\documentclass[twoside,11pt]{article}

\usepackage{jmlr2e}

\usepackage{graphicx}
\usepackage{subfigure}

\usepackage{algorithm}
\usepackage{algorithmic}
\usepackage{bm}
\usepackage{amsmath}
\usepackage{amssymb}
\usepackage{todonotes}
\usepackage{url}
\usepackage{setspace}
\usepackage{tikz}
\usepackage{booktabs}

\usepackage{enumitem}
\usepackage{xspace}

\usepackage{natbib} 
\setcitestyle{round}

\usepackage{hyperref}
\usepackage[capitalise]{cleveref}
\usepackage{booktabs,dcolumn}

\newcommand{\task}{t}

\newcommand{\given}{\,|\,}

\newcommand{\x}{\mathbf{x}}

\newcommand{\EI}{\alpha_\text{EI}}

\newcommand{\E}{\mathbb{E}}
\newcommand{\model}{\mathcal{M}}

\newcommand{\pending}{\omega}
\newcommand{\capacity}{\omega_\text{max}}
\newcommand{\graph}{\mathcal{G}}

\newcommand{\tasks}{\mathcal{T}}
\newcommand{\funcs}{\mathcal{F}}
\newcommand{\resources}{\mathcal{R}} % resources
\newcommand{\xopt}{\x_\star}

\newcommand{\domain}{\mathcal{X}}
\newcommand{\bigO}{\mathcal{O}}
\newcommand{\dimension}{D}
\newcommand{\del}{d}

\newcommand{\appropto}{\mathrel{\vcenter{
  \offinterlineskip\halign{\hfil$##$\cr
    \propto\cr\noalign{\kern2pt}\sim\cr\noalign{\kern-2pt}}}}}

\newcommand{\identity}{\mathbb{I}} %\mathbf{I}

\newcommand{\entropy}{\text{H}}
\newcommand{\data}{\mathcal{D}}
\newcommand{\transpose}{^\top} % can also use ^\itercal

\newcommand{\yvalues}{y^f,y^1,\ldots,y^K}

\newcommand{\functions}{f,c_1,\ldots,c_K}

\newcommand{\gaussian}{\mathcal{N}}

\newcommand{\bo}{Bayesian optimization\xspace}
\newcommand{\acq}{acquisition function\xspace}

\newcommand{\acqs}{acquisition functions\xspace}

\newcommand{\walltime}{wall-clock time\xspace}

\newcommand{\blackbox}{black-box\xspace}

\jmlrheading{17}{2016}{1-53}{12/15; Revised 4/16}{9/16}{J. M. Hern\'andez-Lobato,
M. A. Gelbart, R. P. Adams, M. W. Hoffman and Z. Ghahramani}

\ShortHeadings{Constrained Bayesian Optimization using Information-based Search}
{Hern\'andez-Lobato, Gelbart, Adams, Hoffman and Ghahramani}

\firstpageno{1}

\begin{document}

\title{A General Framework for Constrained Bayesian Optimization using Information-based Search} % miguel thinks 'multi-task is confusing'

% \author{\name Jos\'{e} Miguel Hern\'{a}ndez-Lobato$^*$ \email jmh@seas.harvard.edu \\
%        \addr School of Engineering and Applied Sciences\\
%        Harvard University\\
%        Cambridge, MA 02138, USA
%        \AND
%        \name Michael A. Gelbart$^*$ \email mgelbart@seas.harvard.edu \\
%        \addr Program in Biophysics\\
%        Harvard University\\
%        Cambridge, MA 02138, USA
%        \AND
%        \name Matthew W. Hoffman \email mwh30@cam.ac.uk \\
%        \addr Department of Engineering\\
%        Cambridge University\\
%        Cambridge, CB2 1PZ, UK
%        \AND
%        \name Ryan P. Adams \email rpa@seas.harvard.edu \\
%        \addr School of Engineering and Applied Sciences\\
%        Harvard University\\
%        Cambridge, MA 02138, USA
%        \AND
%        \name Zoubin Ghahramani \email zoubin@eng.cam.ac.uk \\
%        \addr Department of Engineering\\
%        Cambridge University\\
%        Cambridge, CB2 1PZ, UK
% }

\author{\name Jos\'{e} Miguel Hern\'{a}ndez-Lobato$^{1,}$\thanks{Authors contributed equally.}~\hspace*{0.08cm}\email jmh@seas.harvard.edu \\
       \name Michael A. Gelbart$^{3,*}$ \email mgelbart@cs.ubc.ca\\
       \name Ryan P. Adams$^{1,4}$ \email rpa@seas.harvard.edu \\
       \name Matthew W. Hoffman$^2$ \email mwh30@cam.ac.uk \\
       \name Zoubin Ghahramani$^2$ \email zoubin@eng.cam.ac.uk \\
       \addr{1}. School of Engineering and Applied Sciences\\
       Harvard University, Cambridge, MA 02138, USA\\
       \addr{2}. Department of Engineering\\
       Cambridge University, Cambridge, CB2 1PZ, UK\\
       \addr{3}. Department of Computer Science\\
        The University of British Columbia, Vancouver, BC, V6T 1Z4, Canada\\
       \addr{4}. Twitter\\
       Cambridge, MA 02139, USA
}

\editor{Adreas Krause}

\maketitle

\begin{abstract}%   <- trailing '%' for backward compatibility of .sty file
We present an information-theoretic framework for solving global \blackbox optimization
problems that also have \blackbox constraints. Of particular interest to us is to
efficiently solve problems with \emph{decoupled} constraints, in which
subsets of the objective and constraint functions may be evaluated
independently. For example, when the objective is evaluated on a CPU and the
constraints are evaluated independently on a GPU. These problems require an
\acq that can be separated into the contributions of the individual function
evaluations. We develop one such \acq and call it Predictive Entropy Search
with Constraints (PESC). PESC is an approximation to the expected information
gain criterion and it compares favorably to alternative approaches based on improvement
in several synthetic and real-world problems. In addition to this, we consider problems with a
mix of functions that are fast and slow to evaluate. These problems require
balancing the amount of time spent in the meta-computation of PESC and in the actual
evaluation of the target objective. We take a bounded rationality approach and develop a
partial update for PESC which trades off accuracy against speed. We then
propose a method for adaptively switching between the partial and full updates
for PESC.  This allows us to interpolate between versions of PESC that are
efficient in terms of function evaluations and those that are efficient in
terms of \walltime. Overall, we demonstrate that PESC is an effective
algorithm that provides a promising direction towards a unified solution for
constrained Bayesian optimization.  
\end{abstract}


%!TEX root = main.tex
\section{Introduction}
\label{section:introduction}

Many real-world optimization problems involve finding a global minimizer of a black-box objective function subject to a set of black-box constraints
 all being simultaneously satisfied. For example, consider the problem of
optimizing the performance of a speech recognition system, subject to the requirement that it 
operates within a specified time limit. The system may be
implemented as a neural network with hyper-parameters such as the number of
hidden units, learning rates, regularization constants, etc. These
hyper-parameters have to be tuned to minimize the recognition error on some
validation data under a constraint on the maximum runtime of the resulting
system. Another example is the discovery of new materials. Here, we aim to find
new molecular compounds with optimal properties such as the power conversion
efficiency in photovoltaic devices.  Constraints arise from our ability (or inability) to synthesize various molecules.  In this case, the estimation of the
properties of the molecule and its synthesizability can be achieved by
running expensive simulations on a computer. %These are all \emph{black-box} objectives and constraints because we do not have access to their functional forms or derivatives.

More formally, we are interested in finding the global minimum $\xopt$ of a scalar
objective function $f(\x)$ over some bounded domain, typically
${\domain\subset\mathbb{R}^\dimension}$, subject to the non-negativity of a
set of constraint functions ${c_{1},\ldots,c_{K}}$. We write this as
\begin{align}
\min_{\x\in\domain} f(\x) \quad\text{s.t.} \quad c_1(\x) \geq 0,\ldots,c_K(\x)\geq 0\,.
\label{eq:problem}
\end{align}
However, $f$ and ${c_1,\ldots,c_K}$ are unknown and can only be evaluated
pointwise via expensive queries to ``black boxes'' that may provide noise-corrupted
values. Note that we are assuming that~$f$ and each of the constraints~$c_k$ are
defined over the entire space~$\mathcal X$. We seek to find a solution to
\cref{eq:problem} with as few queries as possible. 

For solving unconstrained problems, Bayesian optimization (BO) is a successful approach
to the efficient optimization of black-box functions \citep{Mockus1978}. BO
methods work by applying a Bayesian model to the previous evaluations of the function, with the aim of reasoning about the global structure of the objective function. The
Bayesian model is then used to compute an \emph{\acq} (i.e., expected utility function) that represents how
promising each possible~${\x\in\domain}$ is if it were to be evaluated
next. Maximizing the \acq produces a \emph{suggestion} which is then used as
the next evaluation location. When the evaluation of the objective at the suggested point is complete,
the Bayesian model is updated with the newly collected function observation and the process
repeats. The optimization ends after a maximum number of function evaluations
is reached, a time threshold is exceeded, or some other stopping criterion is met. When this occurs, a
\emph{recommendation} of the solution is given to the user.  This is achieved
for example by optimizing the predictive mean of the Bayesian model, or by choosing the best observed point among the evaluations.  The Bayesian
model is typically a Gaussian process (GP); an in-depth treatment of GPs is
given by \cite{Rasmussen2006}.  A commonly-used \acq is the expected improvement
(EI) criterion \citep{Jones1998}, which measures the expected amount by which
we will improve over some \emph{incumbent} or current best solution, typically given by
the expected value of the objective at the current best recommendation.  
Other acquisition functions aim to approximate the expected information gain 
or expected reduction in the posterior entropy of the global minimizer of the objective
\citep{VillemonteixVW09,hennig-schuler-2012,hernandez2014}.
For more information on BO, we refer to the tutorial by
\cite{brochu-etal-2010a}.

There have been several attempts to extend BO methods to address the
constrained optimization problem in \cref{eq:problem}.
The proposed techniques use GPs and variants of the EI heuristic
\citep{schonlau1998global,parr2013improvement,snoek-2013a,Gelbart2014,Gardner2014,gramacy-augmented-lagrangian,gramacy2010,picheny2014stepwise}.
Some of these methods lack generality since they were designed to work
in specific contexts, such as when the constraints are noiseless or the
objective is known.  Furthermore, because they are based on EI, computing their
\acq requires the current best feasible solution or incumbent: a location in
the search space with low expected objective value and high probability of
satisfying the constraints.  However, the best feasible solution does not exist
when no point in the search space satisfies the constraints with high
probability (for example, because of lack of data). Finally and more
importantly, these methods run into problems when the objective and the constraint
functions are \emph{decoupled}, meaning that the functions $f,c_1,\ldots,c_K$ in
\cref{eq:problem} can be evaluated independently. In
particular, the acquisition functions used by these methods usually consider
joint evaluations of the objective and constraints and cannot produce an
optimal suggestion when only subsets of these functions are being
evaluated.

In this work, we propose a general approach for constrained BO that does
not have the problems mentioned above. Our approach to constraints is based on an extension of Predictive
Entropy Search (PES) \citep{hernandez2014}, an information-theoretic method for unconstrained BO problems.  The resulting technique is called Predictive Entropy Search with
Constraints (PESC) and its \acq approximates the expected information gain with regard to 
the solution of \cref{eq:problem}, which we call $\mathbf{x}_\star$.  At each
iteration, PESC collects data at the location that is the most informative about $\mathbf{x}_\star$, in expectation.  One important property of PESC is that its
\acq naturally separates the contributions of the individual function
evaluations when those functions are modeled independently.  That is, the amount of information that we approximately gain
by jointly evaluating a set of independent functions is equal to the sum of the gains of
information that we approximately obtain by the individual evaluation of each of
the functions. This additive property in its \acq allows PESC to efficiently solve
the general constrained BO problem, including those with decoupled evaluation, something that no other
existing technique can achieve, to the best of our knowledge. 

An initial description of PESC is given by \cite{hernandez2015}. That work
considers PESC only in the coupled evaluation scenario, where all the functions
are jointly evaluated at the same input value. This is the standard setting considered
by most prior approaches for constrained BO. Here, we further extend that
initial work on PESC as follows:
\begin{enumerate}
\item We present a taxonomy of constrained BO
problems. We consider problems in which
the objective and constraints can be split into subsets of functions or
\emph{tasks} that require coupled evaluation, but where different tasks
can be evaluated in a decoupled way. These different tasks
may or may not compete for a limited set of resources.
We propose a general algorithm for solving this type of problems
and then show how PESC can be used for the practical implementation of this algorithm.
\item We analyze PESC in the decoupled scenario. We evaluate the accuracy of
PESC when the different functions (objective or constraint) are evaluated
independently. We show how PESC efficiently solves decoupled problems with an
objective and constraints that compete for the same computational resource.
\item We intelligently balance the computational overhead of the Bayesian
optimization method relative to the cost of evaluating the 
black-boxes. To achieve this, we develop a partial update to the PESC approximation
that is less accurate but faster to compute. We then
automatically switch between partial and full updates so that we can balance
the amount of time spent in the Bayesian optimization subroutines and in the actual
collection of data. This allows us to efficiently solve problems with a mix of
decoupled functions where some are fast and others slow to evaluate.

\end{enumerate}

The rest of the paper is structured as follows. \Cref{section:prior-work}
reviews prior work on constrained BO and considers these methods in the context
of decoupled functions. In \cref{section:decoupling} we present a general
framework for describing BO problems with decoupled functions, which contains
as special cases the standard coupled framework considered in most prior work
as well as the notion of decoupling introduced by \citet{Gelbart2014}.  This
section also describes a general algorithm for BO problems with decoupled
functions.  In \cref{section:pesc} we show how to extend Predictive Entropy
Search (PES) \citep{hernandez2014} to solve \cref{eq:problem} in the context of
decoupling, an approach that we call Predictive Entropy Search with Constraints
(PESC).  We also show how PESC can be used to implement the general algorithm
from \cref{section:decoupling}.  In \cref{section:pesc-fast-updates} we modify
the PESC algorithm to be more efficient in terms of \walltime by adaptively
using an approximate but faster version of the method. In
\cref{section:experiments,section:experiments_decoupling} we perform empirical
evaluations of PESC on coupled and decoupled optimization problems,
respectively. Finally, we conclude in \cref{section:discussion}.

%!TEX root = main.tex
\section{Related Work}
\label{section:prior-work}

Below we discuss previous approaches to \bo with \blackbox constraints, many of
which are variants of the expected improvement (EI) heuristic \citep{Jones1998}. 
In the unconstrained setting, EI measures the expected amount by which
observing the objective $f$ at $\x$ leads to improvement over the current best
recommendation or \emph{incumbent}, the objective value of which is denoted by $\eta$ (thus, $\eta$ has the units of $f$, not $\x$).
The incumbent $\eta$ is usually defined as the lowest 
expected value for the objective over the optimization domain.
The EI acquisition function is then given by
\begin{align}
\EI(\x) & = \!\int\! \max(0, \eta - f(\mathbf{x})) p(f(\mathbf{x})|\mathcal{D}) \,df(\mathbf{x})
= \sigma_f(\x) \left( z_f(\x)\Phi\left( z_f(\x)\right)+\phi\left( z_f(\x)\right) \right)\,,
\end{align}
where $\mathcal{D}$ represents the collected data (previous function evaluations) and
$p(f(\mathbf{x})|\mathcal{D})$ is the predictive distribution for the objective made by a
Gaussian process (GP), $\mu_f(\x)$ and $\sigma^2_f(\x)$ are the GP
predictive mean and variance for $f(\x)$, $z_f(\x) \equiv (\eta-\mu_f(\x)) /
\sigma_f(\x)$, and $\Phi$ and $\phi$ are the standard Gaussian CDF and PDF,
respectively. 

%

% While previous approaches to the problem of Bayesian optimization with unknown
% constraints have been proposed, most are variants of expected improvement (EI)
% \cite{Mockus1978,Jones1998}. Initially proposed by \citet{schonlau1998global}, one
% method of extending EI to the constrained setting considers the expected
% \emph{feasible} improvement, where the constraints are given as above; such
% approaches have recently been independently developed in \citet{Gelbart2014,
% Gardner2014, snoek-2013a}.
% Alternatively \citet{gramacy2010} consider the integrated change in expected improvement
% of all points in the search space with respect to a density given by the probability of feasibility. 
% \citet{picheny2014stepwise} considers the probability of improvement under a
% similar measure. Finally, \citet{gramacy-augmented-lagrangian} propose
% to combine EI with the augmented Lagrangian approach for constrained numerical
% optimization. 
% The strategies used by these methods to select the next evaluation point are all based on the expected level of improvement.  As discussed below, however, this expectation is not always well defined in the presence of constraints.  In contrast to this previous work, the main contribution of this paper is to develop an approach (PESC) that is always well defined.  PESC achieves this by building a Bayesian optimization acquisition function around information gain in the constrained setting, as described in Section~\ref{sec:pesc}.

\subsection{Expected Improvement with Constraints}
\label{section:prior-work:eic}

An intuitive extension of EI in the presence of constraints is to define
improvement as only occurring when the constraints are satisfied. Because we
are uncertain about the values of the constraints, we must weight the original
EI value by the probability of the constraints being satisfied. This results in
what we call Expected Improvement with Constraints (EIC): 
\begin{align}
\label{eq:background:constrained-bo:EIC}
\alpha_\text{EIC}(\x) & = \EI(\x) \prod_{k=1}^K \Pr(c_k(\x) \geq 0|\mathcal{D})
= \EI(\x)\prod_{k=1}^K \Phi\left(\frac{\mu_k(\x)}{\sigma_k(\x)}\right)\, ,
\end{align}
The constraint satisfaction probability factorizes because $f$ and
$c_1,\ldots,c_K$ are modeled by independent GPs.  In this
expression $\mu_k$ and $\sigma^2_k$ are the posterior predictive mean and
variance for $c_k(\x)$. EIC was initially proposed by
\citet{schonlau1998global} and has been revisited by
\citet{parr2013improvement},
\citet{snoek-2013a}, \citet{Gardner2014} and \citet{Gelbart2014}. 

In the constrained setting, the incumbent $\eta$ can be defined as the minimum
expected objective value subject to all the constraints being satisfied at the
corresponding location. However, we can never guarantee that all the
constraints will be satisfied when they are only observed through noisy evaluations.
To circumvent this problem, \cite{Gelbart2014} define $\eta$ as the
lowest expected objective value subject to all the constraints being satisfied
with posterior probability larger than the threshold $1-\delta$, where $\delta$ is a
small number such as $0.05$.  However, this value for $\eta$ still cannot be computed
when there is no point in the search space that satisfies the constraints with posterior
probability higher than $1-\delta$. For example, because of lack of data for
the constraints. In this case, \citeauthor{Gelbart2014} change the original
acquisition function given by \cref{eq:background:constrained-bo:EIC} and
ignore the factor $\EI(\x)$ in that expression.  This allows them to search
only for a feasible location, ignoring the objective $f$ entirely and just
optimizing the constraint satisfaction probability. However, this can lead to
inefficient optimization in practice because the data collected for the objective $f$
is not used to make optimal decisions.

% Their method requires computing the current best feasible solution:
% a location of the search space with high expected objective value and high probability of satisfying the constraints.

% However, when no point in the search space is feasible (that is, 
% satisfies the constraints with high probability, e.g. because of lack of data),
% EI cannot be computed. In this case, they change the acquisition function to
% search only for a feasible location, ignoring the objective $f$ entirely.
% This decoupling between learning the objective and the constraints can produce inefficient results in practice.

\subsection{Integrated Expected Conditional Improvement}
\label{section:background:constrained-bayes-opt-prior-work:ieci}

\citet{gramacy2010} propose an acquisition function called the integrated
expected conditional improvement (IECI), defined as
\begin{align}
\alpha_\text{IECI}(\x) &= \int_\domain \left[\alpha_\text{EI}(\x') - 
\alpha_\text{EI}(\x'|\x) \right]h(\x') \mathrm{d}\x' \, .
\label{eq:ieci}
\end{align}
Here,~$\alpha_\text{EI}(\x')$ is the expected improvement
at~$\x'$,~$\alpha_\text{EI}(\x'|\x)$ is the expected improvement at~$\x'$ given
that the objective has been observed at $\x$ (but without making any
assumptions about the observed value), and $h(\x')$ is an arbitrary density
over~$\x'$. The IECI at~$\x$ is the expected reduction in EI at
$\x'$, under the density~$h(\x')$, caused by observing the objective at~$\x$.
\citeauthor{gramacy2010} use IECI for constrained BO by
setting~$h(\x')$ to the probability of the constraints being satisfied at~$\x'$.
They define the incumbent $\eta$ as the lowest posterior mean for the
objective $f$ over the whole optimization domain, ignoring the fact that the
lowest posterior mean for the objective may be achieved in an infeasible location.

The motivation for IECI is that collecting data at an infeasible location may
also provide useful information. EIC strongly discourages this, because
\cref{eq:background:constrained-bo:EIC} always takes very low values when the
constraints are unlikely to be satisfied. This is not the case with IECI
because \cref{eq:ieci} considers the EI over the whole optimization domain
instead of focusing only on the EI at the current evaluation location, which
may be infeasible with high probability. \citet{Gelbart2014} compare IECI with
EIC for optimizing the hyper-parameters of a topic model with constraints on
the entropy of the per-topic word distribution and show that EIC outperforms
IECI on this problem.

\subsection{Expected Volume Minimization}
\label{section:prior-work:expected-volume}

An alternative approach is given by \citet{picheny2014stepwise}, who proposes
to sequentially explore the location that most decreases the expected volume (EV) of
the feasible region below the best feasible objective value $\eta$ found so
far. This quantity is computed by integrating the product of the probability of
improvement and the probability of feasibility. That is, 
\begin{align}
\alpha_\text{EV}(\x) = \int p[f(\x')\leq \eta] h(\x')\del\x' - 
\int p[f(\x')\leq \min(\eta, f(\x))] h(\x')\del\x'\,,
\label{eq:background:EEV}
\end{align}
where, as in IECI, $h(\x')$ is the probability that the constraints are
satisfied at $\x'$. \citeauthor{picheny2014stepwise} considers noiseless
evaluations for the objective and constraint functions and defines $\eta$ as
the best feasible objective value seen so far or $+\infty$ when no feasible
location has been found.

%, and, as in IAGO (\cref{section:background:information-based-acqs:IAGO}), we
%can ignore the first term on the right-hand side because it does not depend on
%$\x$.  This step-wise uncertainty reduction approach (SURA) is similar to PESC
%in that both methods work by reducing a specific type of  uncertainty measure
%(entropy for PESC and EV for SURA).

A disadvantage of \citeauthor{picheny2014stepwise}'s method is that it requires
the integral in \cref{eq:background:EEV} to be computed over the entire search
domain $\mathcal X$, which is done numerically over a grid on $\x'$. The
resulting \acq must then be globally optimized. This is often performed by
first evaluating the \acq on a grid on~$\x$. The best point in this second grid
is then used as the starting point of a numerical optimizer for the \acq.  This
nesting of grid operations limits the application of this method to small input
dimension $\dimension$. This is also the case for IECI whose \acq in
\cref{eq:ieci} also includes an integral over $\mathcal X$. Our method PESC
requires a similar integral in the form of an expectation with respect to the
posterior distribution of the global feasible minimizer $\xopt$.  Nevertheless,
this expectation can be efficiently approximated by averaging over samples of
$\xopt$ drawn using the approach proposed by \citet{hernandez2014}. This
approach is further described in
\cref{section:implementation:pesc-sampling-x-star}.  Note that the integrals in
\cref{eq:background:EEV} could in principle be also approximated by using
Marcov chain Monte Carlo (MCMC) to sample from the unnormalized density
$h(\mathbf{x}')$. However, this was not proposed by
\citeauthor{picheny2014stepwise} and he only described the grid based method.

\subsection{Modeling an Augmented Lagrangian}
\label{section:prior-work:gramacy-augmented-lagrangian}

\citet{gramacy-augmented-lagrangian} propose to use a
combination of EI and the augmented Lagrangian~(AL) method: an algorithm which
turns an optimization problem with constraints into a sequence of unconstrained
optimization problems. \citeauthor{gramacy-augmented-lagrangian} use BO techniques based on EI to solve the
unconstrained \emph{inner} loop of the AL problem.
When $f$ and $c_1,\ldots,c_K$ are known the unconstrained AL objective is defined as
\begin{align}
L_A(\mathbf{x}|\lambda_1,\ldots,\lambda_K,p) = f(\mathbf{x}) + \sum_{k=1}^K \left[ \frac{1}{2p} \min(0,
c_k(\mathbf{x}))^2 - \lambda_k c_k(\mathbf{x}) \right]\,,\label{eq:auglagrangian}
\end{align}
where ${p > 0}$ is a penalty parameter and $\lambda_1 \geq
0,\ldots,\lambda_K \geq 0$ serve as Lagrange multipliers. The AL method
iteratively minimizes \cref{eq:auglagrangian} with different values for $p$ and
$\lambda_1,\ldots,\lambda_K$ at each iteration. Let $\mathbf{x}^{(n)}_\star$
be the minimizer of \cref{eq:auglagrangian} at iteration $n$ using parameter values~$p^{(n)}$
and~${\lambda_1^{(n)},\ldots,\lambda_K^{(n)}}$. The next parameter values
are~${\lambda_k^{(n+1)} = \max(0,\lambda_k^{(n)} - c_k(\mathbf{x}^{(n)}_\star)
/ p^{(n)})}$ for~${k = 1,\ldots,K}$ and~${p^{(n+1)}=p^{(n)}}$
if~$\mathbf{x}_\star^{(n)}$ is feasible and~${p^{(n+1)}=p^{(n)} / 2}$
otherwise. When~$f$ and~${c_1,\ldots,c_K}$ are unknown we cannot directly minimize
\cref{eq:auglagrangian}. However, if we have observations for~$f$ and~${c_1,\ldots,c_K}$, we can then map such data into observations for the AL
objective. \citeauthor{gramacy-augmented-lagrangian}
fit a GP model to the AL observations and then select the next evaluation
location using the EI heuristic. After collecting the data, the AL parameters
are updated as above using the new values for the constraints and the whole process
repeats.
 
A disadvantage of this approach is that it assumes that the the constraints
$c_1,\ldots,c_k$ are noiseless to guarantee that that $p$ and
$\lambda_1,\ldots,\lambda_K$ can be correctly updated.  Furthermore,
\citet{gramacy-augmented-lagrangian} focus only on the case in which the
objective $f$ is known, although they provide suggestions for extending their
method to unknown $f$. In section \ref{section:experiments:coupled:toy-problem} we show that PESC and
EIC perform better than the AL approach on the synthetic benchmark problem considered by
\citeauthor{gramacy-augmented-lagrangian}, even when the AL method has access to the
true objective function and PESC and EIC do not.

\subsection{Existing Methods for Decoupled Evaluations}
\label{section:prior-work:eic-d}

The methods described above can be used to solve constrained BO problems with
\emph{coupled} evaluations. These are problems in which all the functions
(objective and constraints) are always evaluated jointly at the same input.
\citet{Gelbart2014} consider extending the EIC method from
\cref{section:prior-work:eic} to the \emph{decoupled} setting, where the
different functions can be independently evaluated at different input
locations. However, they identify a problem with EIC in the decoupled scenario.
In particular, the EIC utility function requires two conditions to produce positive values.
First, the evaluation for the objective must achieve a lower value
than the best feasible solution so far and, second, the evaluations for the constraints
must produce non-negative values.  When we evaluate only one function
(objective or constraint), the conjunction of these two conditions cannot be
satisfied by a single observation under a myopic search policy. Thus, the new
evaluation location can never become the new incumbent and the EIC is zero
everywhere.  Therefore, standard EIC fails in the decoupled setting. 

\citet{Gelbart2014} circumvent the problem mentioned above by treating
decoupling as a special case and using a two-stage acquisition function: first,
the next evaluation location~$\x$ is chosen with EIC, and then, given $\x$, the
task (whether to evaluate the objective or one of the constraints) is chosen
according to the expected reduction in the entropy of the global feasible
minimizer $\xopt$, where the entropy computation is approximated using Monte
Carlo sampling as proposed by \citet{VillemonteixVW09}. We call the resulting method
EIC-D. Note that the two-stage decision process used by EIC-D is sub-optimal
and a joint selection of $\x$ and the task should produce better results. As
discussed in the sections that follow, our method, PESC, does not suffer from
this disadvantage and furthermore, can be extended to a wider range of
decoupled problems than EIC-D can.

%!TEX root = main.tex

\section{Decoupled Function Evaluations and Resource Allocation}

\label{section:decoupling}

%Prior work on constrained \bo focuses on the \emph{coupled} scenario, in which
%the objective and constraint(s) must be evaluated jointly;  in our example,
%this corresponds to jointly evaluating the objective and constraint functions
%at each iteration;  in other words, in the coupled scenario one cannot
%evaluate one function without evaluating the other(s).  Here we introduce \bo
%with \emph{decoupling}, in which these different tasks (objective and
%constraints) need not be jointly evaluated. The neural network example is
%naturally decoupled, because the running time (constraint) can presumably be
%evaluated without training the network, which must be done to evaluate the
%objective. 

%Another example of decoupling is the cookie optimization discussed in
%\cref{chapter:intro}: determining the tastiness of a cookie recipe requires a
%taste test with a group of participants, whereas computing the number of
%calories can be done independently without such a group.

We present a framework for describing constrained BO problems. We say
that a set of functions (objective or constraints) are \emph{coupled} when they always
require joint evaluation at the same input location. We say that they are
\emph{decoupled} when they can be evaluated independently at different inputs.  In
practice, a particular problem may exhibit coupled or decoupled functions or a
combination of both.  An example of a problem with coupled functions is given
by a financial simulator that generates many samples from the distribution of
possible financial outcomes. If the objective function is the expected profit and the
constraint is a maximum tolerable probability of default, then these two
functions are computed jointly by the same simulation and are thus coupled to
each other. An example of a problem with decoupled functions is the
optimization of the predictive accuracy of a neural network speech recognition
system subject to prediction time constraints. In this case different neural
network architectures may produce different predictive accuracies and different
prediction times. Assessing the prediction time may not require training the
neural network and could be done using arbitrary network weights. Thus, we can evaluate the timing constraint without evaluating the accuracy objective. 

When problems exhibit a combination of coupled and decoupled functions, we can
then partition the different functions into subsets of functions that require
coupled evaluation. We call these subsets of coupled functions \emph{tasks}. In
the financial simulator example, the objective and the constraint form the only
task. In the speech recognition system there are two tasks, one for
the objective and one for the constraint. Functions within different tasks are
decoupled and can be evaluated independently. These tasks may or may not
compete for a limited set of \emph{resources}. For example, two tasks that both require 
the performance of expensive computations may have to compete for using a single available CPU.
An example with no competition is given by two tasks, one which performs
computations in a CPU and another one which performs computations in a GPU.
Finally, two competitive tasks may also have different evaluation costs and
this should be taken into account when deciding which one is going to be
evaluated next.
%As another example, consider the case in which the data used to evaluate our
%objective function are highly private, and thus the objective function
%evaluation can only be performed on a particular machine with access to these
%data. On the other hand, if the constraint evaluation does not require this
%data set, it could be performed on any machine, or perhaps even in parallel on
%a large number of available CPUs. 

In the previous section we showed that most existing methods for constrained BO
can only address problems with coupled functions. Furthermore, the extension of
these methods to the decoupled setting is difficult because most of them are
based on the EI heuristic and, as illustrated in
\cref{section:prior-work:eic-d}, improvement can be impossible with decoupled
evaluations.  A decoupled problem can, of course, be coupled artificially and then solved as a coupled problem with existing methods.  We examine this approach 
here with a thought experiment and with empirical evaluations in
\cref{section:experiments_decoupling}. Returning to our time-limited speech recognition
system, let us consider the cost of evaluating each of the tasks. Evaluating
the objective requires training the neural network, which is a very expensive
process.  On the other hand, evaluating the constraint (run time) requires
only to time the predictions made by the neural network and this could be done
without training, using arbitrary network weights. Therefore, evaluating the
constraint is in this case much faster than evaluating the objective.
 %assume that the software can be executed quickly and thus an estimate of
 %running time can be achieved by averaging over a few fast executions, whereas
 %the performance evaluation relies on a large data set and takes potentially
 %orders or magnitude more time to complete; in other words, the objective
 %function is much slower than the constraint function. 
%Measuring the running time takes perhaps a few seconds; in short, it is orders
%of magnitude faster than evaluating the objective.
%The naive approach to decoupled problems, to simply treat them as coupled and
%use existing methods for optimization with coupled constraints. However, if
%the tasks are decoupled, it can be highly wasteful to disregard this added
%flexibility and treat the problem as coupled. A thought experiment
%illustrating this point is as follows: 
%Imagine now evaluating both tasks at some input location, only to discover
%immediately that the run time is extremely slow, and thus the constraint is
%violated. To continue training the neural network is then very wasteful,
%because that input location (and probably a nearby region of input space) has
%already been ruled out. 
In a decoupled framework, one could first measure the
run time at several evaluation locations, gaining a sense of the constraint
surface. Only then would we incur the significant expense of evaluating the
objective task, heavily biasing our search towards locations that are
considered to be feasible with high probability.  Put another way, artificially coupling the tasks becomes increasingly inefficient as the cost differential is increased; for example, one might spend a week examining one aspect of a design that could have been ruled out within seconds by examining another aspect.
%The larger the cost
%differential between the two tasks, the more wasteful it becomes to artificially treat the
%problem as coupled.
%\Cref{chapter:experiments} shows the benefits of decoupling in experiments.

In the following sections we present a formalization of constrained \bo
problems that encompasses all of the cases described above. We then show that
our method, PESC (\cref{section:pesc}), is an effective practical solution to
these problems because it naturally separates the contributions of
the different function evaluations in its \acq.

\subsection{Competitive Versus Non-competitive Decoupling and Parallel BO}
\label{section:decoupled:competitive-vs-non-competitive}

We divide the class of problems with decoupled functions into two sub-classes,
which we call \emph{competitive decoupling} (CD) and \emph{non-competitive
decoupling} (NCD).  CD is the form of decoupling considered by
\citet{Gelbart2014}, in which two or more tasks compete for the same resource.
This happens when there is only one CPU available and the optimization problem
includes two tasks with each of them requiring a CPU to perform some expensive
simulations.  In contrast, NCD refers to the case in which tasks require the
use of different resources and can therefore be evaluated independently, in parallel. This occurs, for example, when one of the two tasks uses a CPU and the other task uses a GPU.  %We also consider as a special case one task that
%requires the use of a specific resource when multiple instances of that
%resource are available.  This happens for example when one task performs some
%expensive computations on a CPU and we have access to a computer cluster with
%many CPUs that can evaluate the task in parallel. This special case is the

Note that NCD is very related to
\emph{parallel} \bo 
\citep[see e.g.,][]{ginsbourger2010dealing,snoek-etal-2012b}. %is a special case of NCD in which a resource (such as a CPU) is replicated many times (such as in a compute cluster).
In both parallel BO and NCD we perform multiple task evaluations
concurrently, where each new evaluation location is selected optimally
according to the available data and the locations of all the currently pending
evaluations. The difference between parallel BO and NCD is that in NCD the
tasks whose evaluations are currently pending may be different from the task that
will be evaluated next, while in parallel BO there is only a single task. Parallel BO conveniently fits into the general framework described below.

\subsection{Formalization of Constrained Bayesian Optimization Problems} \label{section:decoupled:formalization}

We now present a framework for describing constrained BO problems of the
form given by \cref{eq:problem}. Our framework can be used to represent general
problems within any of the categories previously described, including coupled
and decoupled functions that may or may not compete for a limited number
of resources, each of which may be replicated multiple times.
Let~$\funcs$ be the set of
functions~$\{f,c_1,\ldots,c_K\}$ and let the set of tasks $\tasks$ be a
partition of~$\funcs$ indicating which functions are coupled and must be
jointly evaluated. Let~${ \resources=\{r_1,\ldots,r_{|\resources|}\} }$ be the set
of resources available to solve this problem. We encode the relationship
between tasks and resources with a bipartite graph~${\graph = \left(
\mathcal{V}, \mathcal{E} \right)}$ with vertices~${\mathcal{V}=\tasks\cup\resources}$ and edges~${\{\task\sim r\}\in\mathcal{E}}$
such that~$\task\in\tasks$ and~$r\in\resources$. The interpretation of an edge
$\{\task\sim r\}$ is that task $\task$ can be performed on resource~$r$.
(We do not address the case in which a task requires
multiple resources to be executed; we leave this as future work.) 
We also introduce a \emph{capacity}~$\capacity$ for each resource~$r$. The
capacity $\capacity(r)\in\mathbb{N}$ represents how many tasks may be
simultaneously executed on resource $r$; for example, if $r$ represents a
cluster of CPUs,~$\capacity(r)$ would be the number of CPUs in the cluster.
Introducing the notion of capacity is simply a matter of convenience since
it is equivalent to setting all capacities to one and replicating each resource
node in~$\graph$ according to its capacity. 

%Assumption: each task can be run in parallel assuming copies of that resource
%are available. sure, this is a bit subtle. probably not even worth mentioning

\begin{figure}[t]
\centering
  \includegraphics[width=0.7\columnwidth]{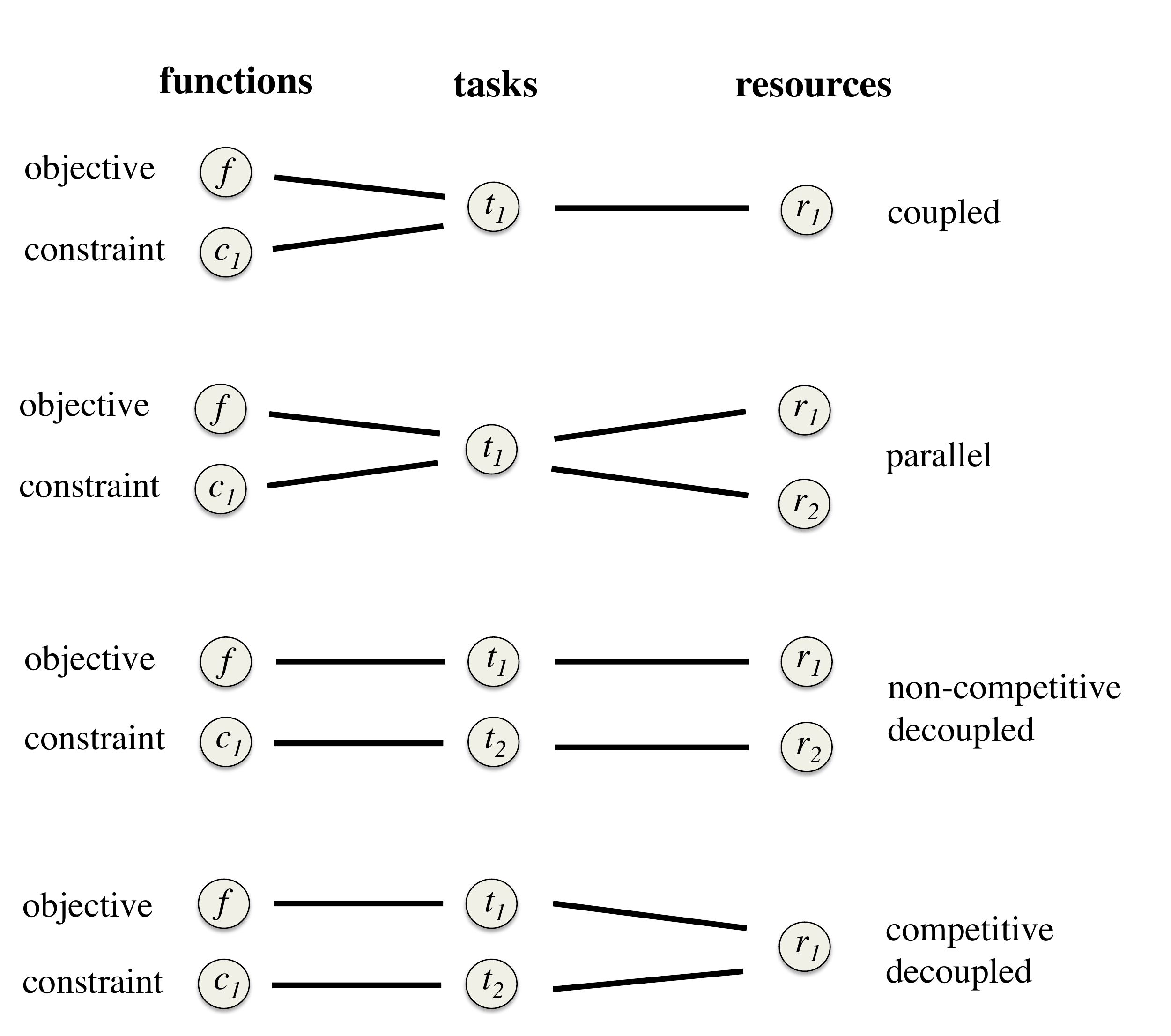}
\caption[Taxonomy of decoupled constraints.]{Schematic comparing the coupled,
parallel, non-competitive decoupled (NCD), and competitive decoupled (CD)
scenarios for a problem with a single constraint $c_1$. In each
case, the mapping between tasks and resources (the right-hand portion of the
figure) is the bipartite graph $\graph$.} \label{fig:decoupling:schematic}
\end{figure}

We can now formally describe problems with coupled evaluations as well as NCD
and CD. In particular, coupling occurs when two functions $g_1$ and $g_2$ 
belong to the same task $\task$. If this task can be evaluated on multiple
resources (or one resource with $\capacity>1$), then this is parallel \bo.  NCD
occurs when two functions $g_1$ and $g_2$ belong to different tasks $\task_1$
and $\task_2$, which themselves require different resources $r_1$ and $r_2$,
(that is, $\task_1\sim r_1$, $\task_2 \sim r_2$ and $r_1 \neq r_2$). CD occurs when two
functions $g_1$ and $g_2$ belong to \emph{different} tasks $\task_1$ and
$\task_2$ (decoupled) that require the \emph{same} resource $r$ (competitive).
These definitions are visually illustrated in \cref{fig:decoupling:schematic}.
The definitions can be trivially extended to cases with more than two functions. The most
general case is an arbitrary task-resource graph $\graph$ encoding a
combination of coupling, NCD, CD and parallel \bo.

%In the next section we describe how PESC can be applied to any such graph.

\subsection{A General Algorithm for Constrained Bayesian Optimization}
\label{section:decoupling:pesc-decoupled}

In this section we present a general algorithm for solving constrained \bo
problems specified according to the formalization from the previous section.
Our approach relies on an \acq that can measure the expected utility of evaluating
any arbitrary subset of functions, that is, of any possible task. When an
\acq satisfies this requirement we say that it is \emph{separable}. As
discussed in \cref{sec:pesc_acquisition_function_new}, our method PESC has %an \acq with 
this property, when the functions are modeled as independent. %In this case PESC's \acq is additive, that is, the utility of evaluating a set of functions is equal to the sum of the utilities obtained by the evaluations of the individual functions. 
This property makes PESC an effective
solution for the practical implementation of our general algorithm.
By contrast, the EIC-D method of \cite{Gelbart2014} is not separable and cannot
be applied in the general case presented here.

% % move some of this to the intro?
% When performing Bayesian optimization with multiple tasks, the \acq must distinguish not only the value of evaluation at different locations $\x$, but also the value of evaluation different tasks.
% \cref{section:eic:decoupled} proposes EIC-D a variant of EIC to address problems with competitive decoupling. As described, the method selects $\x$ first and then $\task$, and thus does not make full use of the available information in the way that a joint decision would. Furthermore, the method is tailored to the specific situation and is not easily adapted to other cases such as non-competitive decoupling.

% PESC does not have these issues because the \acq is \emph{separable} as discussed in \cref{section:pesc:discussion}. A separable \acq is able to provide a measure of promise for evaluating a particular subset of functions in $\funcs$ (typically, those within a single task). On the other hand, an \acq that is not separable can only provide a measure of promise for evaluating all tasks at a particular $\x$. As discussed in \cref{section:pesc:discussion:separability}, PESC is a separable \acq.

%In fact, the PESC \acq is a sum of contributions from each task; that is, the PESC acquisition function for jointly evaluating tasks $\task_1$ and $\task_2$ at $\x$ is simply the sum of the PESC \acq for evaluation task $\task_1$ at $\x$ and $\task_2$ at $\x$.

\Cref{algorithm:decoupling:general} provides a general procedure for
solving constrained \bo problems. The inputs to the algorithm are
the set of functions~$\funcs$, the set of tasks~$\tasks$, the set of
resources~$\resources$, the task-resource
graph~${ \graph=(\mathcal{T}\cup\mathcal{R},\mathcal{E})}$, an acquisition
function for each task, that is,~$\alpha_t$ for $t\in \mathcal{T}$, the search
space~$\domain$, a Bayesian model~$\model$, the initial data set~$\data$,
the resource query functions~$\omega$ and~$\omega_\text{max}$ and a confidence level $\delta$ for making a final recommendation. Recall that
$\omega_\text{max}$ indicates how many tasks can be simultaneously executed on
a particular resource. The function~$\omega$ is introduced here to indicate how
many tasks are currently being evaluated in a resource. The acquisition
function $\alpha_t$ measures the utility of evaluating task $t$ at the location $\mathbf{x}$. This acquisition function depends
on the predictions of the Bayesian model $\mathcal{M}$.  The separability
property of the original \acq guarantees that we can compute an $\alpha_t$ 
for each $t\in \mathcal{T}$.

\begin{algorithm}[tb]
   \caption{A general method for constrained \bo.}
   \label{algorithm:decoupling:general}
\begin{algorithmic}[1] % the "1" means put a line number at every line
   \STATE {\bfseries Input:} $\funcs$, $\graph=(\mathcal{T}\cup\mathcal{R},\mathcal{E})$,
    $\alpha_t$ for $t\in \mathcal{T}$, $\domain$, $\model$, $\data$, $\omega$,~$\omega_\text{max}$ and $\delta$.

%    Set of functions:~$\funcs$,\\
%    Set of tasks:~$\tasks$,\\
%    Set of resources:~$\resources$,\\
%    Task-resource graph:~$\graph=(\mathcal{T}\cup\mathcal{R},\mathcal{E})$,\\
%    Acquisition function for each task:~$\alpha_t$ for $t\in \mathcal{T}$,\\
%    Search space:~$\domain$,\\
%    Bayesian model:~$\model$,\\
%    Initial data set:~$\data$,\\
%    Resource query functions:~$\omega$ and~$\omega_\text{max}$.
   \REPEAT
   \FOR{$r\in\resources$ such that $\pending(r)<\capacity(r)$}
   \STATE Update $\model$ with any new data in $\data$
   \FOR{$t\in\tasks$ such that $\{\task\sim r\}\in\mathcal{E}$}
   \STATE $\x_\task^* \leftarrow \arg \max_{\x\in\domain} \alpha_\task(\x|\mathcal{M})$ 
   \STATE $\alpha_t^* \leftarrow \alpha_\task(\x_\task^*|\mathcal{M})$
   \ENDFOR
   \STATE $\task^{*} \leftarrow \arg \max_{\task} \, \alpha_\task^*$
   \STATE Submit task $\task^*$ at input $\x^*_{\task^*}$ to resource $r$
   \STATE Update $\model$ with the new pending evaluation
   \ENDFOR
   \UNTIL{termination condition is met}
   \STATE {\bfseries Output:}
    $\arg \min_{\x\in\domain} \E_\model[f(\x)] \;\, \mathrm{s.t.} \; 
   p(c_1(\x) \geq 0,\ldots, c_K(\x)\geq0|\mathcal{M}) \geq 1 - \delta$
\end{algorithmic}
\end{algorithm}

\Cref{algorithm:decoupling:general} works as follows. First, in line 3, we
iterate over the resources, checking if they are available. Resource $r$ is
available if its number of currently running jobs~$\pending(r)$ is less than
its capacity~$\capacity(r)$. Whenever resource $r$ is available, we check in
line 4 if any new function observations have been collected.  If this is the case, we then update
the Bayesian model $\mathcal{M}$ with the new data (in most cases we will have new data since the resource $r$
probably became available due to the completion of a previous task).  Next, we iterate in line 5 over the tasks $t$ that can
be evaluated in the new available resource $r$ as dictated by~$\graph$.  In line 6
we find the evaluation location $\mathbf{x}_t^*$ that maximizes the utility
obtained by the evaluation of task $t$, as indicated by the task-specific
acquisition function $\alpha_t$. In line 7 we obtain the corresponding maximum
task utility $\alpha_t^*$. In line 9, we then maximize over tasks, selecting
the task $\task^*$ with highest maximum task utility $\alpha^{*}_\task$ (this is the ``competition'' in CD). Upon
doing so, the pair $(\task^*, \x^{*}_{\task^*})$ forms the next
\emph{suggestion}. This pair represents the experiment with the highest
acquisition function value over all possible $(\task,\x)$ pairs in $\mathcal{T}\times \mathcal{X}$ that can be run
on resource $r$. In line 10, we evaluate the selected task at resource $r$ and
in line 11 we update the Bayesian model $\mathcal{M}$ to take into account that
we are expecting to collect data for task $\task^*$ at input
$\x^{*}_{\task^*}$. This can be done for example by drawing virtual data from~$\mathcal{M}$'s predictive distribution and then
averaging across the predictions made when each virtual data point is assumed
to be the data actually collected by the pending evaluation \citep{schonlau1998global,snoek-etal-2012b}.  
In line 13 the
whole process repeats until a termination condition is met. Finally, in line
14, we give to the user a final \emph{recommendation} of the solution to the optimization problem. 
This is the input that attains the lowest
expected objective value subject to all the constraints being satisfied with
posterior probability larger than $1-\delta$, where $\delta$ is maximum allowable probability of the
recommendation being infeasible according to $\mathcal{M}$.

%We now describe how \Cref{algorithm:decoupling:general} addresses the coupled,
%parallel BO, NCD, and CD cases from \cref{fig:decoupling:schematic}. In the
%coupled case, the loops at lines 3 and 5 are over a single resource and a
%single task, respectively. At each iteration, line 6 maximizes the only
%existing task-specific \acq over $\x\in\mathcal{X}$. After this, the
%maximization over tasks at line 8 is trivial because there is only one task.
%In line 11, $\mathcal{M}$ is updated with some virtual data sampled from the
%posterior. In the coupled case this virtual data is never actually used and
%line 11 can be skipped. The reason for this is that as soon as the resource is
%available again, we will also have new data that replaces the virtual
%observations. Parallel BO is similar to the coupled case. However, we now loop
%over multiple resources at line 3 and we now do use the virtual data produced
%in line 11, since one resource might be available when the task submitted to
%another resource is still pending.  The NCD case is the same as parallel BO,
%except that each resource has now its own associated task-specific \acq.
%Finally, in the CD case, multiple tasks ``compete'' at line 9 and the task
%$\task^*$ with the highest utility value is ultimately selected, along with its
%best evaluation location $\x^*_{\task^*}$.  
\Cref{algorithm:decoupling:general} can solve problems that exhibit
any combination of coupling, parallelism, NCD, and CD.

%Note that it is straightforward to apply the fast method PESC-F (\cref{section:pesc-fast-updates}) by combining \cref{algorithm:decoupling:general,algorithm:pesc:fast-slow}. Note also that \cref{algorithm:decoupling:general} is not specific to PESC but can be applied with any separable \acq.

%Thus, a joint decision $(t,\x)$ is made by maximizing the acquisition function over the entire decision space~$\tasks\otimes\mathcal{X}$.

% Line 7 is where you need this property of an acquisition function
% -- we don't actually need the additivity, do we?? oh i see. i am missing something here??
% right, there is another case. i see. ... hmm. ok. hmm
% adding tasks. when might be evaluate multiple tasks... hmmmm??
% get this straight.
% this only really matters when you are assigning multiple tasks at the same time.
% that's the integer program situation that we're not considering here. we're only assigning
% one task at a time.
% but, interestingly, i think that if you assume additivity, and you have a bunch of tasks/resources
% to simultanously assign, isn't it the case that you can just be greedy and take the best task first
% and then the next best, etc. i mean, isn't it the case, when things are additive, that you don't really
% care about what combinations of tasks you are observing, so you don't really need the whole
% integer program thing?

\subsection{Incorporating Cost Information}

\Cref{algorithm:decoupling:general} always selects, among a group of
competitive tasks, the one whose evaluation produces the highest utility value.
However, other cost factors may render the evaluation of one task more
desirable than another. The most salient of these costs is the run time or
duration of the task's evaluation, which could depend on the
evaluation location~$\mathbf{x}$. For example, in the neural network speech
recognition system, one of the variables to be optimized may be the number of
hidden units in the neural network. In this case, the run time of an evaluation
of the predictive accuracy of the system is a function of $\mathbf{x}$ since
the training time for the network scales with its size.
\citet{snoek-etal-2012b} consider this issue by automatically measuring the
duration of function evaluations. They model the duration as a function
of~$\x$ with an additional Gaussian process (GP). \citet{swersky-etal-2013a}
extend this concept over multiple optimization tasks so that an independent GP
is used to model the unknown duration of each task. This approach can be
applied in \cref{algorithm:decoupling:general} by penalizing the \acq for task
$\task$ with the expected cost of evaluating that task. In particular, we can
change lines 6 and 7 in \Cref{algorithm:decoupling:general} to 
\begin{itemize}
\item[6:] \hspace{1cm} $\x_\task^* \leftarrow \arg \max_{\x\in\domain} \alpha_\task(\x|\mathcal{M}) / \zeta_\task(\x)$
\vspace{-0.15cm}
\item[7:] \hspace{1cm} $\alpha_t^* \leftarrow \alpha_\task(\x_\task^*|\mathcal{M}) / \zeta_\task(\x_\task^*)$
\end{itemize}
where $\zeta_\task(\x)$ is the expected cost associated with the evaluation of
task $\task$ at $\x$, as estimated by a model of the collected cost data.
When taking into account task costs modeled by Gaussian processes, the total number of GP models used by
\cref{algorithm:decoupling:general} is equal to the number of functions in the
constrained BO problem plus the number of tasks, that is,~$|\funcs|+|\tasks|$.
Alternatively, one could fix the cost functions $\zeta_\task(\x)$
\emph{a priori} instead of learning them from collected data.

\section{Predictive Entropy Search with Constraints (PESC)}
\label{section:pesc}

\newcommand{\argmax}[1]{\underset{#1}{\operatorname{arg}\,\operatorname{max}}\;}

\newcommand*\circled[1]{\tikz[baseline=(char.base)]{
            \node[shape=circle,draw,inner sep=0.4pt] (char) {#1};}}

To implement \Cref{algorithm:decoupling:general} in practice we need to
compute an \acq that is separable and can measure the utility of evaluating an
arbitrary subset of functions. In this section we describe how to achieve this.

Our acquisition function approximates the expected gain of information about the
solution to the constrained optimization problem, which is denoted by~$\xopt$. Importantly,
our approximation is additive.  For example, let $\mathcal{A}$ be a set of
functions and let $I(\mathcal{A})$ be the amount of information that we
approximately gain in expectation by jointly evaluating the functions in
$\mathcal{A}$.  Then $I(\mathcal{A})=\sum_{a\in\mathcal{A}}I(\{a\})$. Although
our acquisition function is additive, the exact expected gain of information is
not. Additivity is the result of a factorization assumption in our
approximation (see \cref{sec:ep_approximation_light} for further details). The good results obtained in our
experiments seem to support that this is a reasonable assumption.  Because of
this additive property, we can compute an \acq for any possible subset of $f$,
$c_1,\ldots,c_K$ using the individual acquisition functions for these functions
as building blocks. 

%We now describe Predictive Entropy Search with Constraints (PESC), a general
%method for solving constrained \bo problems of the form given by
%\cref{eq:problem}. \todo{This basically is the same statement as what Alg 1 does.  Perhaps it would be more accurate to say that this section describes an acquisition heuristic that matches the previous formalism?} 
%

%PESC sequentially collects data by approximately maximizing
%the expected gain of information on the solution to the optimization problem,
%denoted by~$\xopt$. A characteristic of PESC is that its \acq is separable. In
%particular, PESC can easily approximate the expected gain of information that
%is obtained by evaluating an arbitrary subset of the functions $f$,
%$c_1,\ldots,c_K$. The reason for this is that PESC's \acq is additive. For
%example, let $\mathcal{A}$ be a set of functions and let $I(\mathcal{A})$ be
%the amount of information that we approximately gain by jointly evaluating the
%functions in $\mathcal{A}$. Then
%$I(\mathcal{A})=\sum_{a\in\mathcal{A}}I(\{a\})$. Note that, although PESC's acquisition
%function is additive, the exact gain of information is not. Additivity results
%from a factorization assumption used by PESC to approximate the exact
%information gain (see \cref{sec:ep_approximation_light} for further details).
%Our experiments seem to support that this is a reasonable assumption. 

We follow \cite{MacKay:1992:IOF:148167.148185} and measure information about
$\xopt$ by the differential entropy of $p(\xopt|\mathcal{D})$, where
$\mathcal{D}$ is the data collected so far. The distribution
$p(\xopt|\mathcal{D})$ is formally defined in the unconstrained case by
\cite{hennig-schuler-2012}. In the constrained case $p(\xopt|\mathcal{D})$ can
be understood as the probability distribution determined by the following sampling process.
First, we draw $f$, $c_1,\ldots,c_K$ from their posterior
distributions given $\mathcal{D}$ and second, we minimize the sampled $f$
subject to the sampled $c_1,\ldots,c_K$ being non-negative, that is, we solve
\cref{eq:problem} for the sampled functions. The solution to \cref{eq:problem}
obtained by this procedure represents then a sample from
$p(\xopt|\mathcal{D})$.  

We consider first the case in which all the black-box functions $f,c_1,\ldots,c_K$ are evaluated
at the same time (coupled).
Let $\text{H}\left[\xopt\given\data\right]$ denote the differential entropy of
$p(\xopt|\mathcal{D})$ and let $y_f$, $y_{c_1},\ldots,y_{c_K}$ denote the measurements
obtained by querying the black-boxes for $f$, $c_1,\ldots,c_K$ at the input
location $\mathbf{x}$.  We encode these
measurements in vector form as
${\mathbf{y}=(y_f,y_{c_1},\ldots,y_{c_K})^\text{T}}$.
Note that 
$\mathbf{y}$ contains the result of the evaluation of all the functions 
at $\mathbf{x}$, that is, 
the objective $f$ and the constraints $c_1,\ldots,c_K$.
We aim to collect data at the
location that maximizes the expected information gain or
the expected reduction in the entropy of $p(\xopt|\mathcal{D})$. The corresponding acquisition function
is
\begin{align}
\alpha(\mathbf{x}) = \text{H}\left[\xopt\given\data\right] - \E_{\mathbf{y}\given\data,\mathbf{x}} 
\left[\text{H}\left[\xopt\given\data\cup\{(\mathbf{x},\mathbf{y})\}\right] \right]\,.
\label{eq:original_acquisition_pesc}
\end{align}
In this expression,
$\text{H}\left[\xopt\given\data\cup\{(\mathbf{x},\mathbf{y})\}\right]$ is the
amount of information on $\xopt$ that is available once we have collected new
data $\mathbf{y}$ at the input location $\mathbf{x}$. However, this new
$\mathbf{y}$ is unknown because it has not been collected yet. To circumvent this
problem, we take the expectation with respect to the predictive distribution
for $\mathbf{y}$ given $\mathbf{x}$ and $\mathcal{D}$. This produces an
expression that does not depend on $\mathbf{y}$ and could in principle be readily computed. 

A direct computation of \cref{eq:original_acquisition_pesc} is
challenging because it requires evaluating the entropy of the intractable
distribution $p(\xopt\given\data)$  when different pairs
$(\mathbf{x},\mathbf{y})$ are added to the data.
To simplify computations, we note that \cref{eq:original_acquisition_pesc} is the mutual information between
$\xopt$ and $\mathbf{y}$ given $\mathcal{D}$ and $\mathbf{x}$, which we denote by
$\text{MI}(\xopt, \mathbf{y})$. The mutual information operator is symmetric,
that is, $\text{MI}(\xopt, \mathbf{y}) = \text{MI}(\mathbf{y}, \xopt)$.
Therefore, we can follow \cite{houlsby2012} and swap the random variables
$\mathbf{y}$ and $\xopt$ in \cref{eq:original_acquisition_pesc}.  The result is
a reformulation of the original equation that is now expressed in terms of
entropies of predictive distributions, which are easier to approximate:
\begin{align}
\alpha(\mathbf{x}) = 
\text{H}\left[\mathbf{y}\given\data,\mathbf{x}\right] - \E_{\xopt\given\data}
\left[\text{H}\left[\mathbf{y}\given\data,\mathbf{x},\xopt\right] \right]\,.
\label{eq:new_acquisition_pesc}
\end{align}
This is the same reformulation used by Predictive Entropy Search (PES) \citep{hernandez2014} for unconstrained \bo, but extended to the case where $\mathbf{y}$ is a vector rather than a scalar. Since we focus
on constrained optimization problems, we call our method Predictive Entropy Search with
Constraints (PESC). \cref{eq:new_acquisition_pesc} is used by PESC to
efficiently solve constrained Bayesian optimization problems with decoupled
function evaluations. In the following section we describe how to obtain a computationally efficient approximation
to \cref{eq:new_acquisition_pesc}. We also show that the resulting approximation is separable.

\subsection{The PESC Acquisition Function}\label{sec:pesc_acquisition_function_new}

We assume that the functions $f$, $c_1,\ldots,c_K$ are independent samples from
Gaussian process (GP) priors and that the noisy measurements $\mathbf{y}$ returned
by the black-boxes are obtained by adding Gaussian noise to the noise-free function
evaluations at $\mathbf{x}$.
Under this Bayesian model for the data, the first term
in \cref{eq:new_acquisition_pesc} can be computed exactly. In particular,
\begin{align}
\text{H}\left[\mathbf{y}\given\data,\mathbf{x}\right] = 
\sum_{i=1}^{K+1} \frac{1}{2}\log \sigma_i^2(\x) + \frac{K+1}{2}\log(2\pi e)\,,
\label{eq:pesc_first_term}
\end{align}
where $\sigma_i^2(\x)$ is the predictive variance for $y_i$ at $\x$ and $y_i$
is the $i$-th entry in $\mathbf{y}$.
To obtain this formula we have used the fact that $f$,
$c_1,\ldots,c_K$ are generated independently, so that
$\text{H}\left[\mathbf{y}\given\data,\mathbf{x}\right] = \sum_{i=1}^{K+1}
\text{H}\left[y_i\given\data,\mathbf{x}\right]$, and that
$p(y_i\given\data,\mathbf{x})$ is Gaussian with variance parameter $\sigma_i^2(\x)$ given
by the GP predictive variance \citep{Rasmussen2006}:
\begin{align}
\sigma_i^2(\x) = k_i(\x) - \mathbf{k}_i(\x)^\text{T} \mathbf{K}^{-1}_i \mathbf{k}_i(\x) + \nu_i\,,\quad\quad i=1,\ldots,K+1\,,
\end{align}
where $\nu_i$ is the variance of the additive Gaussian noise in the $i$-th
black-box, with $f$ being the first one and $c_K$ the
last one.  The scalar $k_i(\x)$ is the prior variance of the noise-free
black-box evaluations at $\x$.  The vector $\mathbf{k}_i(\x)$ contains the prior
covariances between the black-box values at $\x$ and at those locations for
which data from the black-box is available. Finally, $\mathbf{K}_i$ is a
matrix with the prior covariances for the noise-free black-box evaluations at
those locations for which data is available.

The second term in \cref{eq:new_acquisition_pesc}, that is,
$\E_{\xopt\given\data}
\left[\text{H}\left[\mathbf{y}\given\data,\mathbf{x},\xopt\right] \right]$,
cannot be computed exactly and needs to be approximated. We do this
operation as follows. $\circled{1}$: The expectation with respect to
$p(\xopt\given\data)$ is approximated with an empirical average over $M$
samples drawn from $p(\xopt\given\data)$. These samples are generated by
following the approach proposed by \cite{hernandez2014} for sampling $\xopt$ in the
unconstrained case. We draw approximate posterior samples of
$f,c_1,\ldots,c_K$, as described by \citet[][Appendix~A]{hernandez2014}, and
then solve \cref{eq:problem} to obtain $\xopt$ given the sampled functions. More details can be
found in \cref{section:implementation:pesc-sampling-x-star} of this document. Note that this approach only applies for stationary kernels, but this class includes popular choices such as the squared exponential and Mat\'{e}rn kernels. $\circled{2}$: We assume that the
components of $\mathbf{y}$ are independent given $\data$, $\mathbf{x}$ and
$\xopt$, that is,
we assume that the evaluations of $f$, $c_1,\ldots,c_K$ at $\mathbf{x}$
are independent given $\data$ and
$\xopt$.
This factorization assumption guarantees that the acquisition
function used by PESC is additive across the different functions that are being
evaluated. $\circled{3}$: Let $\xopt^j$ be the $j$-th sample from
$p(\xopt\given\data)$. We then find a Gaussian approximation to each
$p(y_i\given\data,\mathbf{x},\xopt^j)$ using %an implementation of the method
expectation propagation (EP) \citep{Minka:EP}. Let $\sigma_i^2(\x|\xopt^j)$
be the variance of the Gaussian approximation to
$p(y_i\given\data,\mathbf{x},\xopt^j)$ given by EP. Then, we obtain
\begin{align}
\E_{\xopt\given\data} \left[\text{H}\left[\mathbf{y}\given\data,\mathbf{x},\xopt\right]\right] & \overset{\circled{1}}{\approx}
\frac{1}{M} \sum_{j=1}^M \text{H}\left[\mathbf{y}\given\data,\mathbf{x},\xopt^j\right]
\overset{\circled{2}}{\approx} 
\frac{1}{M} 
\sum_{j=1}^M \left[\sum_{i=1}^{K+1}\text{H}\left[y_i\given\data,\mathbf{x},\xopt^j\right]\right]\nonumber\\
& \overset{\circled{3}}{\approx}
\sum_{i=1}^{K+1} \left\{ \frac{1}{M} \sum_{j=1}^M \frac{1}{2}\log \sigma_i^2(\x|\xopt^j) \right\} + \frac{K+1}{2}\log(2\pi e)\,,
\label{eq:pesc_second_term}
\end{align}
where each of the approximations has been numbered with the corresponding step
from the description above. Note that in step $\circled{3}$ of
\cref{eq:pesc_second_term} we have swapped the sums over $i$ and $j$.

The \acq used by PESC is then given by the difference between
\cref{eq:pesc_first_term} and the approximation shown in the last line of \cref{eq:pesc_second_term}. In particular, we obtain
\begin{align}
\alpha_\text{PESC}(\mathbf{x}) = \sum_{i=1}^{K+1} \tilde{\alpha}_i(\mathbf{x})\,,\label{eq:pesc_acquisition}
\end{align}
where
\begin{equation}
\tilde{\alpha}_i(\mathbf{x}) = \frac{1}{M} \sum_{j=1}^M  \underbrace{ \frac{1}{2}\log \sigma_i^2(\x) -
\frac{1}{2}\log \sigma_i^2(\x|\xopt^j)}_{ \textstyle \tilde{\alpha}_i(\mathbf{x}|\mathbf{x}_\star^j)}\,,
\quad\quad i=1,\ldots,K+1\,.
\label{eq:individual_acq}
\end{equation}
Interestingly, the factorization assumption that we made in step $\circled{2}$ of
\cref{eq:pesc_second_term} has produced an acquisition function in
\cref{eq:pesc_acquisition} that is the sum of $K+1$ function-specific acquisition
functions, given by the $\tilde{\alpha}_i(\mathbf{x})$ in
\cref{eq:individual_acq}.
Each $\tilde{\alpha}_i(\mathbf{x})$
measures how much
information we gain on average by only evaluating the $i$-th black box,
where the first black-box evaluates $f$ and the last one evaluates $c_{K+1}$. 
Furthermore,
$\tilde{\alpha}_i(\mathbf{x})$
is the
empirical average of
$\tilde{\alpha}_i(\mathbf{x}|\mathbf{x}_\star)$
across $M$ samples from $p(\mathbf{x}_\star|\mathcal{D})$. Therefore, we can
interpret each $\tilde{\alpha}_i(\mathbf{x}|\mathbf{x}_\star)$ 
in \cref{eq:individual_acq}
as a function-specific acquisition function conditioned on
$\mathbf{x}_\star$. 
Crucially, by using bits of information about the minimizer as a common unit of measurement, our \acq can make meaningful comparisons between the usefulness of evaluating the objective and constraints.
%Each
%$\tilde{\alpha}_i(\mathbf{x}_i)$ is obtained by marginalizing 
%$\tilde{\alpha}_i(\mathbf{x}_i|\mathbf{x}_\star)$ across samples from
%$p(\mathbf{x}_\star|\mathcal{D})$.

We now show how PESC can be used to obtain the task-specific acquisition
functions required by the general algorithm from
\cref{section:decoupling:pesc-decoupled}. Let us assume that we plan to evaluate only
a subset of the functions $f$, $c_1,\ldots,c_K$ and let $t \subseteq
\{1,\ldots,K+1\}$ contain the indices of the functions to be evaluated, where
the first function is $f$ and the last one is
$c_{K}$. We assume that the functions indexed by $t$ are coupled and require
joint evaluation. In this case $t$ encodes a
\emph{task} according to the definition from
\cref{section:decoupled:formalization}. We can then approximate the expected
gain of information that is obtained by evaluating this task at
input $\mathbf{x}$. The process is similar to the one used
above when all the black-boxes are evaluated
at the same time.
However, instead of working with the full vector $\mathbf{y}$,
we now work with the components of $\mathbf{y}$ indexed by $t$.
One can then show that the expected
information gain obtained after evaluating task $t$ at input $\mathbf{x}$ can be approximated as
\begin{align}
\alpha_t(\mathbf{x}) = \sum_{i\in t} \tilde{\alpha}_i(\mathbf{x})\,,\label{eq:task_specific_acq_function}
\end{align}
where the $\tilde{\alpha}_i$ are given by \cref{eq:individual_acq}.  PESC's
\acq is therefore separable since \cref{eq:task_specific_acq_function} can be
used to obtain an acquisition function for each possible task.  The process for
constructing these task-specific acquisition functions is also efficient since
it requires only to use the individual acquisition functions from
\cref{eq:individual_acq} as building blocks. These two properties make PESC an
effective solution for the practical implementation of the general algorithm from
\cref{section:decoupling:pesc-decoupled}.

\begin{figure}[t!]
\centering
\includegraphics[width=0.99\columnwidth]{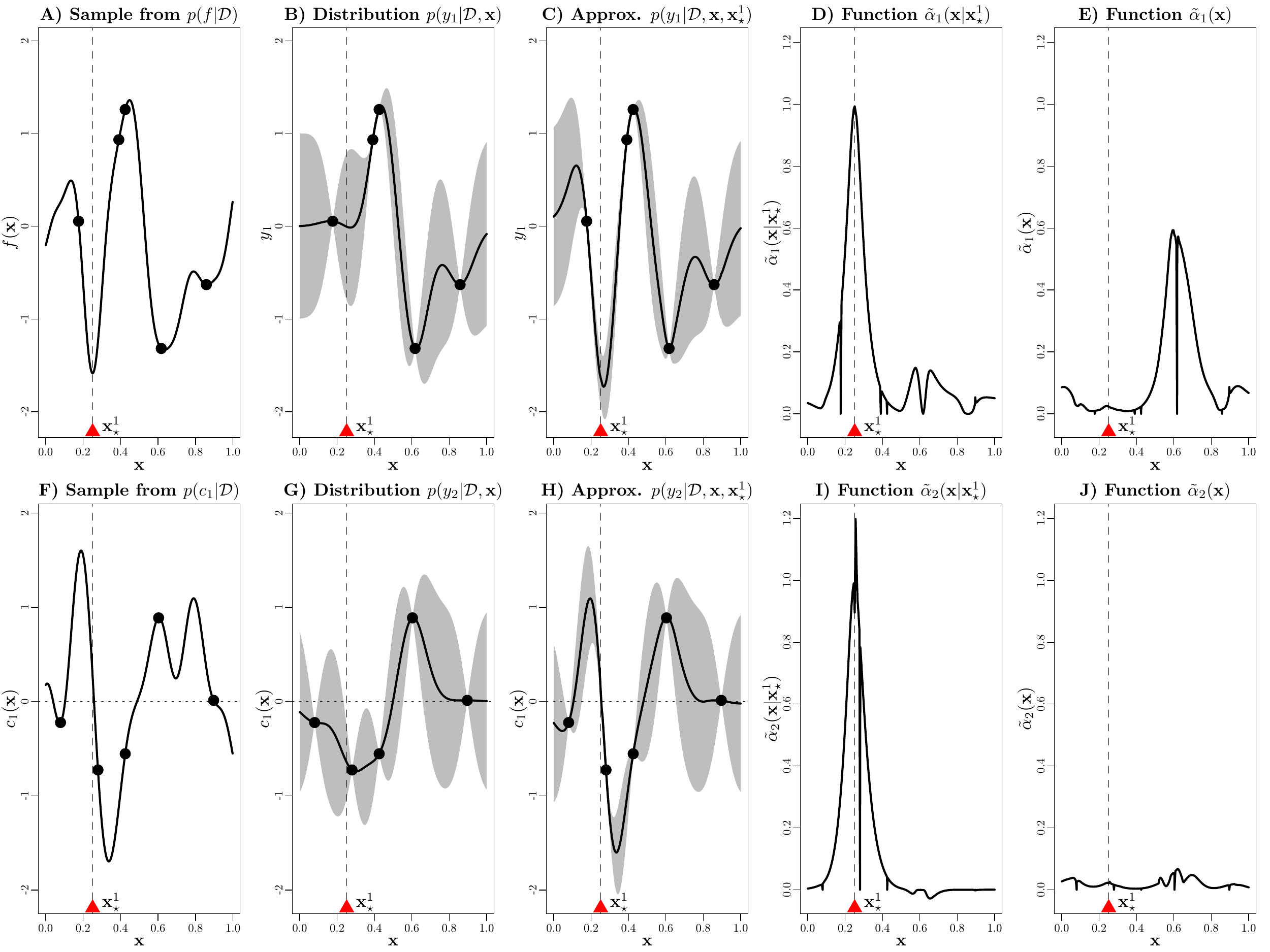}
\caption{Illustration of the process followed to compute the function-specific
acquisition functions given by \cref{eq:individual_acq}. See the main text for
details.}\label{fig:pesc:visualizing_approximation}
\end{figure}

\cref{fig:pesc:visualizing_approximation} illustrates with a toy example the
process for computing the function-specific acquisition functions from
\cref{eq:individual_acq}. In this example there is only one constraint
function. Therefore, the functions in the optimization problem are only $f$ and
$c_1$.  The search space $\mathcal{X}$ is the unit interval $[0,1]$ and we have
collected four measurements for each function. The data for $f$ are shown as
black points in panels A, B and C. The data for $c_1$ are shown as black points
in panels F, G and H. We assume that $f$ and $c_1$ are independently sampled
from a GP with zero mean function and squared exponential covariance function
with unit amplitude and length-scale $0.07$. The noise variance for the
black-boxes that evaluate $f$ and $c_1$ is zero.  Let $y_1$ and $y_2$ be the
black-box evaluations for $f$ and $c_1$ at input $\mathbf{x}$.
Under the assumed GP model we can
analytically compute the predictive distributions for $y_1$ and $y_2$, that is,
$p(y_1|\mathcal{D},\mathbf{x})$ and $p(y_2|\mathcal{D},\mathbf{x})$.
Panels B and G show the means of these distributions with confidence
bands equal to one standard deviation.  The first step to compute the
$\tilde{\alpha}_i(\mathbf{x})$ from \cref{eq:individual_acq} is to draw $M$ samples from
$p(\xopt|\mathcal{D})$. To generate each of these samples, we first
approximately sample $f$ and $c_1$ from their posterior distributions
$p(f|\mathcal{D})$ and $p(c_1|\mathcal{D})$ using the method described by
\citet[][Appendix~A]{hernandez2014}.
Panels A and F show one of the samples obtained for $f$
and $c_1$, respectively. We then solve the optimization problem given by
\cref{eq:problem} when $f$ and $c_1$ are known and equal to the samples
obtained. The solution to this problem is the input that minimizes $f$ subject
to $c_1$ being positive.  This produces a sample $\xopt^1$ from
$p(\xopt|\mathcal{D})$ which is shown as a discontinuous vertical line with a
red triangle in all the panels. The next step is to find a Gaussian approximation to the
predictive distributions when we condition to $\xopt^1$, that is,
$p(y_1|\mathcal{D},\mathbf{x},\xopt^1)$ and
$p(y_2|\mathcal{D},\mathbf{x},\xopt^1)$.  This step is performed using %the method 
expectation propagation (EP) as described in \cref{sec:ep_approximation_light} and Appendix \ref{appendix:implementation}. Panels C
and H show the approximations produced by EP for
$p(y_1|\mathcal{D},\mathbf{x},\xopt^1)$ and
$p(y_2|\mathcal{D},\mathbf{x},\xopt^1)$, respectively. 
Panel C shows that conditioning to $\xopt^1$ decreases the posterior
mean of $y_1$ in the neighborhood of $\xopt^1$. The reason for this is that
$\xopt^1$ must be the global feasible solution and this means that
$f(\xopt^1)$ must be lower than any other feasible point.
Panel H shows that conditioning to $\xopt^1$ increases the posterior
mean of $y_2$ in the neighborhood of $\xopt^1$. The reason for this is that
$c_1(\xopt^1)$ must be positive because $\xopt^1$ has to be feasible. In particular, by conditioning 
to $\xopt^1$ we are giving zero probability to all $c_1$ such that $c_1(\xopt^1) < 0$.
Let $\sigma_1^2(\mathbf{x}|\xopt^1)$ and $\sigma_2^2(\mathbf{x}|\xopt^1)$
be the variances of the Gaussian approximations to
$p(y_1|\mathcal{D},\mathbf{x},\xopt^1)$ and
$p(y_2|\mathcal{D},\mathbf{x},\xopt^1)$ and let
$\sigma_1^2(\mathbf{x})$ and $\sigma_2^2(\mathbf{x})$
be the variances of $p(y_1|\mathcal{D},\mathbf{x})$ and
$p(y_2|\mathcal{D},\mathbf{x})$.  We use these quantities to obtain
$\tilde{\alpha}_1(\mathbf{x}|\xopt^1)$ and
$\tilde{\alpha}_2(\mathbf{x}|\xopt^1)$ according to \cref{eq:individual_acq}.
These two functions are shown in panels D and I.  The whole process is repeated
$M=50$ times and the resulting $\tilde{\alpha}_1(\mathbf{x}|\xopt^j)$ and
$\tilde{\alpha}_2(\mathbf{x}|\xopt^j)$, $j=1,\ldots,M$, are averaged
according to \cref{eq:individual_acq} to obtain the
function-specific acquisition functions $\tilde{\alpha}_1(\mathbf{x})$ and
$\tilde{\alpha}_2(\mathbf{x})$, whose plots are shown in panels E and J.
These plots indicate that evaluating the objective $f$ is in this case 
more informative than evaluating the constraint $c_1$. But this is certainly not always the case, as will be demonstrated in the experiments later on.

\subsection{How to Compute the Gaussian Approximation to $p(y_i|\mathcal{D},\mathbf{x},\xopt^j)$}\label{sec:ep_approximation_light}

We briefly describe the process followed to find a Gaussian approximation to
$p(y_i|\mathcal{D},\mathbf{x},\xopt^j)$ using expectation
propagation (EP) \citep{minka2001family}.  Recall that the variance
of this approximation, that is, $\sigma_i^2(\x|\xopt^j)$, is used to compute
$\tilde{\alpha}_i(\mathbf{x}|\xopt^j)$ in \cref{eq:individual_acq}.  Here we only
provide a sketch of the process; full details can be found in
Appendix \ref{appendix:implementation}.

We start by assuming that the search space has finite size, that is,
$\mathcal{X}=\{\tilde{\mathbf{x}}_1,\ldots,\tilde{\mathbf{x}}_{|\mathcal{X}|}\}$.
In this case the functions $f$, $c_1,\ldots,c_K$ are encoded as finite
dimensional vectors denoted by $\mathbf{f}$,
$\mathbf{c}_1,\ldots,\mathbf{c}_K$. The $i$-th entries in these vectors are the
result of evaluating $f$, $c_1,\ldots,c_K$ at the $i$-th element of
$\mathcal{X}$, that is, $f(\tilde{\mathbf{x}})$,
$c_1(\tilde{\mathbf{x}}_i),\ldots,c_K(\tilde{\mathbf{x}}_i)$.  Let us assume
that $\xopt^j$ and $\mathbf{x}$ are in~$\mathcal{X}$.  Then
$p(\mathbf{y}|\mathcal{D},\mathbf{x},\xopt^j)$ can be defined by the following
rejection sampling process. First, we sample $\mathbf{f}$,
$\mathbf{c}_1,\ldots,\mathbf{c}_K$ from their posterior distribution given the
assumed GP models.  We then solve the optimization problem given by
\cref{eq:problem}. For this, we find the entry of $\mathbf{f}$ with lowest
value subject to the corresponding entries of
$\mathbf{c}_1,\ldots,\mathbf{c}_K$ being positive.  Let
$i\in\{1,\ldots,|\mathcal{X}|\}$ be the index of the selected entry.  Then, if
$\xopt^j \neq \tilde{\mathbf{x}}_i$, we reject the sampled
$\mathbf{f}$, $\mathbf{c}_1,\ldots,\mathbf{c}_K$ and start again. Otherwise, we
take the entries of $\mathbf{f}$, $\mathbf{c}_1,\ldots,\mathbf{c}_K$ indexed by
$\mathbf{x}$, that is, $f(\mathbf{x})$,
$c_1(\mathbf{x}),\ldots,c_K(\mathbf{x})$ and then obtain $\mathbf{y}$
by adding to each of these values a Gaussian random variable with zero mean and 
variance $\nu_1,\ldots,\nu_{K+1}$, respectively.
The probability distribution implied by this rejection sampling process can be
obtained by first multiplying the posterior for $\mathbf{f}$,
$\mathbf{c}_1,\ldots,\mathbf{c}_K$ with indicator functions that take value zero
when $\mathbf{f}$, $\mathbf{c}_1,\ldots,\mathbf{c}_K$ should be rejected and
one otherwise. We can then multiply the resulting quantity by the likelihood for $\mathbf{y}$
given $\mathbf{f}$, $\mathbf{c}_1,\ldots,\mathbf{c}_K$.
The desired distribution is finally obtained by marginalizing out $\mathbf{f}$, $\mathbf{c}_1,\ldots,\mathbf{c}_K$.

We introduce several indicator functions to implement the approach described above. The first
one $\Gamma(\mathbf{x})$ takes value one when $\mathbf{x}$ is a feasible solution and value zero otherwise,
that is,
\begin{align}
\Gamma(\mathbf{x}) = \prod_{k=1}^K \Theta [c_k(\mathbf{x})]\,,\label{eq:indicator_gamma}
\end{align}
where $\Theta[\cdot]$ is the Heaviside step function which is
equal to one if its input is non-negative and zero otherwise. 
The second indicator function $\Psi(\mathbf{x})$ takes value
zero if $\mathbf{x}$ is a better solution than $\xopt^j$ according to the
sampled functions. Otherwise $\Psi(\mathbf{x})$ takes value one. In particular,
\begin{align}
\Psi(\mathbf{x}) = \Gamma(\mathbf{x}) \Theta [ f(\mathbf{x}) - f(\xopt^j) ] + 
\left(1 - \Gamma(\mathbf{x})\right)\,.\label{eq:indicator_psi}
\end{align}
When $\mathbf{x}$ is infeasible, this expression takes value one. In this case,
$\mathbf{x}$ is not a better solution than $\xopt^j$ (because $\mathbf{x}$ is
infeasible) and we do not have to reject. When $\mathbf{x}$ is feasible,
the factor $\Theta [ f(\mathbf{x}) - f(\xopt^j) ]$ in \cref{eq:indicator_psi} is zero when $\mathbf{x}$
takes lower objective value than $\xopt^j$.
This will allow us to reject $\mathbf{f}$, $\mathbf{c}_1,\ldots,\mathbf{c}_K$
when $\mathbf{x}$ is a better solution than $\xopt^j$.
Using \cref{eq:indicator_gamma} and \cref{eq:indicator_psi}, we can then
write $p(\mathbf{y}|\mathcal{D},\mathbf{x},\xopt^j)$ as
\begin{equation}
\resizebox{.9\hsize}{!}{$
p(\mathbf{y}|\mathcal{D},\mathbf{x},\xopt^j) \propto \int 
p(\mathbf{y}|\mathbf{f},\mathbf{c}_1,\ldots,\mathbf{c}_K,\mathbf{x})
\underbrace{p(\mathbf{f},\mathbf{c}_1,\ldots,\mathbf{c}_K|\mathcal{D}) \Gamma(\xopt^j)
\left\{ \prod_{\mathbf{x}'\in \mathcal{X}} \Psi(\mathbf{x}') \right\}}_{\textstyle f(\mathbf{f},\mathbf{c}_1,\ldots,\mathbf{c}_K|\xopt^j)}
\,d\mathbf{f}\,d\mathbf{c}_1\cdots\mathbf{c}_K$}\,,\label{eq:conditioned_predictive_distribution}
\end{equation}
where $p(\mathbf{f},\mathbf{c}_1,\ldots,\mathbf{c}_K|\mathcal{D})$ is the GP
posterior distribution for the noise-free evaluations of $f$, $c_1,\ldots,c_K$
at $\mathcal{X}$ and
$p(\mathbf{y}|\mathbf{f},\mathbf{c}_1,\ldots,\mathbf{c}_K,\mathbf{x})$ is the likelihood function, that is, the
distribution of the noisy evaluations produced by the black-boxes with input
$\mathbf{x}$ given the true function values:
\begin{equation}
\resizebox{.9\hsize}{!}{$
p(\mathbf{y}|\mathbf{f},\mathbf{c}_1,\ldots,\mathbf{c}_K,\mathbf{x}) = 
\mathcal{N}(y_1|f(\mathbf{x}), \nu_1)
\mathcal{N}(y_2|c_1(\mathbf{x}), \nu_2)
\cdots
\mathcal{N}(y_{K+1}|c_K(\mathbf{x}), \nu_{K+1})$}\,.\label{eq:black_box_noise}
\end{equation}
The product of the indicator functions $\Gamma$ and $\Psi$ in
\cref{eq:conditioned_predictive_distribution} takes value zero whenever
$\xopt^j$ is not the best feasible solution according to
$\mathbf{f},\mathbf{c}_1,\ldots,\mathbf{c}_K$.  The indicator $\Gamma$ in
\cref{eq:conditioned_predictive_distribution} guarantees that $\xopt^j$ is a
feasible location. The product of all the $\Psi$ in
\cref{eq:conditioned_predictive_distribution} guarantees that no other point in
$\mathcal{X}$ is better than $\xopt^j$. Therefore, the product of $\Gamma$ and
the $\Psi$ in \cref{eq:conditioned_predictive_distribution} rejects any value
of $\mathbf{f},\mathbf{c}_1,\ldots,\mathbf{c}_K$ for which $\xopt^j$ is not the
optimal solution to the constrained optimization problem.

The factors $p(\mathbf{f},\mathbf{c}_1,\ldots,\mathbf{c}_K|\mathcal{D})$ and
$p(\mathbf{y}|\mathbf{f},\mathbf{c}_1,\ldots,\mathbf{c}_K,\mathbf{x})$ in
\cref{eq:conditioned_predictive_distribution} are Gaussian. Thus, their product
is also Gaussian and tractable. However, the integral in
\cref{eq:conditioned_predictive_distribution} does not have a closed form
solution because of the complexity introduced by the the product of indicator
functions $\Gamma$ and $\Psi$. This means that
\cref{eq:conditioned_predictive_distribution} cannot be exactly computed and has to be
approximated. For this, we use EP to
fit a Gaussian approximation to the product of
$p(\mathbf{f},\mathbf{c}_1,\ldots,\mathbf{c}_K|\mathcal{D})$ and the indicator
functions $\Gamma$ and $\Psi$ in \cref{eq:conditioned_predictive_distribution},
which we have denoted by
$f(\mathbf{f},\mathbf{c}_1,\ldots,\mathbf{c}_K|\xopt^j)$,
with a tractable Gaussian distribution given by
\begin{align}
q(\mathbf{f},\mathbf{c}_1,\ldots,\mathbf{c}_K|\xopt^j) = \mathcal{N}(\mathbf{f}|\mathbf{m}_1,\mathbf{V}_1)
\mathcal{N}(\mathbf{c}_1|\mathbf{m}_2,\mathbf{V}_2)
\cdots
\mathcal{N}(\mathbf{c}_K|\mathbf{m}_{K+1},\mathbf{V}_{K+1})\,,\label{eq:ep_approximation_q}
\end{align}
where $\mathbf{m}_1,\ldots,\mathbf{m}_{K+1}$ and
$\mathbf{V}_1,\ldots,\mathbf{V}_{K+1}$ are mean vectors and covariance matrices
to be determined by the execution of EP. Let $v_i(\mathbf{x})$ be the
diagonal entry of
$\mathbf{V}_i$ corresponding to the evaluation location given by
$\mathbf{x}$, where $i=1,\ldots,K+1$. Similarly,
let $m_i(\mathbf{x})$ be the entry of $\mathbf{m}_i$ corresponding to the
evaluation location $\mathbf{x}$ for $i=1,\ldots,K+1$.  Then, by replacing 
$f(\mathbf{f},\mathbf{c}_1,\ldots,\mathbf{c}_K|\xopt^j)$ in \cref{eq:conditioned_predictive_distribution}
with $q(\mathbf{f},\mathbf{c}_1,\ldots,\mathbf{c}_K|\xopt^j)$, we obtain
\begin{equation}
p(\mathbf{y}|\mathcal{D},\mathbf{x},\xopt^j) \approx \prod_{i=1}^{K+1} \mathcal{N}(y_i|m_i(\mathbf{x}),v_i(\mathbf{x}) + \nu_i)\,.
\label{eq:ep_approximation}
\end{equation}
Consequently, $\sigma_i^2(\x|\xopt^j)=v_i(\mathbf{x})+ \nu_i$ can be
used to compute $\tilde{\alpha}_i(\mathbf{x}|\xopt^j)$ in \cref{eq:individual_acq}.

The previous approach does not work when the search space $\mathcal{X}$ has
infinite size, for example when $\mathcal{X}=[0,1]^d$ with $d$ being the
dimension of the inputs to $f,c_1,\ldots,c_K$.  In this case the product of
indicators in \cref{eq:conditioned_predictive_distribution} includes an
infinite number of factors $\Psi(\mathbf{x}')$, one for each possible
$\mathbf{x}'\in \mathcal{X}$. To solve this problem we perform an additional
approximation. For the computation of
\cref{eq:conditioned_predictive_distribution}, we consider that $\mathcal{X}$
is well approximated by the finite set $\mathcal{Z}$, which contains only
the locations at which the objective $f$ has been evaluated so far,
the value of $\xopt^j$ and 
$\mathbf{x}$. Therefore, we approximate the factor $\prod_{\mathbf{x}'\in
\mathcal{X}} \Psi(\mathbf{x}')$ in
\cref{eq:conditioned_predictive_distribution} with the factor
$\prod_{\mathbf{x}'\in \mathcal{Z}} \Psi(\mathbf{x}')$, which has now finite size.
We expect this approximation to become more and more accurate as we increase
the amount of data collected for $f$. Note that our approximation to
$\mathcal{X}$ is finite, but it is also different for each location
$\mathbf{x}$ at which we want to evaluate
\cref{eq:conditioned_predictive_distribution} since $\mathcal{Z}$ is defined to
contain $\mathbf{x}$.  A detailed description of the resulting EP
algorithm, indicating how to compute the variance functions $v_i(\mathbf{x})$
shown in \cref{eq:ep_approximation}, is given in Appendix \ref{appendix:implementation}. 

The EP approximation to \cref{eq:ep_approximation}, performed after replacing
$\mathcal{X}$ with $\mathcal{Z}$, depends on the values of $\mathcal{D}$,
$\xopt^j$ and $\mathbf{x}$. Having to re-run EP for each value of $\mathbf{x}$
at which we may want to evaluate the acquisition function given by
\cref{eq:pesc_acquisition} is a very expensive operation.
To avoid this, we split the EP computations between those that depend only on
$\mathcal{D}$ and $\xopt^j$, which are the most expensive ones, and those that
depend only on the value of $\mathbf{x}$. We perform the former computations only once
and then reuse them for each different value of $\mathbf{x}$. This allows us to evaluate
the EP approximation to \cref{eq:conditioned_predictive_distribution} at
different values of $\mathbf{x}$ in a computationally efficient way.
See Appendix \ref{appendix:implementation} for further details.

\subsection{Efficient Marginalization of the Model Hyper-parameters}\label{sec:hyper_marginalization}

So far we have assumed to know the optimal hyper-parameter values, that is, the
amplitude and the length-scales for the GPs and the noise variances for the
black-boxes. However, in practice, the hyper-parameter values are unknown and have to be
estimated from data. This can be done for example by drawing samples from the
posterior distribution of the hyper-parameters under some non-informative
prior. Ideally, we should then average the GP predictive distributions with
respect to the generated samples before approximating the
information gain. However, this approach is too computationally expensive in practice.
Instead, we follow \cite{snoek-etal-2012b} and average the PESC acquisition
function with respect to the generated hyper-parameter samples.
In our case, this involves marginalizing each of the function-specific
acquisition functions from \cref{eq:individual_acq}. For this, we follow the
method proposed by \cite{hernandez2014} to average the acquisition function
of Predictive Entropy Search in the unconstrained case. Let $\bm \Theta$ denote
the model hyper-parameters. First, we draw $M$ samples $\bm \Theta^1,\ldots,\bm
\Theta^M$ from the posterior distribution of $\bm \Theta$ given the data
$\mathcal{D}$. %In general, we will draw the same number of samples of $\bm \Theta$ and $\xopt$.
%In particular, 
Typically, for each of the posterior samples $\bm \Theta^j$ of $\bf \Theta$ we draw a single
corresponding sample $\xopt^j$ from the posterior distribution of $\xopt$ %when the
%hyper-parameters are fixed to 
given $\bm \Theta^j$, that is, $p(\xopt\given\data,\bm
\Theta^j)$. Let $\sigma_i^2(\x|\bm \Theta^j)$ be the variance of the GP
predictive distribution for $y_i$ when the hyper-parameter values are fixed to
$\bm \Theta^j$, that is, $p(y_i|\mathcal{D},\mathbf{x},\bm \Theta^j)$,
and let $\sigma_i^2(\x|\xopt^j,\bm \Theta^j)$  be the variance of the
Gaussian approximation to the predictive distribution for $y_i$ when we condition to the
solution of the optimization problem being $\xopt^j$ and the hyper-parameter
values being $\bm \Theta^j$. Then, the version of \cref{eq:individual_acq} that marginalizes
out the model hyper-parameters is given by
\begin{align}
\tilde{\alpha}_i(\mathbf{x}) = \frac{1}{M} \sum_{j=1}^M \left\{ \frac{1}{2}\log \sigma_i^2(\x|\bm \Theta^j) -
\frac{1}{2}\log \sigma_i^2(\x|\xopt^j,\bm \Theta^j) \right\}\,,
\quad\quad i=1,\ldots,K+1\,.
\label{eq:marginalzed_individual_acq}
\end{align}
Note that $j$ is now an index over joint posterior samples of the model hyper-parameters
$\bm \Theta$ and the constrained minimizer $\xopt$. Therefore, we can
marginalize out the hyper-parameter values without adding any additional
computational complexity to our method because a loop over $M$ samples of $\xopt$ is just replaced with a loop over $M$ joint samples of $(\Theta,\xopt)$. This is a consequence of our
reformulation of \cref{eq:original_acquisition_pesc} into \cref{eq:new_acquisition_pesc}. 
By contrast, other techniques that work by
approximating the original form of the acquisition function used in \cref{eq:original_acquisition_pesc}
do not have this property.  An example in
the unconstrained setting is Entropy Search
\citep{hennig-schuler-2012}, which requires re-computing an approximation to
the acquisition function for each hyper-parameter sample $\bm \Theta^j$.

\subsection{Computational Complexity}

In the coupled setting, the complexity of PESC is $\bigO(MKN^3)$, where $M$ is
the number of posterior samples of the global constrained minimizer $\xopt$, $K$ is the number of constraints, and $N$ is
the number of collected data points. This cost is determined by the cost of each EP iteration,
which requires computing the inverse of the covariance matrices
$\mathbf{V}_1,\ldots,\mathbf{V}_{K+1}$ in \cref{eq:ep_approximation}. The dimensionality of each of these
matrices grows with the size of
$\mathcal{Z}$, which is determined by the number $N$ of objective evaluations 
(see the last paragraph of \cref{sec:ep_approximation_light}).
Therefore each EP iteration has cost $\mathcal{O}(KN^3)$ and we have to run an
instance of EP for each of the $M$ samples of $\xopt$. If $M$ is also the
number of posterior samples for the GP hyperparameters, as explained in
\cref{sec:hyper_marginalization}, this is the same computational complexity as
in EIC. However, in practice PESC is slower than EIC because of the cost
of running multiple iterations of the EP algorithm.

In the decoupled setting the cost of PESC is ${\bigO(M\sum_{k=2}^{K+1}
(N_1+N_k)^3)}$ where $N_1$ is the number of evaluations of the objective and
$N_k$ is the number of evaluations for constraint $k-1$. The origin of this
cost is again the size of the matrices $\mathbf{V}_1,\ldots,\mathbf{V}_{K+1}$
in \cref{eq:ep_approximation}. While $\mathbf{V}_1$ still scales as a function
of $|\mathcal{Z}|$, we have that $\mathbf{V}_2,\ldots,\mathbf{V}_{K+1}$ scale
now as a function of $|\mathcal{Z}|$ plus the number of observations for the
corresponding constraint function.  The reason for this is that
$\prod_{\mathbf{x}'\in \mathcal{Z}} \Psi(\mathbf{x}')$ is used to approximate
$\prod_{\mathbf{x}'\in \mathcal{X}} \Psi(\mathbf{x}')$ in
\cref{eq:conditioned_predictive_distribution} and each factor in
$\prod_{\mathbf{x}'\in \mathcal{Z}} \Psi(\mathbf{x}')$ represents then
a virtual data point for each GP.
See Appendix \ref{appendix:implementation} for details.

The cost of sampling the GP hyper-parameters is $\bigO(MKN^3)$ and
therefore, it does not affect the overall computational complexity of PESC.

\subsection{Relationship between PESC and PES}\label{section:pesc:discussion:relationship-to-PES}

PESC can be applied to unconstrained optimization problems. For this we only
have to set $K=0$ and ignore the constraints.  The resulting technique is very
similar to the method PES proposed by \cite{hernandez2014} as an
information-based approach for unconstrained Bayesian optimization. However,
PESC without constraints and PES are not identical. PES approximates
$p(y|\mathcal{D},\mathbf{x},\xopt^j)$ by multiplying the GP predictive
distribution by additional factors that enforce $\xopt^j$ to be the location
with lowest objective value. These factors guarantee that 1) the value of the
objective at $\xopt^j$ is lower than the minimum of the values for the objective
collected so far, 2) the gradient of the objective is zero at $\xopt$ and 3)
the Hessian of the objective is positive definite at $\xopt$.  We do not
enforce the last two conditions since the global optimum may be on the boundary
of a feasible region and thus conditions 2) and 3) do not necessarily
hold (this issue also arises in PES because the optimum may be on the
boundary of the search space $\domain$). Condition 1) is implemented in
PES by taking the minimum observed value for the objective, denoted by $\eta$,
and then imposing the soft condition $f(\xopt^j) < \eta + \epsilon$, where
$\epsilon \sim \mathcal{N}(0,\nu)$ accounts for the additive Gaussian noise 
with variance $\nu$ in the black-box that evaluates the objective.
In PESC this is achieved in a more principled way by using
the indicator functions given by \cref{eq:indicator_psi}.

\subsection{Summary of the Approximations Made in PESC} \label{section:pesc:summary-of-approximations}

We describe here all the approximations performed in the practical implementation of PESC.
PESC approximates the expected reduction in the posterior entropy of $\xopt$ (see Eq. \ref{eq:original_acquisition_pesc}) with
the acquisition function given by \cref{eq:pesc_acquisition}. This involves the following approximations:
\begin{enumerate}
\item The expectation over $\xopt$ in \cref{eq:new_acquisition_pesc} is approximated with Monte Carlo sampling.

\item The Monte Carlo samples of $\xopt$ come from samples of $f,c_1,\ldots,c_K$ drawn approximately using a finite
basis function approximation to the GP covariance function, as described by
\citet[][Appendix~A]{hernandez2014}.

\item We approximate the factor $\prod_{\mathbf{x}'\in \mathcal{X}}
\Psi(\mathbf{x}')$ in \cref{eq:conditioned_predictive_distribution} with the
factor $\prod_{\mathbf{x}'\in \mathcal{Z}} \Psi(\mathbf{x}')$. Unlike the
original search space $\mathcal{X}$, $\mathcal{Z}$ has now finite size and the
corresponding product of $\Psi$ indicators is easier to approximate. The set
$\mathcal{Z}$ is formed by the locations of the current observations for the
objective $f$ and the current evaluation location $\mathbf{x}$ of the
acquisition function.

\item After replacing $\prod_{\mathbf{x}'\in \mathcal{X}} \Psi(\mathbf{x}')$
with $\prod_{\mathbf{x}'\in \mathcal{Z}} \Psi(\mathbf{x}')$ in
\cref{eq:conditioned_predictive_distribution}, we further approximate the
factor $f(\mathbf{f},\mathbf{c}_1,\ldots,\mathbf{c}_K|\xopt^j)$ in this
equation with the Gaussian approximation given by the right-hand-side of
\cref{eq:ep_approximation}. We use the method expectation propagation (EP) for
this task, as described in Appendix \ref{appendix:implementation}.  Because the EP approximation in
\cref{eq:ep_approximation_q} factorizes across $\mathbf{f}$,
$\mathbf{c}_1,\ldots,\mathbf{c}_K$, the execution of EP implicitly includes
the factorization assumption performed in step $\circled{2}$ of
\cref{eq:pesc_second_term}.

\item As described in the last paragraph of \cref{sec:ep_approximation_light},
in the execution of EP we separate the computations that depend on
$\mathcal{D}$ and $\xopt^j$, which are very expensive, from those that depend
on the location $\mathbf{x}$ at which the PESC acquisition function will be
evaluated. This allows us to evaluate the approximation to
\cref{eq:conditioned_predictive_distribution} at different values of
$\mathbf{x}$ in a computationally efficient way.

\item To deal with unknown hyper-parameter values, we marginalize the
acquisition function over posterior samples of the hyper-parameters.
Ideally, we should instead marginalize the predictive distributions with respect to the hyper-parameters before
computing the entropy, but this is too computationally expensive in practice. \label{hyper:approximation}

\end{enumerate}
In \cref{section:experiments:pesc-approximation-quality}, we assess the
accuracy of these approximations (except the last one) and show that PESC performs %almost as well as
on par with a ground-truth method based on rejection sampling.

Note that in addition to the mathematical approximations described above, additional sources of error are introduced by the numerical computations involved. In addition to the usual roundoff error, etc., we draw the reader's attention to the fact that the $\xopt$ samples are the result of numerical global optimization of the approximately drawn samples of $f,c_1,\ldots,c_K$, and then the suggestion is chosen by another numerical global optimization of the \acq. At present, we do not have guarantees that the true global optimum is found by our numerical methods in each case.

%!TEX root = main.tex
\section{PESC-F: Speeding Up the BO Computations}
\label{section:pesc-fast-updates}

One disadvantage of PESC is that sampling $\xopt$ and then computing the
corresponding EP approximation can be slow. If PESC is slow with respect to the
evaluation of the black-box functions $f,c_1,\ldots,c_K$, the entire \bo (BO)
procedure may be inefficient. For the BO approach to be useful, the time spent
doing meta-computations has to be significantly shorter than the time spent actually evaluating the objective and constraints. This issue can be avoided in the coupled case by,
for example, switching to a faster \acq like EIC or abandoning BO entirely for methods such as the popular CMA-ES \citep{CMA-ES} evolutionary strategy.
However, in the decoupling setting, one can encounter problems in which some tasks are fast and others are slow. In this case, a cumbersome BO method might be undesirable because it would be unreasonable to spend minutes making a decision about a task that only takes seconds to complete; and, yet, a method that is fast but inefficient in terms of function evaluations would be ill-suited to making decisions about a task that takes hours complete. This situation calls for an optimization algorithm that can adaptively adjust its own decision-making time. %, we can obtain significant gains by using a method that can
%handle decoupling such as PESC and that can also make fast decisions when they
%are required. 
For this reason, we introduce additional approximations in the
computations made by PESC to reduce their cost when necessary. The new method
that adaptively switches between fast and slow decision-making computations is called PESC-F. The two
main challenges are how to speed up the original computations made by PESC and
how to decide when to switch between the slow and the fast versions of those
computations. In the following paragraphs we address these issues.

%\subsection{}
%\label{section:pesc:fast-pesc:how-to-make-it-fast}

\crefname{enumi}{step}{steps}
We propose ways to reduce the cost of the computations performed by PESC after collecting each new data point.
These computations include
\begin{enumerate}
\item \label{enum:pesc:fast:gp} Drawing posterior samples of the GP
hyper-parameters and then for each sample computing the Cholesky decomposition of the kernel matrix.
\item \label{enum:pesc:fast:ep} Drawing approximate posterior samples of $\xopt$ and
then running an EP algorithm for each of these samples.
\item \label{enum:pesc:fast:max} Globally maximizing the resulting acquisition functions.
\end{enumerate}
We shorten each of these steps. First, we reduce the cost of
\cref{enum:pesc:fast:gp} by skipping the sampling of the GP hyper-parameters and
instead considering the hyper-parameter samples already used at an earlier
iteration. This also allows for additional speedups by using fast ($\mathcal{O}(N^2)$) updates of
the Cholesky decomposition of the kernel matrix instead of recomputing it from
scratch.  Second, we shorten \cref{enum:pesc:fast:ep} by skipping the sampling
of $\xopt$ and instead considering the samples used at the previous
iteration.  We also reuse the EP solutions computed at the previous iteration 
(see \cref{appendix:implementation} for further details on how to reuse the EP solutions). Finally, we shorten
\cref{enum:pesc:fast:max} by using a coarser termination condition tolerance
when maximizing the \acq. This allows the optimization process to converge
faster but with reduced precision. Furthermore, if the \acq is maximized using
a local optimizer with random restarts and/or a grid initialization, we can
shorten the computation further by reducing the number of restarts and/or grid
size.

%computed 
%previously computed EP approximation instead of recomputing it. In particular,
%we reuse the approximations $\tilde{h}_n$ and $\tilde{g}_k$ in
%\cref{eq:pesc-supplement:q-in-terms-of-g-and-h} and then recompute the
%approximation to $q(\vectors)$ (\cref{eq:pesc:finite-approx-q}) using
%\cref{eq:pesc-supplement:means-and-variances-in-q-tilde}. This allows us to
%skip all the computations in \cref{section:pesc-supplement:EP-approximation}.
%We then perform the remaining calculations given in \cref{section:pesc} in the
%usual way. Since computing the EP approximation to $q(\vectors)$ is much more
%expensive than the ensuing computations, this caching strategy leads to a
%significant speedup.

\subsection{Choosing When to Run the Fast or the Slow Version}
\label{section:pesc:fast-pesc:choosing-fast-or-slow}

The motivation for PESC-F is that the time spent in the BO computations should
be small compared to the time spent evaluating the black-box functions.
Therefore, our approach is to switch between two distinct types of BO
computations: the full (slow) and the partial (fast) PESC computations. Our
goal is to approximately keep constant the fraction of total \walltime consumed
by such computations. To achieve this, at each iteration of the BO process, we
use the slow version of the computations if and only if 
\begin{align}
\frac{\tau_\text{now}-\tau_\text{last}}{\tau_\text{slow}} > \gamma \,,
\label{eq:pesc:fast-pesc:rationality-condition}
\end{align}
where $\tau_\text{now}$ is the current time, $\tau_\text{last}$ is the time at
which the last slow BO computations were complete, $\tau_\text{slow}$ is
the duration of the last execution of the slow BO computations (this includes
the time passed since the actual collection of the data until the maximization
of the acquisition function) and $\gamma>0$ is a constant called the rationality
level. The larger the value of $\gamma$, the larger the amount of time spent in
rational decision making, that is, in performing BO computations.
\cref{algorithm:pesc:fast-slow} shows the steps taken by PESC-F for the decoupled competitive case.
In this case each function $f,c_1,\ldots,c_K$ represents a different task, that is,
the different functions can be evaluated in a decoupled manner and in addition to this, all of them compete
for using a single computational resource.

One could replace $\tau_\text{slow}$ with an average over the
durations of past slow computations. While this approach is less noisy, we opt
for using only the duration of the most recent slow update since these
durations may exhibit deterministic trends. For example, the cost of computations
tends to increase at each iteration due to the increase in data set size. If
indeed the update duration increases monotonically, then the duration of
the most recent update would be a more accurate estimate of the duration of the
next slow update than the average duration of all past updates.

\begin{algorithm}[tb]
   \caption{PESC-F for competitive decoupled functions.}
   \label{algorithm:pesc:fast-slow}
\begin{algorithmic}[1] % the "1" means put a line number at every line
   \STATE {\bfseries Inputs:} $\mathcal{T} =\{ \{f \}, \{c_1\},\ldots,\{c_K\} \}$, $\data$, $\gamma$, $\delta$.
   \STATE $\tau_\text{last} \leftarrow  0$
   \STATE $\tau_\text{slow} \leftarrow 0$
   \REPEAT
   \STATE $\tau_\text{now}\leftarrow\text{current time}$
   \IF{$(\tau_\text{now}-\tau_\text{last})/\tau_\text{slow} > \gamma$}
   \STATE Sample GP hyper-parameters
   \STATE Fit GP to $\data$
   \STATE Generate new samples of $\xopt$
   \STATE Compute the EP solutions from scratch
   \STATE $\tau_\text{slow} \leftarrow \text{current time} - \tau_\text{now}$
   \STATE $\tau_\text{last} \leftarrow \text{current time}$
   \STATE $\{\x^*, t^* \} \leftarrow \arg \max_{\x\in\domain,t\in\mathcal{T}} \alpha_t(\x)$ (expensive optimization)
   \ELSE
   \STATE Update fit of GP to $\data$
   \STATE Reuse previous EP solutions
   \STATE $\{\x^*, t^* \} \leftarrow \arg \max_{\x\in\domain,t\in\mathcal{T}} \alpha_t(\x)$ (cheap optimization)
   \ENDIF
   \STATE Add to $\mathcal{D}$ the evaluation of the function in task $t^*$ at input $\x^*$
   \UNTIL{termination condition is met}
   \STATE {\bfseries Output:} 
    $\arg \min_{\x\in\domain} \E_\text{GP}[f(\x)] \;\, \mathrm{s.t.} \; 
   p(c_1(\x) \geq 0,\ldots, c_K(\x)\geq0|\text{GP}) \geq 1 - \delta$
\end{algorithmic}
\end{algorithm}

PESC-F can be used as a generalization of PESC, since it reduces to PESC in the
case of sufficiently slow function evaluations. To see this, note that the time
spent in a function evaluation will be upper bounded by $\tau_\text{now} -
\tau_\text{last}$ and according to \cref{eq:pesc:fast-pesc:rationality-condition}, the slow
computations are performed when ${\tau_\text{now} -\tau_\text{last} >
\gamma \tau_\text{slow}}$.  When the function evaluation takes a very large amount of time, we have
that $\tau_\text{slow}$ will always be smaller than that amount of time and the condition
${\tau_\text{now} -\tau_\text{last} > \gamma \tau_\text{slow}}$ will
always be satisfied for reasonable choices of $\gamma$.
Thus, PESC-F will always perform slow computations as we would
expect. On the other hand, if the evaluation of the black-box function is very fast, PESC-F
will mainly perform fast computations but will still occasionally perform slow ones,
with a frequency roughly proportional to the function evaluation duration.

\subsection{Setting the Rationality Level in PESC-F}
\label{section:pesc:fast-pesc:setting-gamma}

PESC-F is designed so that the ratio of time spent in BO computations to time
spent in function evaluations is at most $\gamma$. This notion is approximate
because the time spent in function evaluations includes the time spent doing
fast computations. The optimal value of~$\gamma$ may be problem-dependent, but
we propose values of $\gamma$ on the order of $0.1$ to $1$, which correspond to
spending roughly $50-90\%$ of the total time performing function evaluations.
The optimal $\gamma$ may also change at different stages of the BO process.
Selecting the optimal value of $\gamma$ is a subject for future research. Note that in
PESC-F we are making sub-optimal decisions because of time constraints.
Therefore, PESC-F is a simple example of \emph{bounded rationality}, which
has its roots in the traditional AI literature. For example, \cite{russel1991}
proposes to treat computation as a possible action that consumes time but
increases the expected utility of future actions.

\subsection{Bridging the Gap Between Fast and Slow Computations}
\label{section:pesc:discussion:pesc-fast-slow}

As discussed above, PESC-F can be applied even when function evaluations are very slow, as it automatically reverts to standard PESC when $\tau_\text{eval}>\tau_\text{slow}$. However, if the function evaluations are extremely fast, that is, faster even that the fast PESC updates, then even PESC-F violates the condition that the decision-making should take less time than the function evaluations. 
% PESC-F can switch between fast, yet inaccurate BO computations and accurate,
% but slow ones. The objective of PESC-F is to keep the ratio of time spent in BO
% computations to time spent in function evaluations below a user-specified
% value. However, if the function evaluations are so fast that the time spent in
% the fast updates becomes significant or even the dominant factor, then PESC-F
% will fail to achieve the desired ratio of computation to meta-computation.
% Of course, when the function evaluations
% are very fast then BO may not be a good approach at all.  
We have already
defined $\tau_\text{slow}$ as the duration of the slow BO computations. Let us
also define $\tau_\text{fast}$ as the duration of the fast BO computations and
$\tau_\text{eval}$ as the duration of the evaluation of the functions.
Then, the intuition described above can be put into symbols by saying that PESC-F is most useful
when~${\tau_\text{fast}<\tau_\text{eval}<\tau_\text{slow}}$.
% When~${\tau_\text{eval}>\tau_\text{slow}}$, PESC-F is not necessary and the
% original PESC method can be used.  However, PESC-F also works well in this
% setting because it automatically reverts to standard PESC with slow BO
% computations when~${\tau_\text{eval}>\tau_\text{slow}}$. 

Many aspects of PESC-F are not specific to PESC and could easily be
adapted to other \acqs like EIC or even unconstrained \acqs like PES and EI.
In particular, lines 9, 10 and 16 of \cref{algorithm:pesc:fast-slow} are specific
to PESC, whereas others are common to other techniques. For example, when using
vanilla unconstrained EI, the computational bottleneck is likely to be the sampling
of the GP hyper-parameters (\cref{algorithm:pesc:fast-slow}, line 7) and maximizing
the \acq (\cref{algorithm:pesc:fast-slow}, line 13). The ideas presented above,
namely to skip the hyper-parameter sampling and to optimize the \acq with a
smaller grid and/or coarser tolerances, are applicable in this situation and
might be useful in the case of a fairly fast objective function. However, as mentioned above, in the single-task case one retains the option to abandon BO entirely for a faster method, whereas in the multi-task case considered here, neither a purely slow nor a purely fast method suits the nature of the optimization problem. An interesting direction for future research is to further pursue this notion of optimization algorithms that bridge the gap between those designed for optimizing cheap (fast) functions and those designed for optimizing expensive (slow) functions.

% When the function evaluations are very fast, that is, when $\tau_\text{eval} <
% \tau_\text{fast}$, model-based methods such as PESC-F are unlikely to perform
% well. Instead, alternative techniques such as evolutionary strategies, for
% example CMA-ES \citep{CMA-ES}, are likely to be better because they can perform
% many more function evaluations in a fixed amount of time. However, the user may
% not know the values of the constants $\tau_\text{eval}$ and $\tau_\text{fast}$ beforehand
% and thus may not know which method to use. Furthermore, in a general scenario
% with multiple decoupled tasks, the user might be faced with some very slow
% tasks and some very fast tasks for which $\tau_\text{eval} < \tau_\text{fast}$.
% In this case, neither a BO method such as PESC-F nor a fast approach like
% CMA-ES are well-suited to the entire optimization problem. 
% Therefore, an interesting direction for future research is to create algorithms that can
% bridge the gap between very fast and slower function evaluations.

%!TEX root = main.tex
\section{Empirical Analyses in the Coupled Case}
\label{section:experiments}

We first evaluate the performance of PESC in experiments with different types of
coupled optimization problems. First, we consider synthetic problems of functions sampled from the GP prior distribution. Second, we consider analytic
benchmark problems that were previously used in the literature on Bayesian optimization
with unknown constraints. Finally, we address the meta-optimization of machine
learning algorithms with unknown constraints. 

For the first synthetic case, we follow the experimental setup used by
\citet{hennig-schuler-2012} and \cite{hernandez2014}. The search space is the unit
hypercube of dimension $\dimension$, and the ground truth objective~$f$ is a
sample from a zero-mean GP with a squared exponential covariance function of
unit amplitude and length scale~${\ell = 0.1}$ in each dimension. We represent
the function~$f$ by first sampling from the GP prior on a grid of 1000 points
generated using a Halton sequence \citep[see][]{leobacher-2014} and then
defining~$f$ as the resulting GP posterior mean. We
use a single constraint function~$c_1$ whose ground truth is sampled in the
same way as~$f$.  The evaluations for~$f$ and~$c_1$ are contaminated with
i.i.d. Gaussian noise with variance~${\nu_1 = \nu_2 =0.01}$. 

\subsection{Assessing the Accuracy of the PESC Approximation}
\label{section:experiments:pesc-approximation-quality}

We first analyze the accuracy of the PESC approximation to the acquisition
function shown in~\cref{eq:new_acquisition_pesc}. We compare the PESC
approximation with a ground truth for the acquisition function obtained by
rejection sampling (RS). The RS method works by discretizing the search space
using a fine uniform grid. The expectation with respect to~$p(\xopt\given\data)$ in
\cref{eq:new_acquisition_pesc} is then approximated by Monte Carlo. To achieve
this, $\functions$ are sampled on the grid and the grid cell with non-negative
$c_1,\ldots,c_K$ (feasibility) and the lowest value of $f$ (optimality) is
selected. For each sample of
$\xopt$,~$\entropy\left[\yvalues\given\data,\x,\xopt\right]$ is approximated by
rejection sampling: we sample $\functions$ on the grid and select
those samples whose corresponding feasible optimal solution is the sampled
$\xopt$ and reject the other samples. We assume that the selected samples
for $\functions$ have a multivariate Gaussian distribution.  Under this
assumption, $\entropy\left[\yvalues\given\data,\x,\xopt\right]$ can be
approximated using the formula for the entropy of a multivariate Gaussian
distribution, with the covariance parameter in the formula being equal to the
empirical covariance of the selected samples for $f$ and $c_1,\ldots,c_K$ at
$\x$ plus the corresponding noise variances $\nu_1$ and
$\nu_2,\ldots,\nu_{K+1}$ in its diagonal. In our experiments, this approach
produces entropy estimates that are very similar, faster to obtain and less noisy than the
ones obtained with non-parametric entropy estimators. We compared this
implementation of RS with another version that ignores
correlations in the samples of $f$ and $c_1,\ldots,c_K$. In practice, both methods
produced equivalent results. Therefore, to speed up the method, we ignore
correlations in our implementation of RS.

\begin{figure}[t!]
\centering

\subfigure[Marginal posteriors]{
  \includegraphics[height=0.24\textwidth]{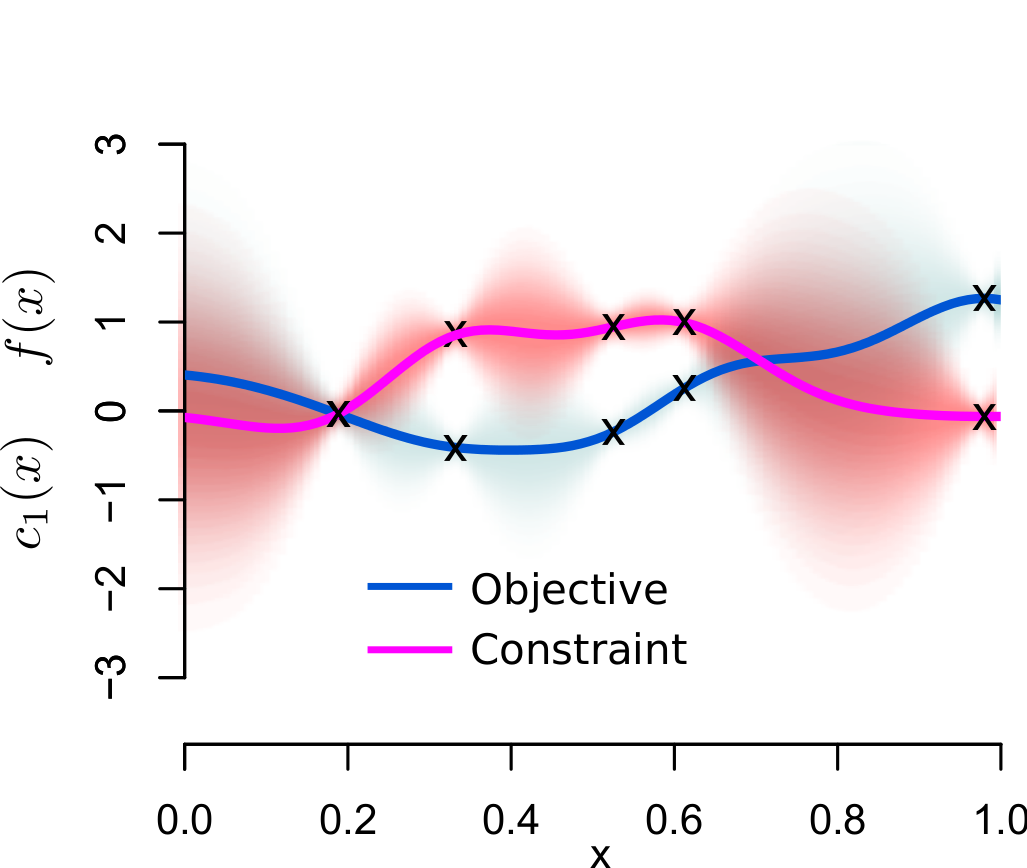}
  \label{fig:experiments:pesc:accuracy-marginal-posteriors}
}
\subfigure[Acquisition functions]{
  \includegraphics[height=0.24\textwidth]{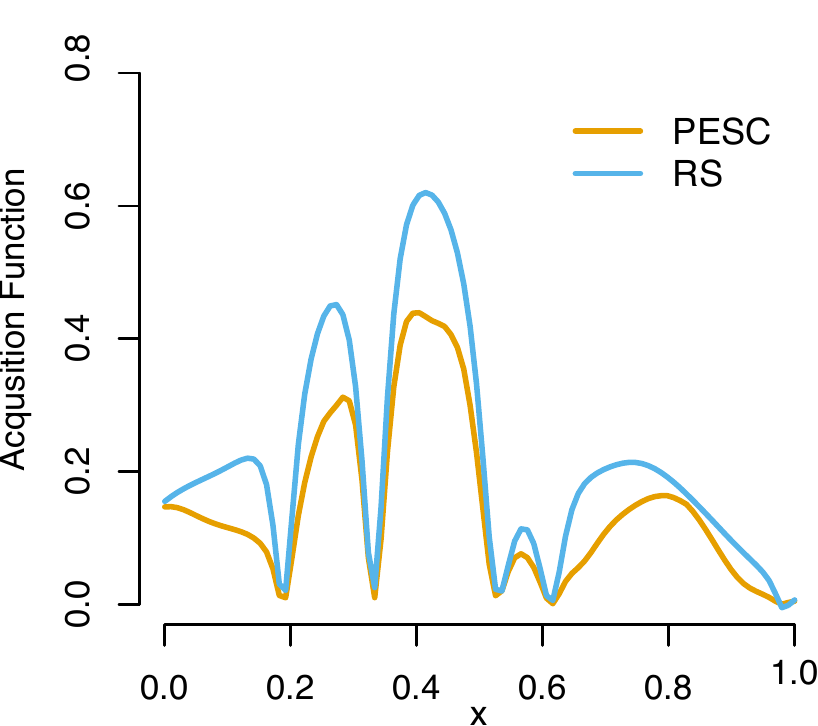}
  \label{fig:experiments:pesc:accuracy-acquisition-functions} 
}
\subfigure[Performance in $1\dimension$]{
  \includegraphics[height=0.24\textwidth]{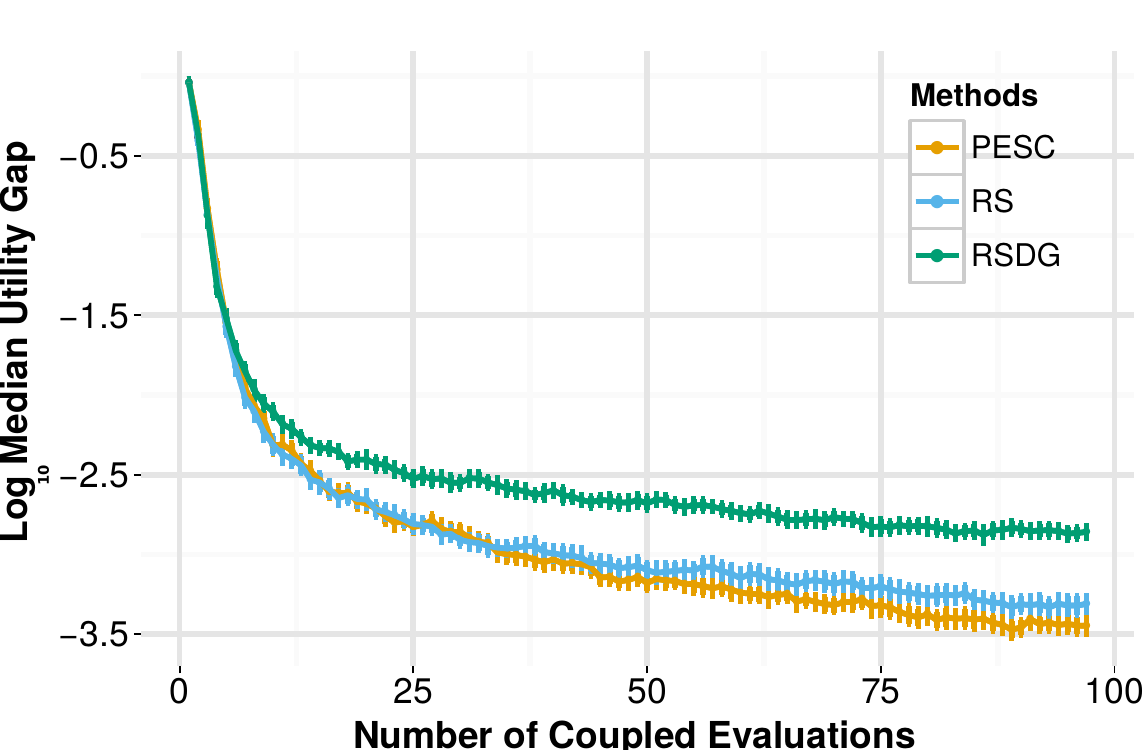}
  \label{fig:experiments:pesc:accuracy-performance} 
}

\caption[Accuracy of the PESC approximation.]{
Accuracy of the PESC approximation. (a) Marginal posterior
distributions for the objective and constraint given some collected data
denoted by $\times$'s. (b) PESC and RS acquisition functions given the data in
(a). (c) Median utility gap for PESC, RS and RSDG in the experiments with
synthetic functions sampled from the GP prior with $\dimension=1$.}
\label{fig:experiments:pesc-approximation-accuracy}
\end{figure}
 
\Cref{fig:experiments:pesc:accuracy-marginal-posteriors} shows the posterior
distribution for $f$ and $c_1$ given 5 observations sampled from the GP prior
with ${\dimension=1}$. The posterior is computed using the optimal GP
hyperparameters. The corresponding approximations to the acquisition function
generated by PESC and RS are shown in
\cref{fig:experiments:pesc:accuracy-acquisition-functions}. In the figure, both
PESC and RS use a total of $M=50$ samples from $p(\xopt\given\data)$ when
approximating the expectation in \cref{eq:new_acquisition_pesc}. The PESC
approximation is quite accurate, and importantly its maximum value is very
close to the maximum value of the RS approximation.  The approximation produced
by the version of RS that does not ignore correlations in the samples of
$\functions$ (not shown) cannot be visually distinguished from the one shown in
\cref{fig:experiments:pesc:accuracy-acquisition-functions}.

One disadvantage of the RS method is its high cost, which scales with the size
of the grid used.  This grid has to be large to guarantee good performance,
especially when $\dimension$ is large.
An alternative is to use a small dynamic grid that changes as data is
collected.  Such a grid can be obtained by sampling from $p(\xopt\given\data)$
using the same approach as in PESC to generate these samples (a similar
approach is taken by \citet{hennig-schuler-2012}, in which the dynamic grid is
sampled from the EI \acq). The samples obtained then form the dynamic grid,
with the idea that grid points are more concentrated in areas that we expect
$p(\xopt\given\data)$ to be high. The resulting method is called Rejection
Sampling with a Dynamic Grid (RSDG).

We compare the performance of PESC,  RS and RSDG in experiments with synthetic
data corresponding to 500 pairs of $f$ and $c_1$ sampled from the GP prior with
${\dimension=1}$. RS and RSDG draw the same number of samples of
$\xopt$ as PESC. We assume that the GP hyper-parameters are known to each
method and fix ${\delta = 0.05}$, that is, recommendations are made by finding
the location with highest posterior mean for $f$ such that $c_1$ is
non-negative with probability at least ${1-\delta}$.  For reporting purposes,
we set the utility~$u(\x)$ of a recommendation $\x$ to be $f(\x)$ if~$\x$
satisfies the constraint, and otherwise a penalty value of the worst (largest)
objective function value achievable in the search space.  For each
recommendation $\x$, we compute the utility gap $|u(\x) - u(\xopt)|$, where
$\xopt$ is the true solution to the optimization problem. Each method is
initialized with the same three random points drawn with Latin hypercube
sampling.

\Cref{fig:experiments:pesc:accuracy-performance} shows the median of the
utility gap for each method for the 500 realizations of $f$ and~$c_1$.  The
$x$-axis in this plot is the number of joint function evaluations for $f$ and
$c_1$.  We report the median because the empirical distribution of the utility
gap is heavy-tailed and in this case the median is more representative of the
location of the bulk of the data than the mean. The heavy tails arise because
we are averaging 
%measuring performance across 
over 500 different optimization problems with very different degrees of
difficulty.  In this and all of the following experiments, unless otherwise specified,
error bars are computed using the bootstrap method. The plot shows that PESC and RS
are better than RSDG. Furthermore, PESC is very similar to RS, with PESC even
performing slightly better, 
%at the end of the data collection process 
perhaps because PESC is not confined to a grid as RS is. These results seem to indicate
that PESC yields an accurate approximation of the information gain.  
%Furthermore, although RSDG performs worse than PESC, RSDG is faster because
%the rejection sampling operation (with a small grid) is less expensive than
%the EP algorithm. When optimizer speed becomes an issue, the PESC-F method
%(\cref{section:pesc:fast-updates}) is a useful alternative to PESC.  Thus,
%RSDG is an attractive alternative to PESC when the available computing time is
%very limited.

\subsection{Synthetic Functions in 2 and 8 Input Dimensions}
\label{section:experiments:pesc-synthetic-2-and-8-dim}

\begin{figure}[t!]
\centering
\subfigure[$\dimension=2$]{
  \includegraphics[width=0.45\textwidth]{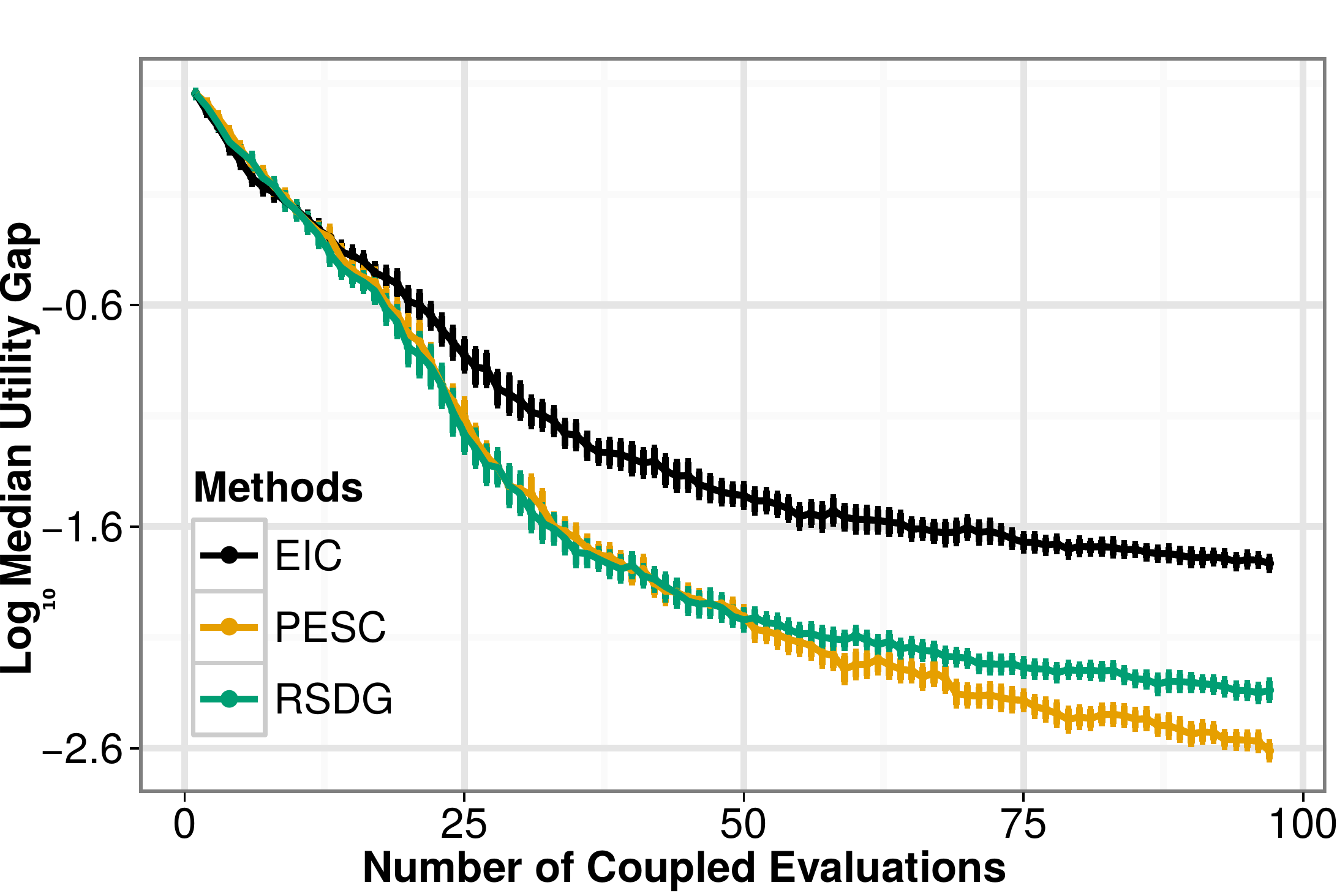}
  \label{fig:experiments:pesc-synthetic-2-dim}
}
\subfigure[$\dimension=8$]{
  \includegraphics[width=0.45\textwidth]{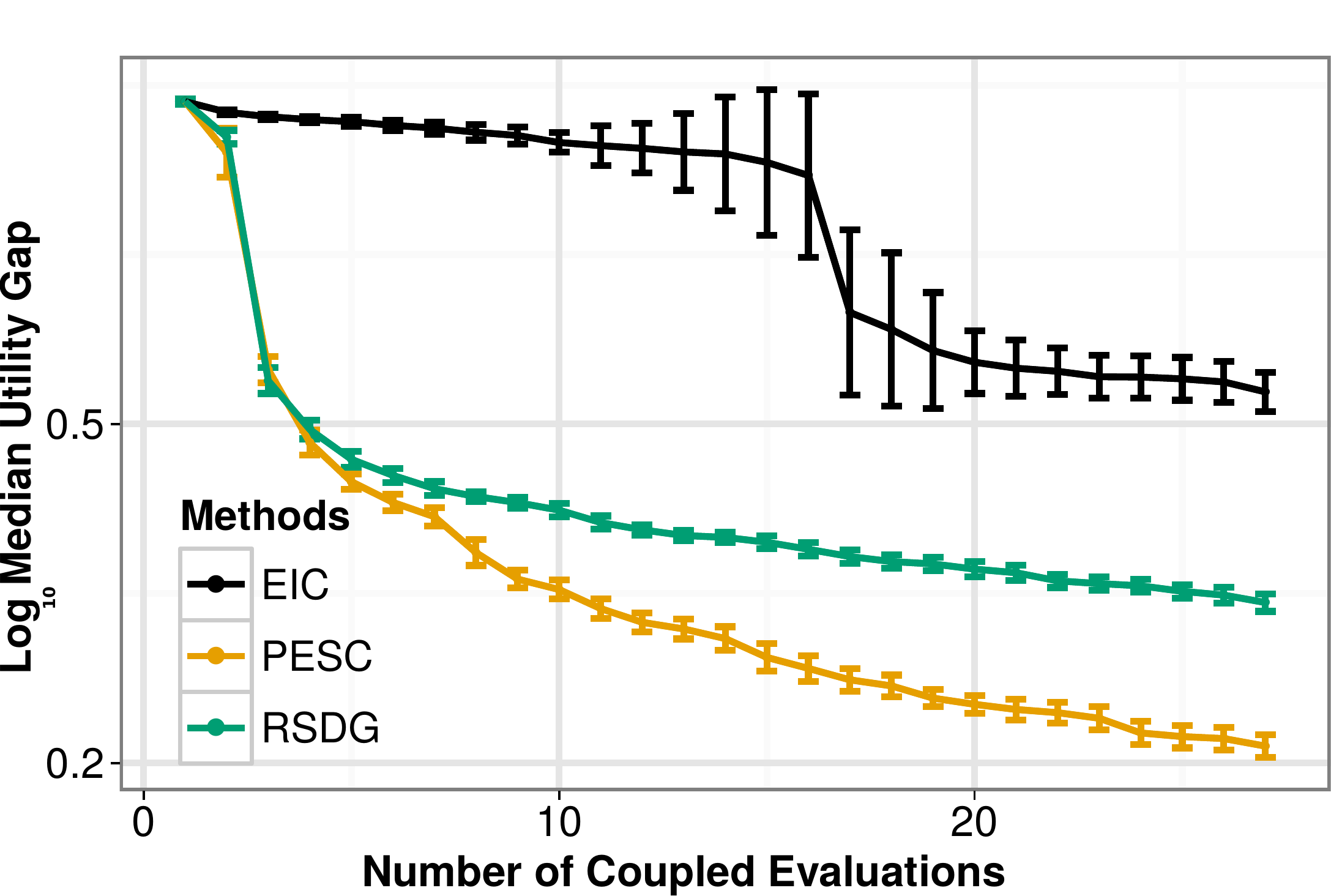}
  \label{fig:experiments:pesc-synthetic-8-dim}
}
\caption[Evaluating PESC with synthetic data.]{Optimizing samples from the GP
prior with (a) $\dimension=2$ and (b) $\dimension=8$. 
%, comparing EIC (black), PESC (orange), and RSDG (green).
}
\label{fig:experiments:pesc-synthetic-2-and-8-dim}
\end{figure}

We compare the performance of PESC and RSDG with EIC using the same
experimental protocol as in the previous section, but with
dimensionalities~${\dimension=2}$ and~${\dimension=8}$.  We do not compare with
RS here because its use of grids does not scale to higher dimensions.
\cref{fig:experiments:pesc-synthetic-2-and-8-dim} shows the utility gap for
each method across 500 different samples of $f$ and $c_1$ from the GP prior
with (a) ${\dimension=2}$ and (b) ${\dimension=8}$. Overall, PESC is the best
method, followed by RSDG and EIC.  RSDG performs similarly to PESC
when~${\dimension=2}$, but is significantly worse when~${\dimension=8}$. This
shows that, when $\dimension$ is high, grid based approaches (e.g. RSDG) are at
a disadvantage with respect to methods that do not require a grid (e.g. PESC).

\subsection{A Toy Problem}
\label{section:experiments:coupled:toy-problem}

\begin{figure}[t]
\centering
\subfigure[Visualization of the problem.]{
\includegraphics[width=0.34\columnwidth]{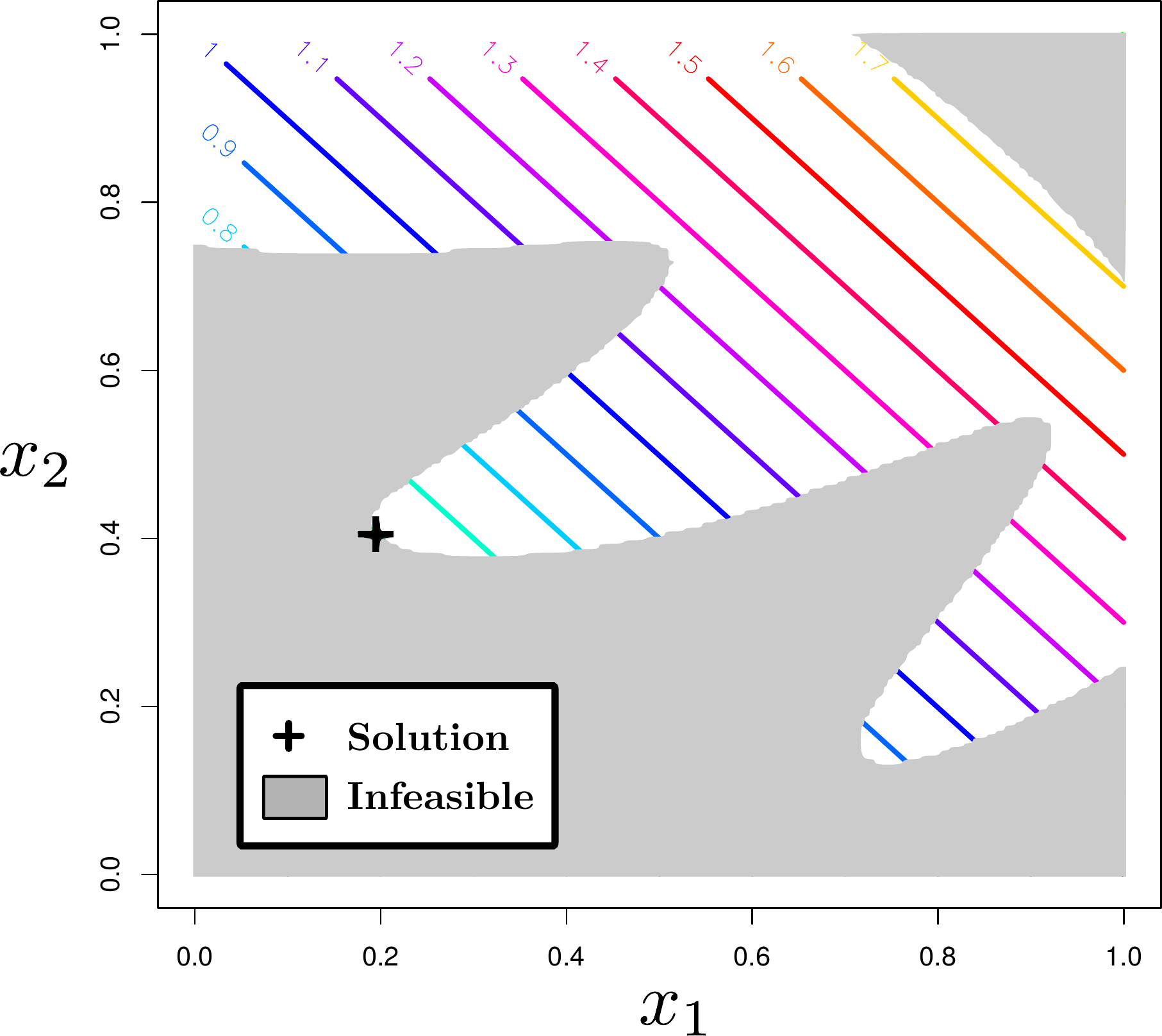}
\label{fig:visualization-toy-example}
}
\hspace{0.5cm}
\subfigure[Performance of PESC, AL and EIC.]{
\includegraphics[width=0.5\columnwidth]{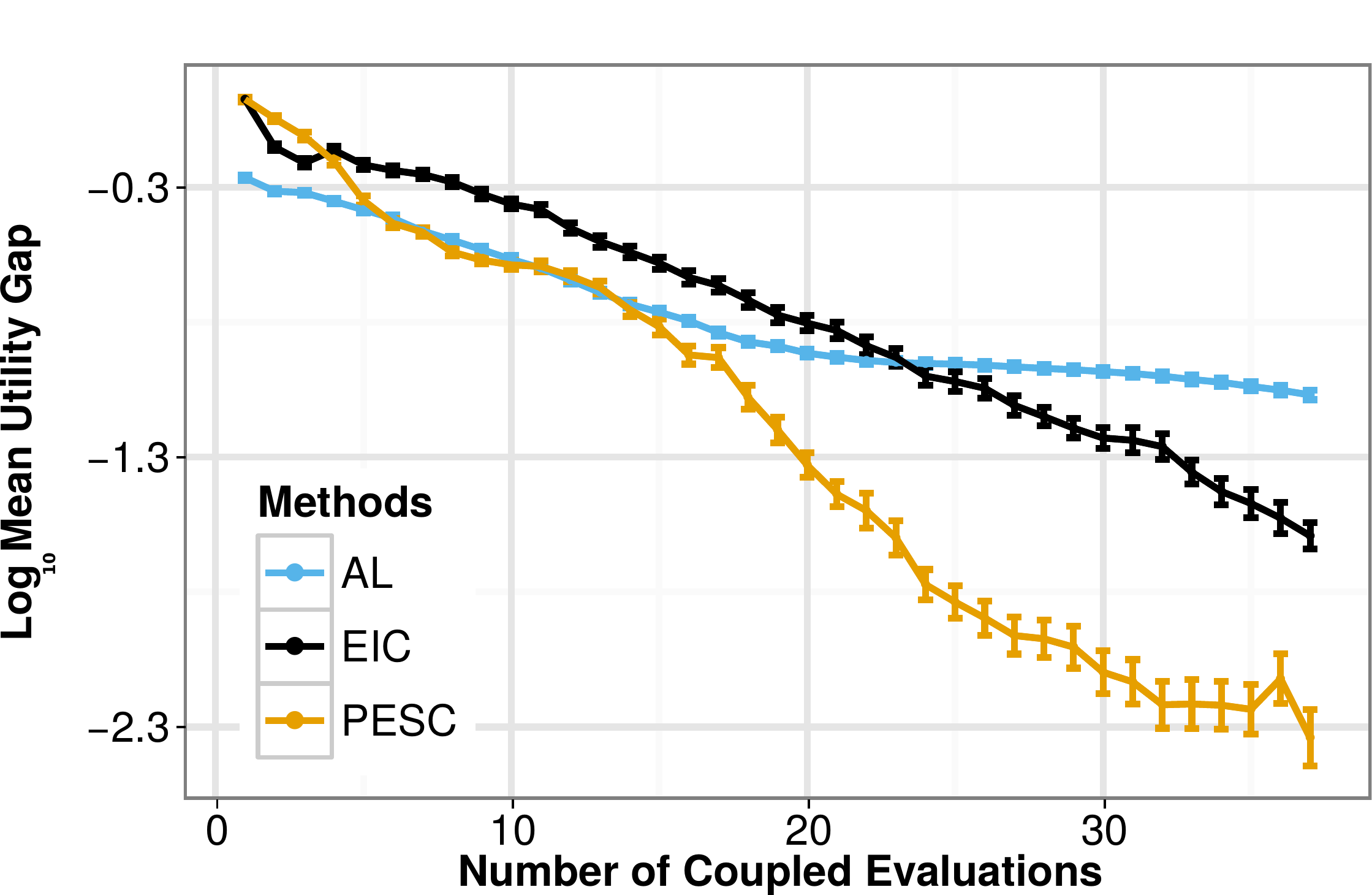}
\label{fig:results-toy-example}
}
\caption[Comparing PESC, AL, and EIC on a toy problem.]{Comparing PESC, AL, and
EIC in the toy problem described by \citet{gramacy-augmented-lagrangian}. (a)
Visualization of the linear objective function and the feasible and infeasible
regions. (b) Results obtained by PESC, AL and EIC on the toy problem.} 
\label{fig:toy-example}
\end{figure}

Next, we compare PESC with EIC and AL (\citet{gramacy-augmented-lagrangian},
\cref{section:prior-work:gramacy-augmented-lagrangian}) on the toy problem
described by \citet{gramacy-augmented-lagrangian}, namely,
\begin{align}
\label{eq:experiments:toy-problem}
\min_{\x\in[0,1]^2}  f(\x) & \text{ s.t. } c_1(\x)\geq0, \, c_2(\x)\geq 0 \, , \\
f(\x) & = x_1 + x_2 \, , \nonumber \\
c_1(\x) &= 0.5\sin{(2\pi(x_1^2-2x_2))}+x_1+2x_2-1.5 \, , \nonumber \\
c_2(\x) &= -x_1^2-x_2^2 + 1.5 \, . \nonumber
\end{align}
This optimization problem has two local minimizers and one global minimizer. At
the global solution, which is at $\xopt\approx[0.1954, 0.4404]$, only one of
the two constraints ($c_1$) is active. Since the objective is linear and $c_2$
is not active at the solution, learning about $c_1$ is the main challenge of
this problem. \cref{fig:visualization-toy-example} shows a visualization of the
linear objective function and the feasible and infeasible regions, including the
location of the global constrained minimizer $\xopt$.

In this case, the evaluations for $f$, $c_1$ and $c_2$ are noise-free.
To produce recommendations in PESC and EIC, we use the confidence value
${\delta = 0.05}$. We also use a squared
exponential GP kernel. PESC uses $M=10$ samples of $\xopt$ when approximating the
expectation in \cref{eq:new_acquisition_pesc}. We use the AL implementation provided
by \citet{gramacy-augmented-lagrangian} in the R~package \emph{laGP}, which is
based on the squared exponential kernel and assumes the objective $f$ is known.
Thus, in order for this implementation to be used, AL has an advantage over
other methods in that it has access to the true objective function. In all
three methods, the GP hyperparameters are estimated by maximum likelihood.

\Cref{fig:results-toy-example} shows the mean utility gap for each
method across 500 repetitions. Each repetition corresponds to a
different initialization of the methods with three data points selected with
Latin hypercube sampling.  
%Here, we report the mean because we are now measuring performance across
%realizations of the same optimization problem and the heavy-tailed effect
%described in \cref{section:experiments:pesc-approximation-quality} is less
%severe. 
The results show that PESC is significantly better than EIC and AL for this
problem.  EIC is superior to AL, which performs slightly better at the
beginning, possibly because it has access to the ground truth objective $f$.

\subsection{Finding a Fast Neural Network}
\label{section:nnet}

In this experiment, we tune the hyper-parameters of a three-hidden-layer neural
network subject to the constraint that the prediction time must not exceed 2 ms
on an NVIDIA GeForce GTX~580 GPU (also used for training). 
We use the Mat\'ern 5/2 kernel for the GP priors.
The search space
consists of 12 parameters: 2 learning rate parameters (initial value and decay rate),
2 momentum parameters (initial and final values, with linear interpolation), 2 dropout
parameters (for the input layer and for other layers), 2 additional regularization parameters
(weight decay and max weight norm), the number of hidden units in each of the 3
hidden layers, and the type of activation function (RELU or sigmoid). The network is
trained using the \emph{deepnet} package\footnote{\url{https://github.com/nitishsrivastava/deepnet}}, and the prediction time is computed
as the average time of 1000 predictions for mini-batches of size 128. The
network is trained on the MNIST digit classification task with momentum-based
stochastic gradient descent for 5000 iterations. The objective is reported as
the classification error rate on the standard validation set. For reporting
purposes, we treat constraint violations as the worst possible objective value
(a classification error of 1.0).
This experiment is inspired by a real need for neural networks that can make
fast predictions with high accuracy. An example is given by computer vision
problems in which the prediction time of the best performing neural network is
not fast enough to keep up with the fast rate at which new data is available
(e.g., YouTube, connectomics). 

\begin{figure}
\centering

\subfigure[Finding a fast neural network.]{
  \includegraphics[width=0.45\columnwidth]{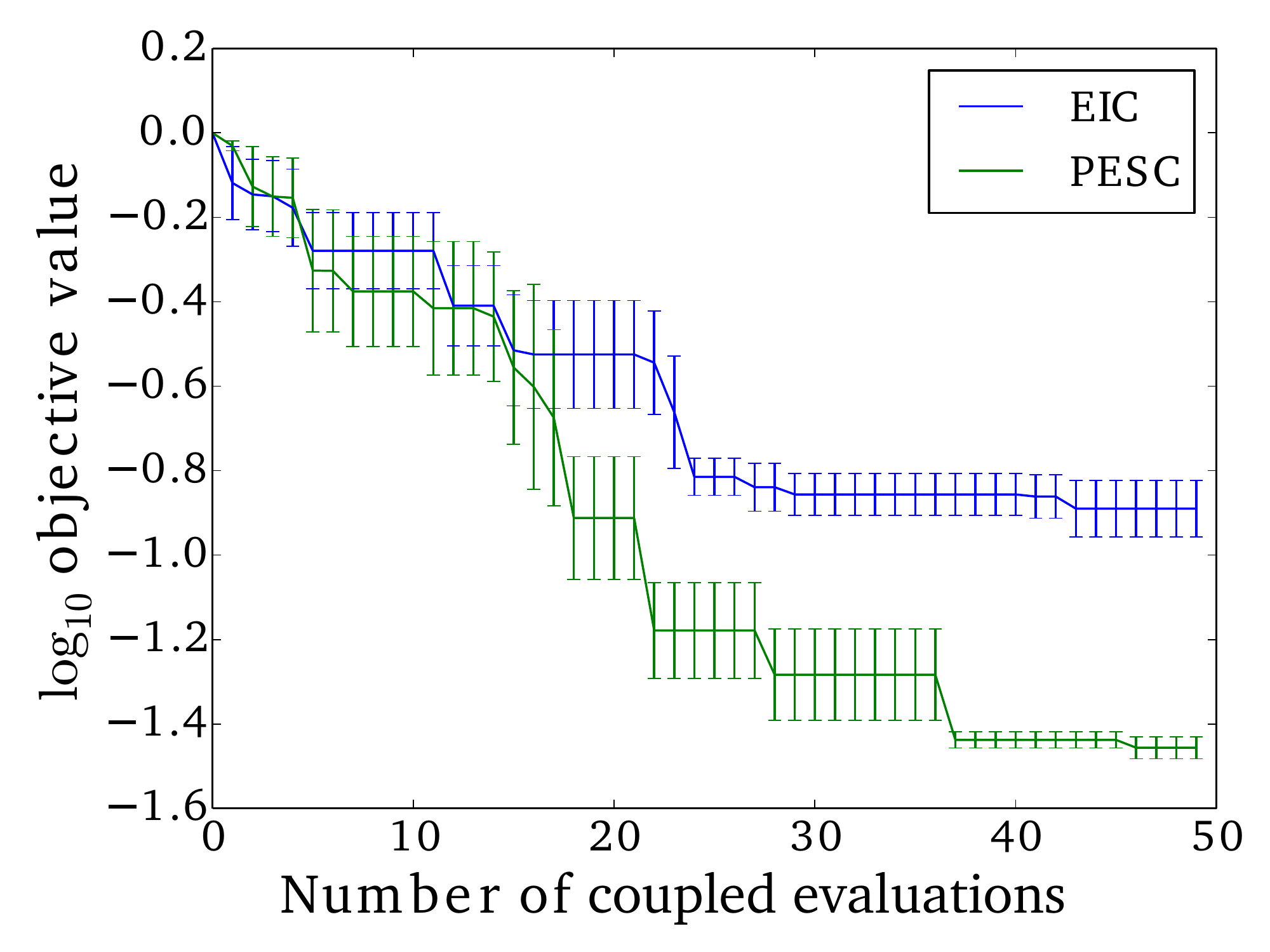}
  \label{fig:experiments:pesc:net-coupled}
}
\subfigure[Tuning Hamiltonian Monte Carlo]{
  \includegraphics[width=0.45\columnwidth]{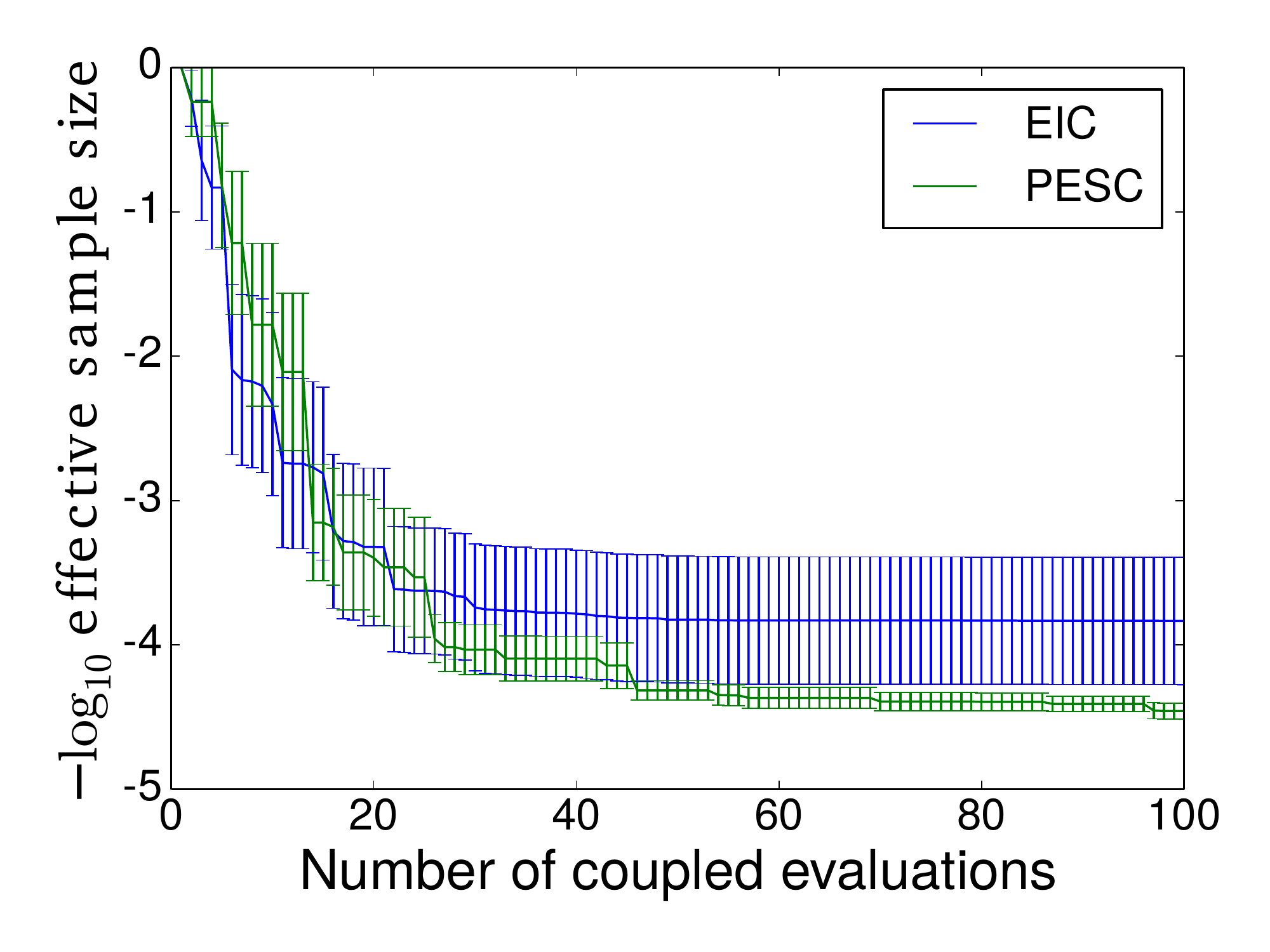}
  \label{fig:experiments:pesc:hmc-coupled}
}
\caption[Comparing PESC and EIC on machine learning problems.]{Results for PESC
and EIC on the tuning of machine learning methods with coupled constraints. (a) Tuning a
neural network subject to the constraint that it makes predictions in under 2
ms. (b) Tuning Hamiltonian Monte Carlo to maximize the number of effective
samples within 5 minutes of compute time, subject to the constraints passing
the Geweke and Gelman-Rubin convergence diagnostics and integrator stability.}
\label{fig:experiments:pesc:neuralnet-and-hmc}
\end{figure}

\Cref{fig:experiments:pesc:net-coupled} shows the results of 50 iterations of
the Bayesian optimization process. In this experiment and in the next one, the
$y$-axis represents the best objective value observed so far, with
recommendations produced using ${\delta=0.05}$ and observed constraint
violations resulting in objective values equal to 1.0. Curves show averages over
five independent experiments. In this case, PESC performs significantly better
than EIC. 

When the constraints are noisy, reporting the best observation is an overly
optimistic metric because the best feasible observation might be infeasible in
practice.
On the other hand,
ground-truth is not available. Therefore, to validate our results further, we
used the recommendations made at the final iteration of the Bayesian
optimization process for each method (EIC and PESC) and evaluated the functions
with these recommended parameters. We repeated the evaluation 10 times for each
of the 5 repeated experiments. The result is a ground-truth score obtained as
the average of 50 function evaluations. This procedure yields a score of $7.0
\pm 0.6 \%$ for PESC and $49 \pm 4 \%$ for EIC (as in the figure, constraint
violations are treated as a classification error of $100\%$), where the numbers
after the $\pm$ symbol denote the empirical standard deviation.  This result is
consistent with \cref{fig:experiments:pesc:net-coupled} in that PESC performs
significantly better than EIC. 
%, but also demonstrates that, due to noise, Figure~\ref{fig:net-coupled} is
%overly optimistic. While we may believe this optimism to affect both methods
%equally, the ground-truth measurement provides a more reliable result and a
%much clearer understanding of the classification error attained by Bayesian
%optimization.
%We propose the use of PESC-based Bayesian optimization to search for
%high-performing yet fast network architectures.

\subsection{Tuning Markov Chain Monte Carlo}
\label{section:experiments:pesc:hmc}

Hamiltonian Monte Carlo (HMC) \citep{duane1987hybrid} is a popular MCMC
technique that takes advantage of gradient information for rapid mixing. HMC
contains several parameters that require careful tuning. The two basic
parameters are the number of leapfrog steps and the step size.  HMC may also
include a mass matrix which introduces $\bigO(\dimension^2)$ additional
parameters for problems in $\dimension$ dimensions, although the matrix is
often fixed to be diagonal ($\dimension$~parameters) or a multiple of the
identity matrix (1~parameter) \citep{nealbook}.  In this experiment, we
optimize the performance of HMC.
We use again the Mat\'ern 5/2 kernel for the GP priors.
%; see \citet{mahendran-2012a} for a similar approach. 
We tune the following parameters: the number of leapfrog steps, the step size,
a mass parameter and the fraction of the allotted computation time spent
burning in the chain. Our experiment measures the number of effective samples
obtained in a fixed computation time. We impose the constraints that the
generated samples must pass the Geweke \citep{Geweke92} and Gelman-Rubin
\citep{gelmanrubin} convergence diagnostics.  In particular, we require the
worst (largest absolute value) Geweke test score across all variables and
chains to be at most 2.0, and the worst (largest) Gelman-Rubin score between
chains and across all variables to be at most 1.2.  We use the \emph{coda} R
package \citep{rcoda} to compute the effective sample size and the Geweke
convergence diagnostic, and the \emph{PyMC} python package \citep{PyMC} to
compute the Gelman-Rubin diagnostic over two independent traces. 
%The chosen thresholds for the convergence diagnostics are based on the PyMC
%and LaplacesDemon documentation. 

The HMC integration may also diverge for large values of the step size. We
treat this as a hidden constraint, and set ${\delta=0.05}$. We use HMC to
sample from the posterior of a logistic regression binary classification
problem using the German credit data set from the UCI repository
\citep{Frank2010}. The data set contains 1000 data points, and is normalized to
have zero mean unit variance for each feature. We initialize each chain
randomly with ${\dimension=25}$ independent draws from a Gaussian distribution
with mean zero and standard deviation $10^{-3}$. For each set of inputs, we
compute two chains, each one with five minutes of computation time on a single
core of a compute node.

 %We use the \emph{coda} R package \citep{rcoda} to compute the effective
 %sample size and the Geweke convergence diagnostic, and the \emph{PyMC} python
 %package \citep{PyMC} to compute the Gelman-Rubin diagnostic over two
 %independent traces. 

\Cref{fig:experiments:pesc:hmc-coupled} compares EIC and PESC on this task,
averaged over ten realizations of the experiment. 
%The weaker performance and large error bars of EIC are due to a pathology in
%this acquisition function that sometimes causes it remain in infeasible
%regions for long periods before finding a feasible region. 
As above, we perform a ground-truth assessment of the final recommendations.
For each method (EIC and PESC), we used the recommendations made at the final
iteration of the Bayesian optimization process and evaluated the functions with
these recommended parameters multiple times.  The resulting average effective
sample size is $3300 \pm 1200$ for PESC and $2300 \pm 900$ for EIC, where the
number after the $\pm$ symbol denotes the empirical standard deviation.  Here, the
difference between the two methods is within the margin of error.  When we
compare these results with the ones in \cref{fig:experiments:pesc:hmc-coupled}
we observe that the latter results are overly optimistic, indicating that this
experiment is very noisy. The noise presumably comes from the randomness in the
initialization and the execution of HMC, which causes the passing or the
failure of the convergence diagnostics to be highly stochastic.

\section{Empirical Analyses with Decoupled Functions}
\label{section:experiments_decoupling}

\Cref{section:experiments} focused on the evaluation of the performance of PESC
in experiments with coupled functions. Here, we evaluate the performance of PESC
in the decoupled case, where the different functions can be evaluated independently.

\subsection{Accuracy of the PESC Approximation}
\label{section:experiments:pesc-approximation-quality-decoupled}

\begin{figure}[t]
\centering

\subfigure[Marginal posteriors]{
  \includegraphics[width=0.31\textwidth]{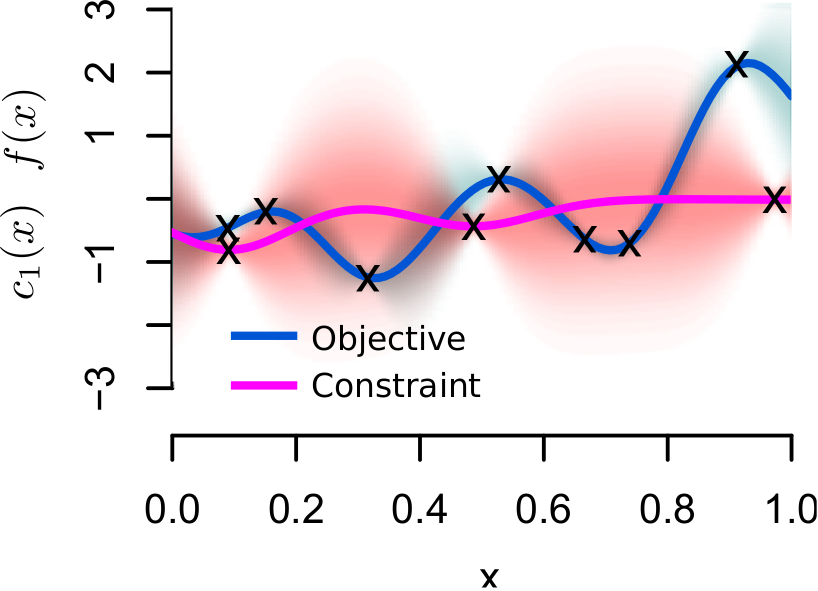}
  \label{fig:experiments:decoupled:accuracy:posteriors1}
}
\subfigure[$\alpha_f(x)$]{
  \includegraphics[width=0.31\textwidth]{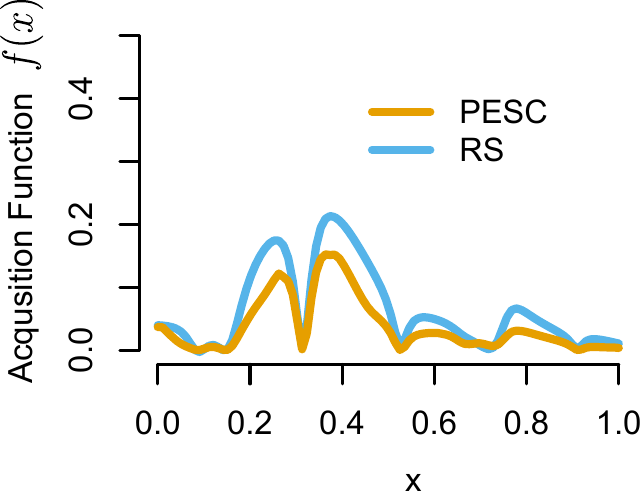}
  \label{fig:experiments:decoupled:accuracy:acqs1f}
}
\subfigure[$\alpha_c(x)$]{
  \includegraphics[width=0.31\textwidth]{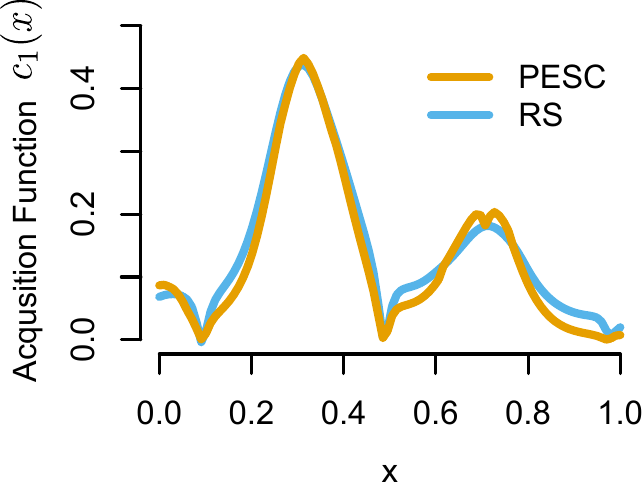}
  \label{fig:experiments:decoupled:accuracy:acqs1c}
}
%\subfigure[$\xopt$ histogram]{
%  \includegraphics[width=0.22\textwidth]{figures/decoupling/acquisitionFunctions/xstar_histo1.pdf}
%  \label{fig:experiments:decoupled:accuracy:hist1}
%}
\subfigure[Marginal posteriors]{
  \includegraphics[width=0.31\textwidth]{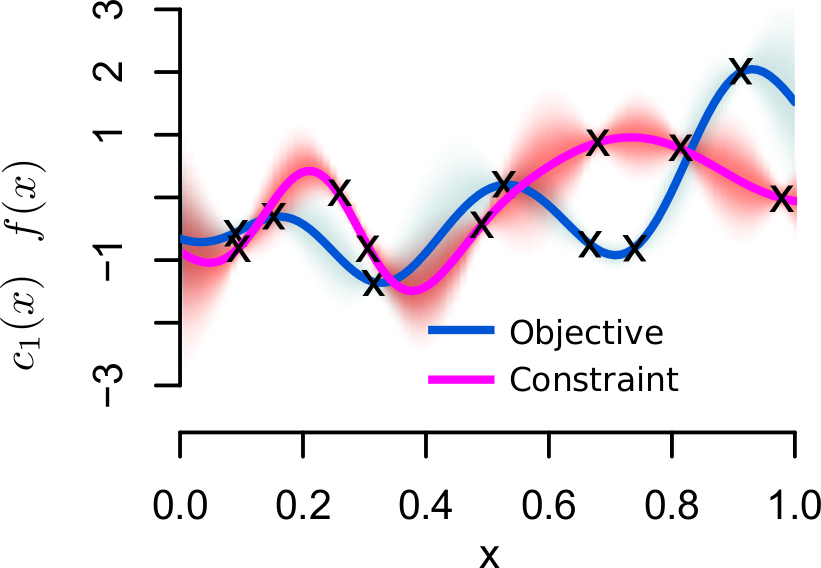}
  \label{fig:experiments:decoupled:accuracy:posteriors2}
}
\subfigure[$\alpha_f(x)$]{
  \includegraphics[width=0.31\textwidth]{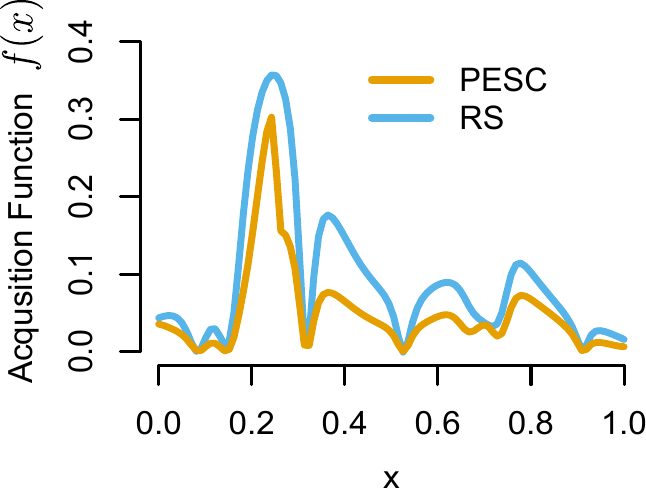}
  \label{fig:experiments:decoupled:accuracy:acqs2f}
}
\subfigure[$\alpha_c(x)$]{
  \includegraphics[width=0.31\textwidth]{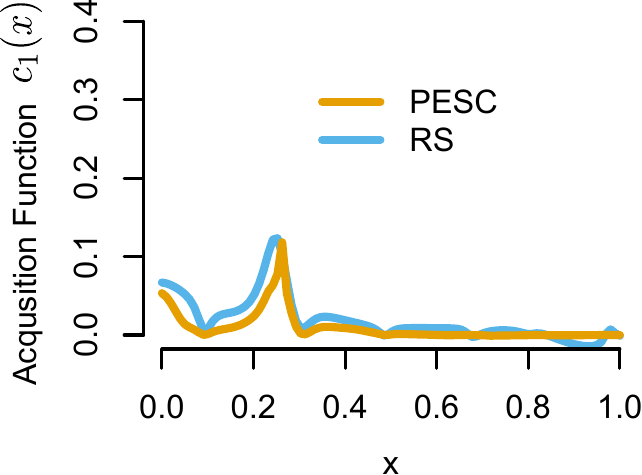}
  \label{fig:experiments:decoupled:accuracy:acqs2c}
}
%\subfigure[$\xopt$ histogram]{
%  \includegraphics[width=0.22\textwidth]{figures/decoupling/acquisitionFunctions/xstar_histo2.pdf}
%  \label{fig:experiments:decoupled:accuracy:hist2}
%}

\caption[Assessing the accuracy of the decoupled PESC approximation.]{Assessing
the accuracy of the decoupled PESC approximation for the partial acquisition
functions for $\alpha_f(x)$ and $\alpha_c(x)$. Between the top and bottom rows,
three additional observations of the constraint have been made.}
\label{fig:experiments:decoupled:accuracy}
\end{figure}

We first evaluate the accuracy of PESC when approximating the function-specific
acquisition functions from \cref{{eq:individual_acq}}. We consider a synthetic
problem with input dimension ${\dimension=1}$ and including an objective
function and a single constraint function, both drawn from the GP prior
distribution.  \Cref{fig:experiments:decoupled:accuracy:posteriors1} shows the
marginal posterior distributions for $f$ and $c_1$ given 7 observations for the
objective and 3 for the constraint.
\Cref{fig:experiments:decoupled:accuracy:acqs1f,fig:experiments:decoupled:accuracy:acqs1c}
show the PESC approximations to the acquisition functions for the objective and
the constraint, respectively. These functions approximate how much information
we would obtain by the individual evaluation of the objective or the constraint
at any given location.  We also include in
\cref{fig:experiments:decoupled:accuracy:acqs1f,fig:experiments:decoupled:accuracy:acqs1c}
the value of a ground truth obtained by rejection
sampling (RS). The RS solution is obtained in the same way as in
\cref{section:experiments:pesc-approximation-quality}.  Both PESC and RS use a
total of $M=50$ samples from $p(\xopt\given\data)$.  The PESC approximation is
quite accurate, and importantly its maximum value is very close to the maximum
value of the RS approximation.
\Cref{fig:experiments:decoupled:accuracy:acqs1f,fig:experiments:decoupled:accuracy:acqs1c}
indicate that the highest expected gain of information is obtained by
evaluating the constraint at $x\approx 0.3$.  The reason for this is that, as
\cref{fig:experiments:decoupled:accuracy:posteriors1} shows, the objective
is low near $x\approx0.3$ but the constraint has not been evaluated at that location yet.

\Cref{fig:experiments:decoupled:accuracy:posteriors2} shows the marginal
posterior distributions for $f$ and $c_1$ when three more observations have
been collected for the constraint.  The corresponding approximations given by
PESC and RS to the function-specific acquisition functions are shown in
\cref{fig:experiments:decoupled:accuracy:acqs2f,fig:experiments:decoupled:accuracy:acqs2c}.
As before, the PESC approximation is very similar to the RS one.
In this case, evaluating the constraint is no longer as informative as before and
the highest expected gain of information is obtained by
evaluating the objective at $x\approx 0.25$.
Intuitively, as we collect more constraint observations the constraint becomes
well determined and the optimizer turns its attention to the objective.

\subsection{Comparing Coupled and Decoupled PESC}

We now compare the performance of coupled and decoupled versions of PESC in the
same decoupled optimization problem. This allows us to empirically demonstrate
the benefits of treating a decoupled problem as such.  

We first consider the toy problem from
\cref{section:experiments:coupled:toy-problem} given by
\cref{eq:experiments:toy-problem}. We assume that there are three decoupled
tasks: one for the objective and another one for each constraint function. We
further assume that there is a single resource $r$ with capacity
$\capacity(r)=3$. Each task requires to use resource $r$ for its evaluation and
the evaluation of each task takes always the same amount of time, which is
assumed to be much larger than the BO computations. At each iteration resource
$r$ is used to evaluate 3 functions in parallel. We compare the performance of
four versions of PESC, which differ in how they select the 3 parallel
evaluations that will be performed at each iteration. The first method is a
coupled approach (Coupled) which, at each iteration, evaluates jointly the
three tasks at the same input. The second method is a non-competitive
decoupling approach (NCD) which, at each iteration, evaluates all the different
tasks once but not necessarily at the same input. This is equivalent to
assuming that there are 3 resources with capacity 1 and each task can only be
evaluated in one resource: the tasks do not have to compete because each one
can only be evaluated in its corresponding resource. The third method is a
competitive-decoupling approach (CD) which allows the different tasks to
compete such that, at each iteration, three not necessarily unique functions
are evaluated at three not necessarily unique locations. We also consider an
implementation of CD that is not based on PESC and uses the EIC-D approach, as described in
\cref{section:prior-work:eic-d}. We call this method EIC-CD.  EIC-CD works like
CD, with the difference that, at each step, we first determine the next
evaluation location~$\x$ by maximizing the EIC acquisition function.  After
this, the next task to be evaluated at~$\x$ is chosen according to the expected
reduction in the entropy of the global feasible minimizer $\xopt$. The original
description of this method given by \citet{Gelbart2014} approximates the
expected reduction in entropy using Monte Carlo sampling. This is in general
computationally very expensive. To speed up EIC-CD, we replace the Monte Carlo
sampling step by the approximation of the expected reduction in entropy given
by PESC.

All the methods have to update the GP model, the posterior samples of $\xopt$
and the EP solutions just after collecting the data from resource $r$. However,
the method CD and EIC-CD have to do two additional update operations after
sending the first and the second evaluations to resource $r$, respectively.
These updates correspond to step 11 in \cref{algorithm:decoupling:general} and
they allow CD and EIC-CD to condition on pending evaluations that are not
complete yet. 
% This conditioning can be done by drawing virtual data from the GP
% predictive distribution and then averaging across the predictions made when
% each virtual data point is assumed to be the data actually collected
% \citep{snoek-etal-2012b}. In practice we follow this approach using only one
%virtual data point for each pending evaluation, 
In our experiments we use the Kriging believer approach,
in which we pretend that the pending function evaluations have completed and returned the values of the GP predictive mean at those locations. This allows the methods CD and EIC-CD
to update the GP model in a fast way, at the cost of ignoring uncertainty in the
predictions of the GP model.
The samples of $\xopt$ and the EP
solutions are, however, recomputed from scratch once the GP model has been
updated. This can be expensive in practice. To address this problem we
introduce the method CD-F, which works like CD, but replaces the full updates
for the samples of $\xopt$ and the EP solutions with the corresponding fast
updates used by PESC-F in \cref{section:pesc-fast-updates}. Therefore, by
comparing CD and CD-F, we can evaluate the loss in performance that is obtained
by using the fast PESC-F updates. Note that CD-F uses the fast updates only
after sending the first and the second evaluations to resource $r$. Once the
new data is collected, CD-F uses the original slow updates.

\begin{figure}[t!]
\centering
\subfigure[Performance]{
  \includegraphics[width=0.45\textwidth]{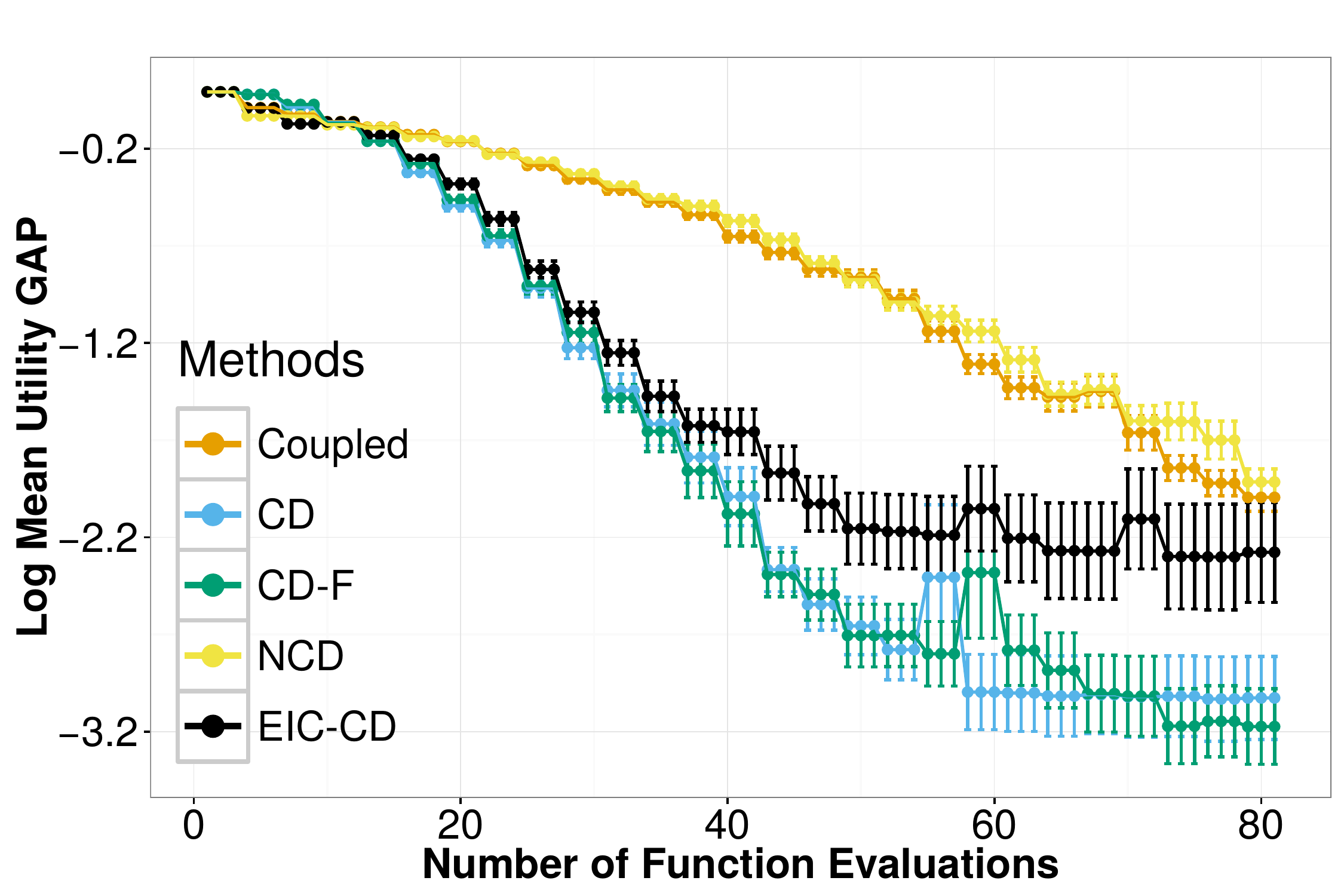}
  \label{fig:experiments:decoupling:toy-problem:performance}
}
\subfigure[Cumulative task evaluations]{
  \includegraphics[width=0.45\textwidth]{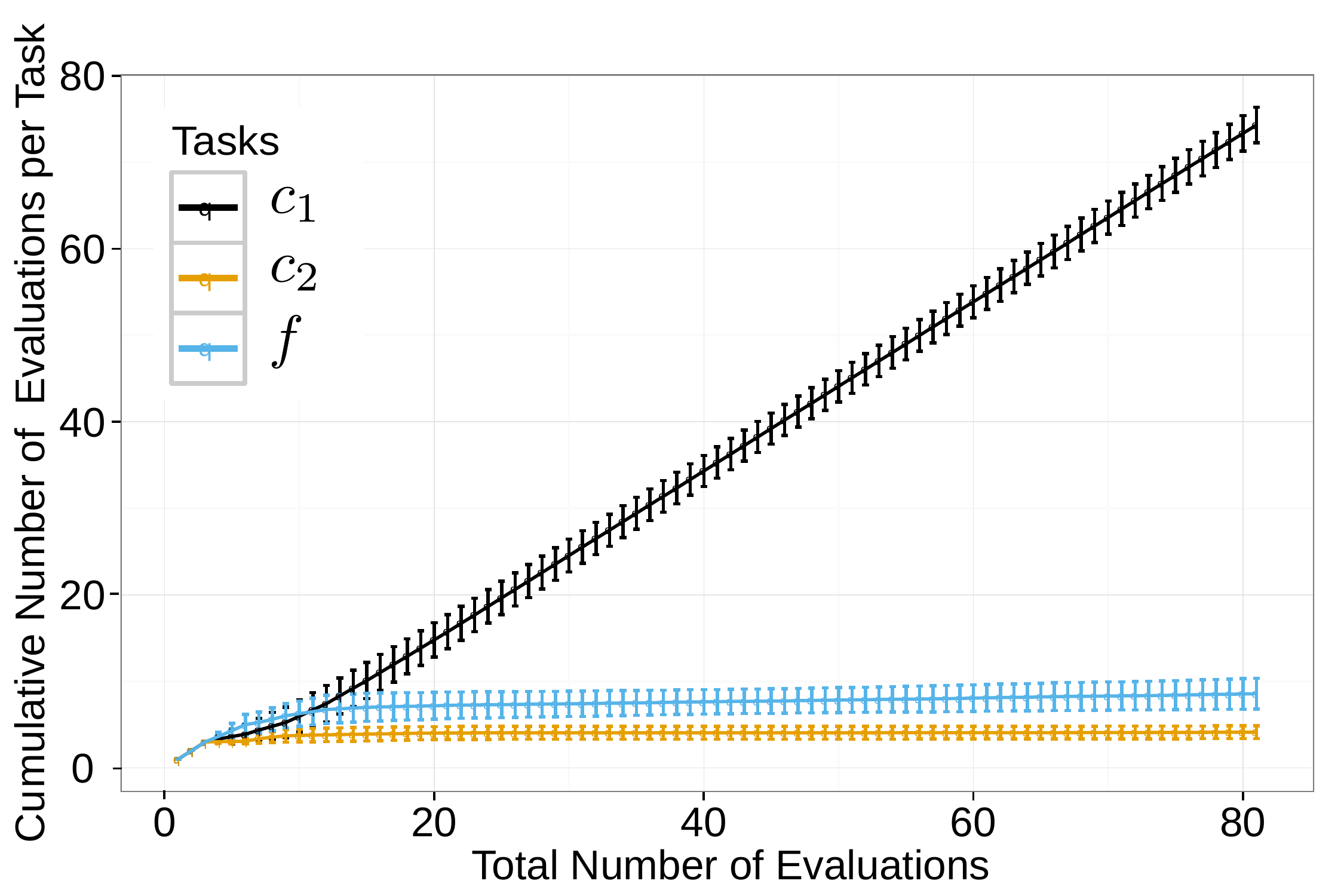}
  \label{fig:experiments:decoupling:toy-problem:evals}
}
\caption[Comparing Coupled, CD, NCD and EIC-CD approaches on the toy problem.]{
Results for the decoupled toy problem (\cref{eq:experiments:toy-problem}) when using
a resource $r$ that can evaluate 3 tasks ($f$, $c_1$ or $c_2$) in parallel.
(a) Performance comparison of Coupled (orange), NCD (yellow), CD (blue), CD-F (green) and EIC-CD (black) approaches.
(b) Cumulative number of evaluations for each task performed by CD-F. The algorithm automatically discovers that the constraint $c_1$ is much more important than the objective $f$ or the other constraint $c_2$.
} \label{fig:experiments:decoupling:toy-problem}
\end{figure}

\Cref{fig:experiments:decoupling:toy-problem:performance} shows the results
obtained by each method across 500 repetitions of the experiment starting from
random initializations. Recommendations are computed with
$\delta=0.01$. The horizontal axis in the plot denotes the number of
function evaluations performed so far. Since $\capacity(r)=3$, these evaluation
are performed in parallel in blocks of three.  The vertical axis denotes the
average utility gap, computed as in
\cref{section:experiments:pesc-approximation-quality}.  Overall, CD and CD-F
perform the best; the fact that CD and CD-F
obtain similar results implies that the fast PESC-F updates incur no
significant performance loss in this synthetic optimization problem.  EIC-CD is worse
than these two methods. This is a result of the the sub-optimal two-stage
decision process used by EIC-CD to select the next evaluation location and
the next task to be evaluated at that location; see \cref{section:prior-work:eic-d} for more details. NCD performs about the same as Coupled
which means that, in this problem, the benefits of decoupling come from
choosing an unequal distribution of tasks to evaluate, rather than from the
additional freedom of evaluating the three tasks at potentially different
locations. This hypothesis is corroborated by
\cref{fig:experiments:decoupling:toy-problem:evals}, which shows the average
cumulative number of evaluations performed by CD for each task ($f$, $c_1$ or
$c_2$) at each iteration. CD chooses to evaluate the constraint $c_1$ far more
often than the objective or the other constraint $c_2$. This makes sense since
the objective is a linear function and $c_1$, which has a complicated form, is
the only active constraint at the global solution. Thus, the PESC algorithm has automatically discovered that the constraint $c_1$ is much more important (both in the sense of being complicated and in the sense of being active at the true solution) than the objective $f$ or the other constraint $c_2$. This demonstrates the true power of competitive decoupling, as the algorithm avoids wasting time on uninteresting tasks that might be, in the worst case scenario, even more expensive than the interesting ones.

%In \cref{section:experiments:wall-time-experiments}, we show that
%PESC-F is also effective when conditioning on actual new observations.

% \begin{figure}[p]
% \centering
% \subfigure[Performance, all approaches]{
%   \includegraphics[width=0.5\textwidth]{figures/decoupling/synthetic-decoupled-2d.pdf}
%   \label{fig:experiments:decoupling:pesc-synthetic:2d-evals:active}
% }\\
% \subfigure[Constraint active at solution]{
%   \includegraphics[width=0.45\textwidth]{figures/decoupling/synthetic-decoupled-2d-constraint-active.pdf}
%   \label{fig:experiments:decoupling:pesc-synthetic:2d-evals:active}
% }
% \subfigure[Constraint not active at solution]{
%   \includegraphics[width=0.45\textwidth]{figures/decoupling/synthetic-decoupled-2d-constraint-not-active.pdf}
%   \label{fig:experiments:decoupling:pesc-synthetic:2d-evals:not-active}
% }
% \caption[Comparing coupled and decoupled approaches on synthetic data.]{Optimizing samples from the GP in $\dimension=2$. (a) Performance with coupled (black), competitive decoupled (CD, orange) and parallel CD (blue) PESC. (b-c) Cumulative function evaluations of the objective (orange) and constraint (black) when (a) the constraint is active at the true solution (about 30\% of cases) and (b) the constraint is not active at the true solution (about 70\% of cases). Curves reflect the mean over 500 realizations.
% }
% \label{fig:experiments:decoupling:pesc-synthetic:2d-evals}
% \end{figure}

\begin{figure}
\centering
\subfigure[Performance]{
\includegraphics[width=0.31\textwidth]{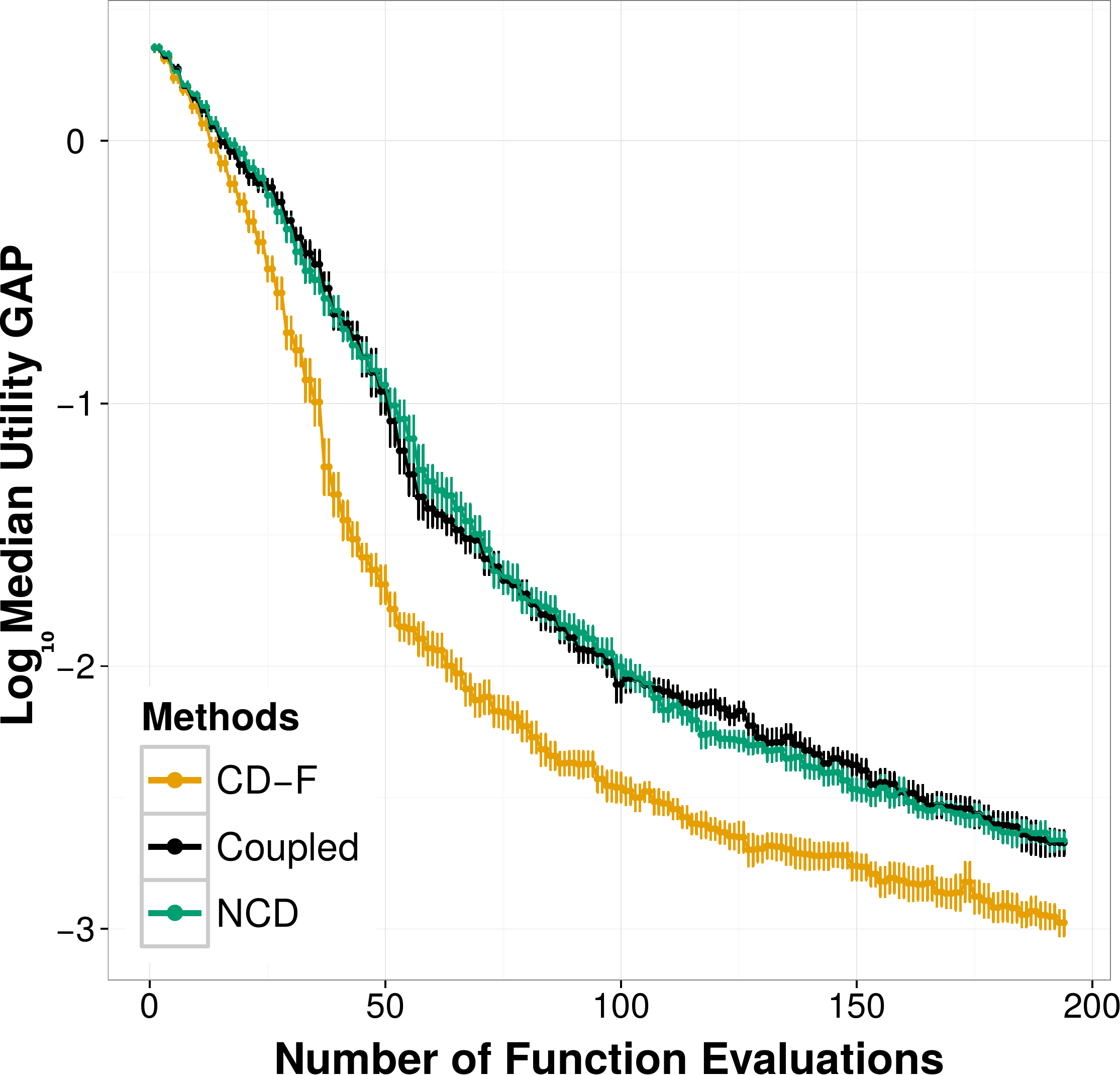}
  \label{fig:experiments:decoupling:pesc-synthetic:2d-performance}
}
\subfigure[Constraint active]{
  \includegraphics[width=0.31\textwidth]{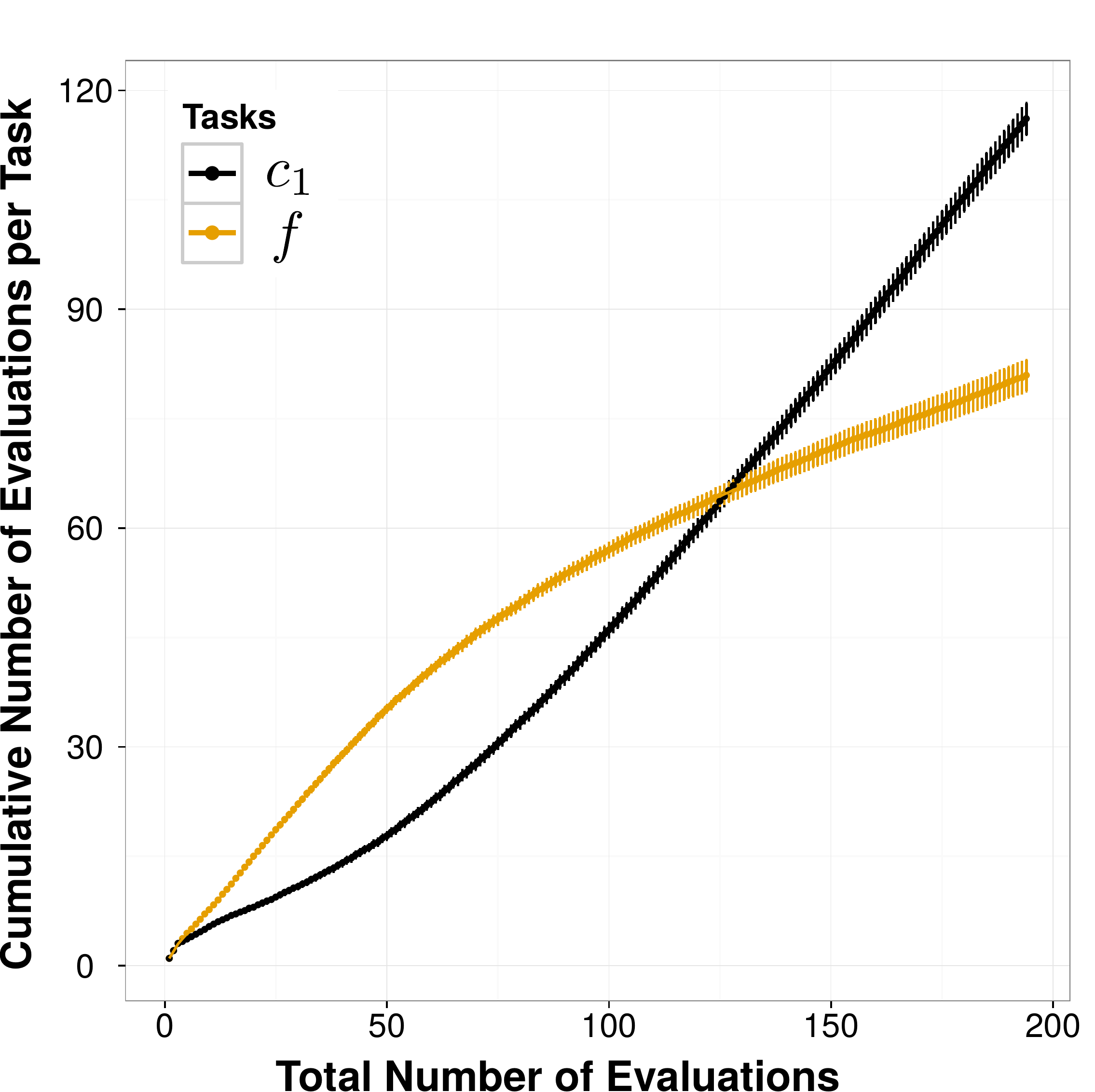}
  \label{fig:experiments:decoupling:pesc-synthetic:2d-evals:active}
}
\subfigure[Constraint not active]{
  \includegraphics[width=0.31\textwidth]{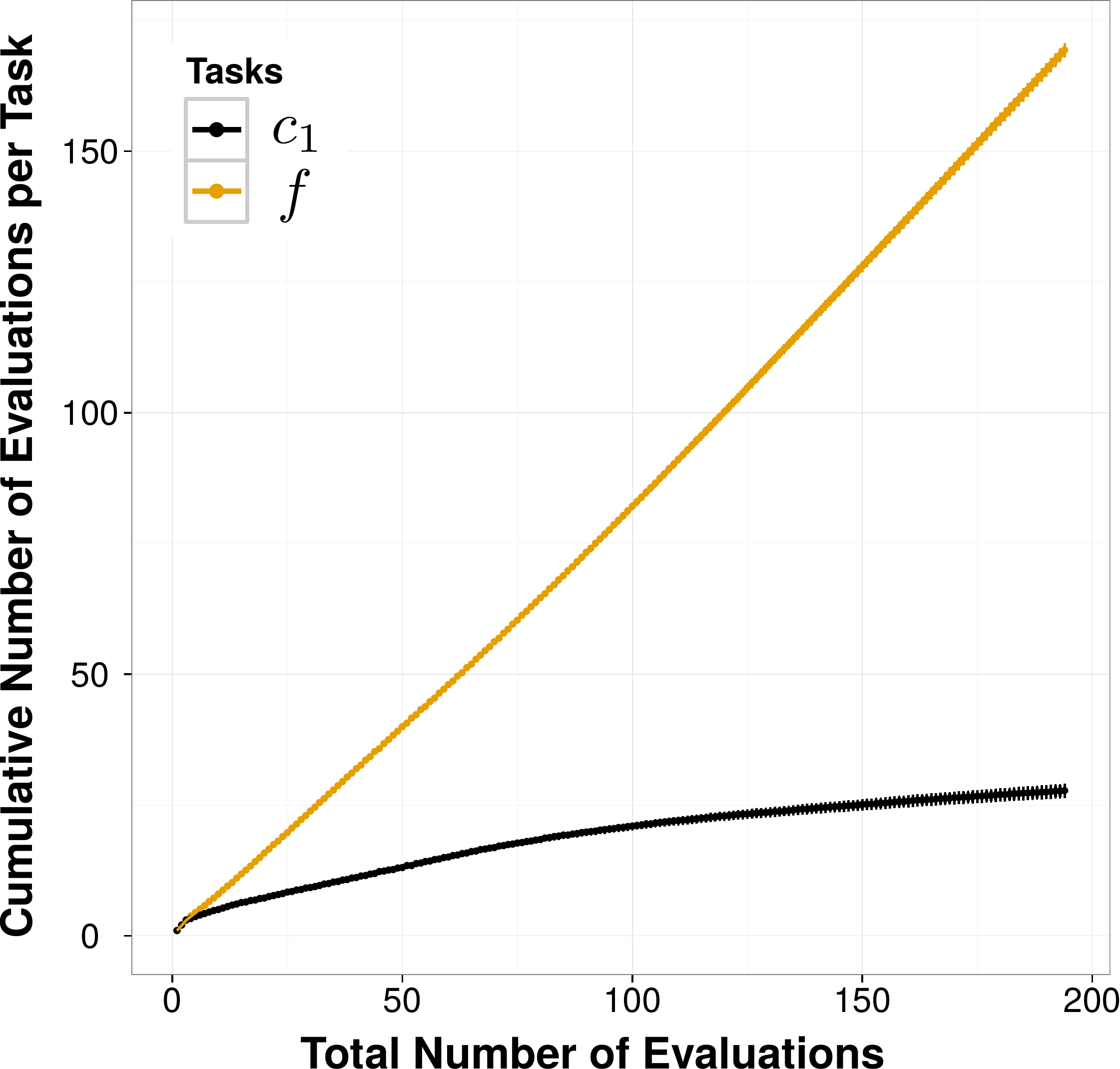}
  \label{fig:experiments:decoupling:pesc-synthetic:2d-evals:not-active}
}
\caption[Comparing coupled and decoupled approaches on synthetic data.]{
Results on synthetic problems with $\dimension = 2$ sampled from the GP prior
as in \cref{section:experiments:pesc-synthetic-2-and-8-dim} when using
a resource $r$ that can evaluate 2 tasks ($f$ or $c_1$) in parallel.
(a) Performance comparison of Coupled (black), NCD (green), and CD-F (orange)
approaches.  (b) Cumulative number of task evaluations performed by CD-F when
$c_1$ is active at the solution.  (c) Cumulative number of task evaluations
performed by CD-F when $c_1$ is not active at the solution.
}
\end{figure}

We perform another comparison of the methods Coupled, NCD and CD-F in synthetic
problems in which the objective $f$ and a single constraint function $c_1$ are
drawn from the GP prior with $\dimension=2$. This is done in the same way as in
\cref{section:experiments:pesc-synthetic-2-and-8-dim}. We set $\delta = 0.05$
and follow an experimental protocol similar to the one from the previous
experiment: we assume that there are two tasks, given by $f$ and $c_1$, which
can be evaluated at a resource $r$ with capacity~$\capacity(r)=2$.  Therefore, at
any iteration we will be evaluating 2 tasks in parallel.
\Cref{fig:experiments:decoupling:pesc-synthetic:2d-performance} shows the
median utility gap obtained by each method across 500 different realizations of
the experiment. As in the previous toy problem, CD-F outperforms Coupled, while
Coupled performs similar to NCD. Again, decoupling is useful when we can choose
the tasks to evaluate (CD-F) and evaluating the tasks at potentially different
locations (NCD) does not seem to produce significant improvements with respect
to the coupled approach.

In the previous toy problem, CD-F outperformed NCD and Coupled because it
learned that evaluating the constraint $c_1$ is much more useful than
evaluating the objective $f$ or the constraint~$c_2$. We perform a similar
analysis here by plotting the cumulative number of evaluations for each task
performed by CD-F.  We divide the 500 realizations into those cases in which
the constraint $c_1$ is active at the true solution
(\cref{fig:experiments:decoupling:pesc-synthetic:2d-evals:active}) and those in
which $c_1$ is not active at the true solution
(\cref{fig:experiments:decoupling:pesc-synthetic:2d-evals:not-active}). The
plots in
\cref{fig:experiments:decoupling:pesc-synthetic:2d-evals:active,fig:experiments:decoupling:pesc-synthetic:2d-evals:not-active}
show that when the constraint $c_1$ is active at the solution, CD-F chooses to
evaluate $c_1$ much more frequently. By contrast, when the constraint is not
active, $c_1$ is evaluated much less. Presumably, in the latter case $c_1$ need
only be evaluated until it is determined that it is very unlikely to be active
at the solution. After this point, further evaluations of $c_1$ are not very
informative.  These results indicate that the task-specific acquisition
functions used by PESC and given by \cref{eq:task_specific_acq_function} are
able to successfully measure the usefulness of evaluating each different task.

%\begin{figure}[p]
%\centering
%\subfigure[Constraint active at solution]{
%  \includegraphics[width=0.45\textwidth]{figures/decoupling/synthetic-decoupled-2d-constraint-active.pdf}
%  \label{fig:experiments:decoupling:pesc-synthetic:2d-evals:active}
%}
%\subfigure[Constraint not active at solution]{
%  \includegraphics[width=0.45\textwidth]{figures/decoupling/synthetic-decoupled-2d-constraint-not-active.pdf}
%  \label{fig:experiments:decoupling:pesc-synthetic:2d-evals:not-active}
%}
%\caption[Cumulative function evaluations of decoupled PESC on synthetic data.]{Optimizing samples from the GP prior with decoupled PESC in $\dimension=2$ (\cref{fig:experiments:decoupling-pesc-synthetic-2d-performance}). Cumulative function evaluations of the objective (orange) and constraint (black) when (a) the constraint is active at the true solution (about 30\% of cases) and (b) the constraint is not active at the true solution (about 70\% of cases). Curves reflect the mean over 500 realizations.
%}
%\label{fig:experiments:decoupling:pesc-synthetic:2d-evals}
%\end{figure}

\subsection{Performance of PESC-F with Respect to Wall-clock Time}
\label{section:experiments:wall-time-experiments}

We now evaluate the performance of PESC with fast BO computations (PESC-F,
\cref{section:pesc-fast-updates}) while considering the \walltime of each
experiment. Again, we focus on the toy problem from
\cref{section:experiments:coupled:toy-problem} given by
\cref{eq:experiments:toy-problem}. To highlight what can go wrong in decoupled optimization problems, we will assume that evaluating the objective is
instantaneous, evaluating~$c_1$ takes 2 seconds, and evaluating~$c_2$ takes 1
minute. Each of these functions forms a different task so that all of them can
be evaluated independently. We also consider that there is a single resource
$r$ with $\capacity(r)=1$, that is, only one task can be evaluated at any given
time with no possible parallelism. This setup corresponds to the competitive
decoupling scenario from \cref{fig:decoupling:schematic}.  We limit each
experiment time to 15 minutes and consider the following methods: Coupled, and
competitive decoupled (CD) with PESC-F and rationality levels $\gamma=\{\infty,
1, 0.1, 0\}$. According to \cref{eq:pesc:fast-pesc:rationality-condition},
setting $\gamma=\infty$ is simply another way of saying that fast BO
computations are not used.

\begin{figure}
\centering
\subfigure[Performance, all approaches]{
  \includegraphics[width=0.4\textwidth]{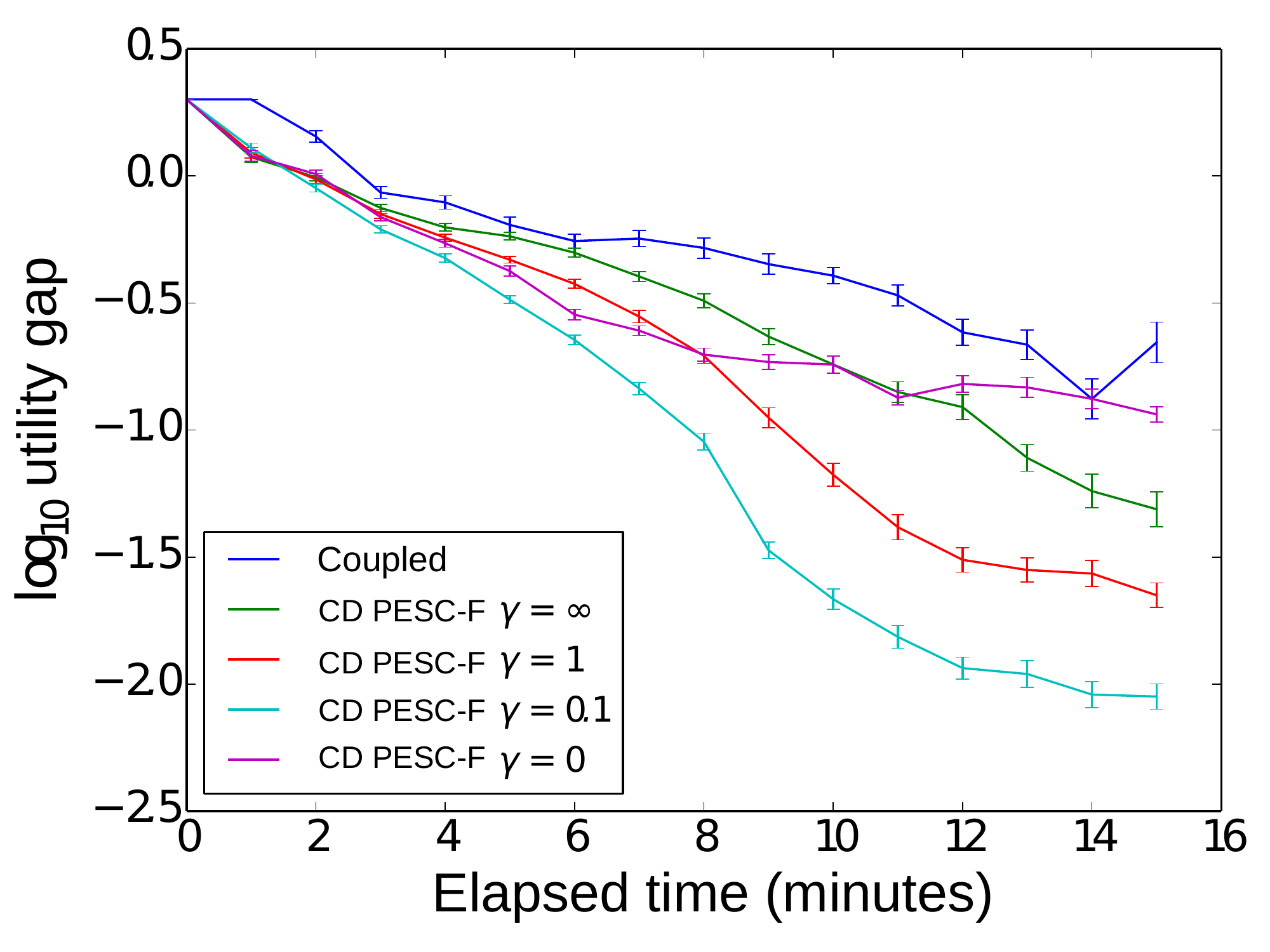}
  \label{fig:experiments:wall-time-toy:performance}
}
\subfigure[Coupled]{
  \includegraphics[width=0.4\textwidth]{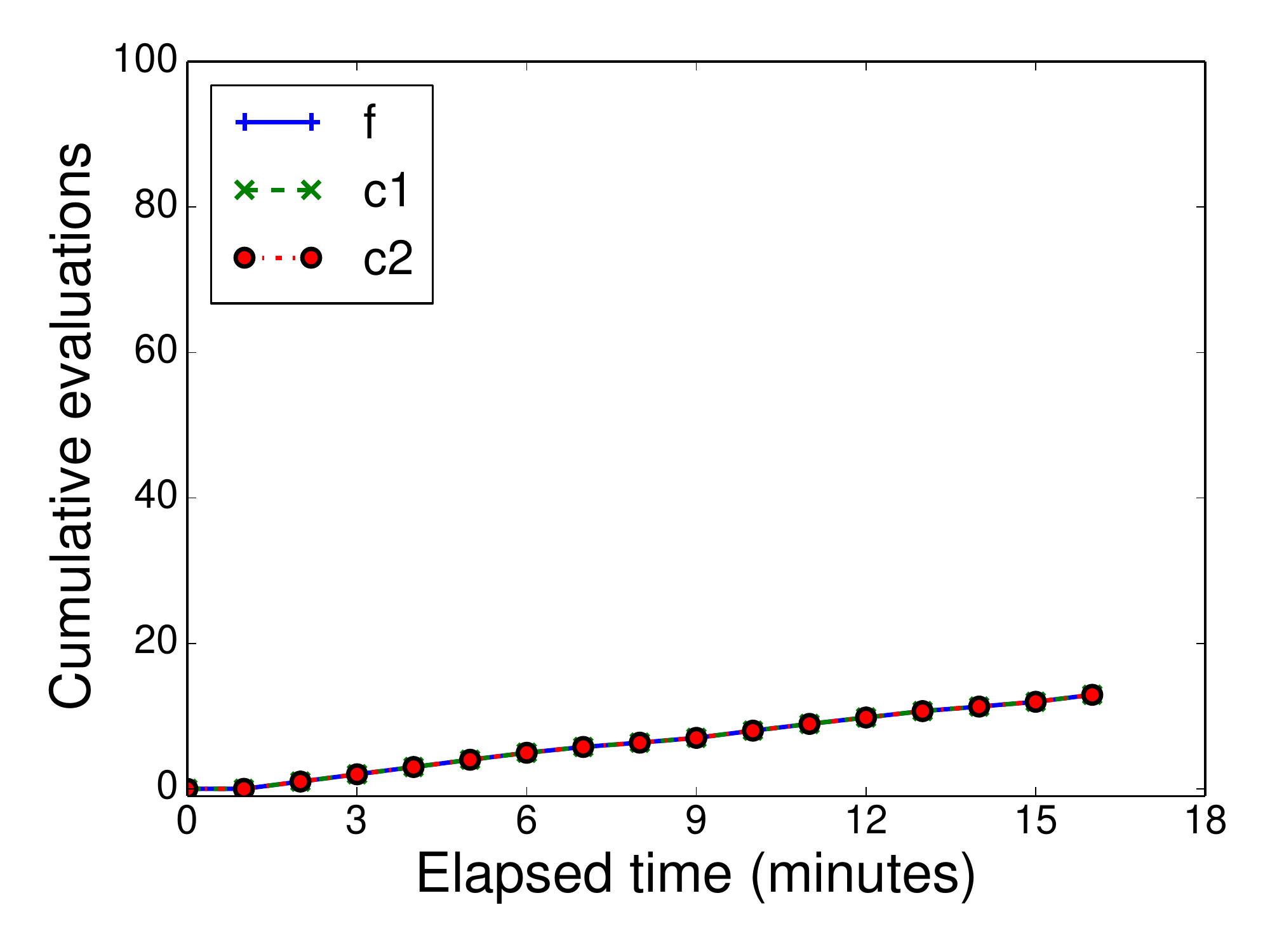}
  \label{fig:experiments:wall-time-toy:evals:coupled}
}\\
\subfigure[CD PESC-F, $\gamma=\infty$]{
  \includegraphics[width=0.4\textwidth]{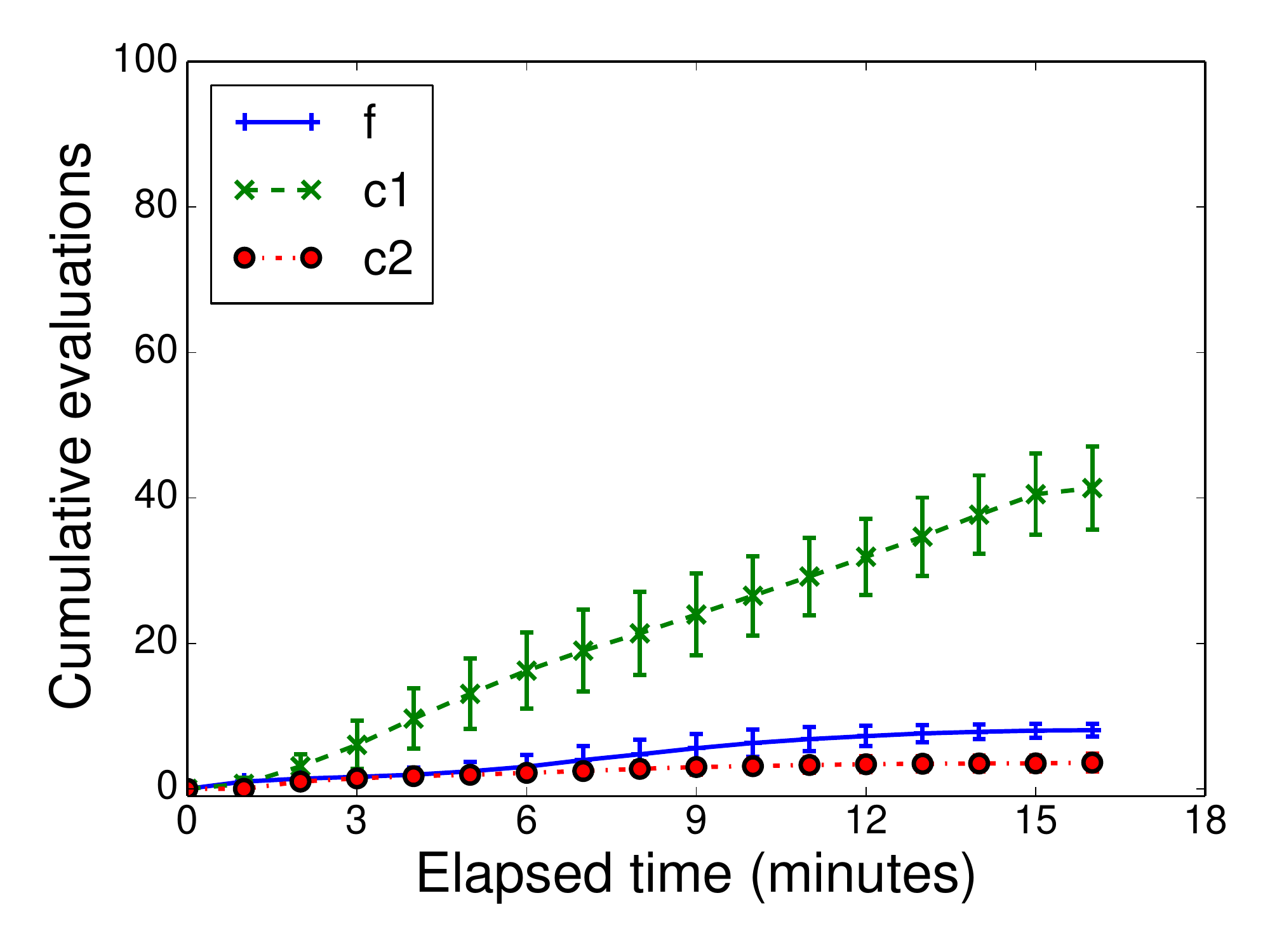}
  \label{fig:experiments:wall-time-toy:evals:slow}
}
\subfigure[CD PESC-F, $\gamma=1$]{
  \includegraphics[width=0.4\textwidth]{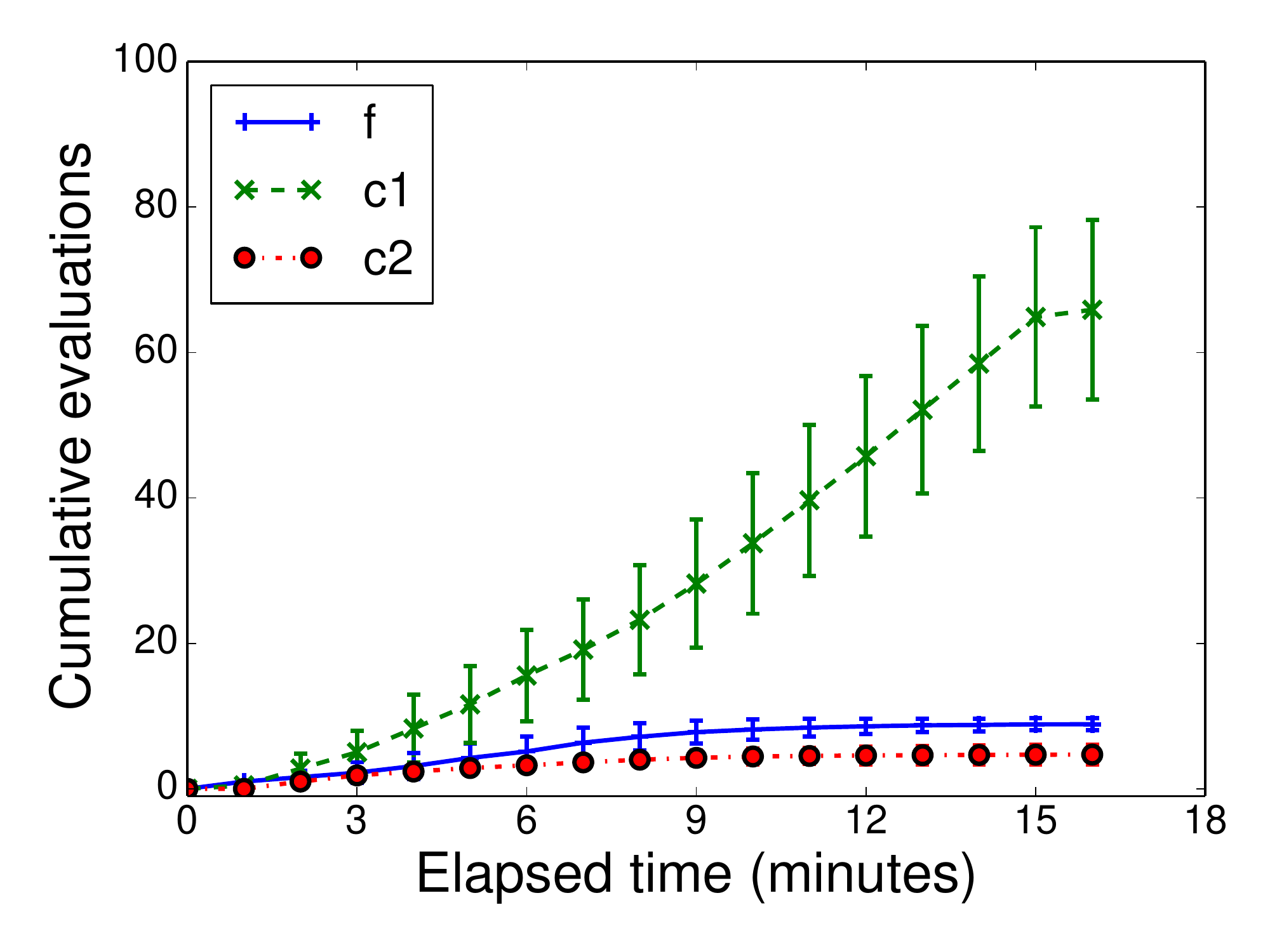}
  \label{fig:experiments:wall-time-toy:evals:fast1}
}\\
\subfigure[CD PESC-F, $\gamma=0.1$]{
  \includegraphics[width=0.4\textwidth]{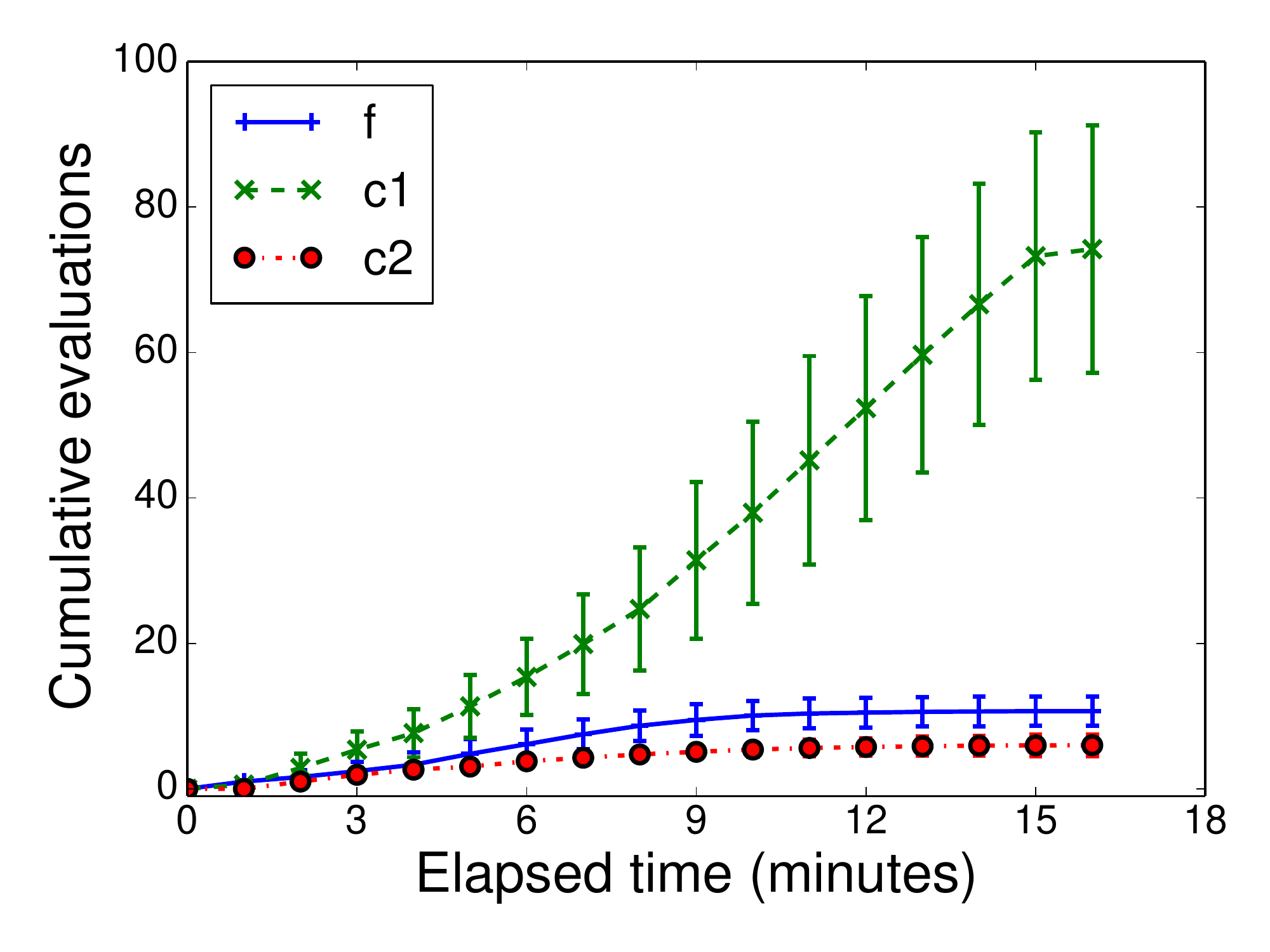}
  \label{fig:experiments:wall-time-toy:evals:fast0.1}
}
\subfigure[CD PESC-F, $\gamma=0$]{
  \includegraphics[width=0.4\textwidth]{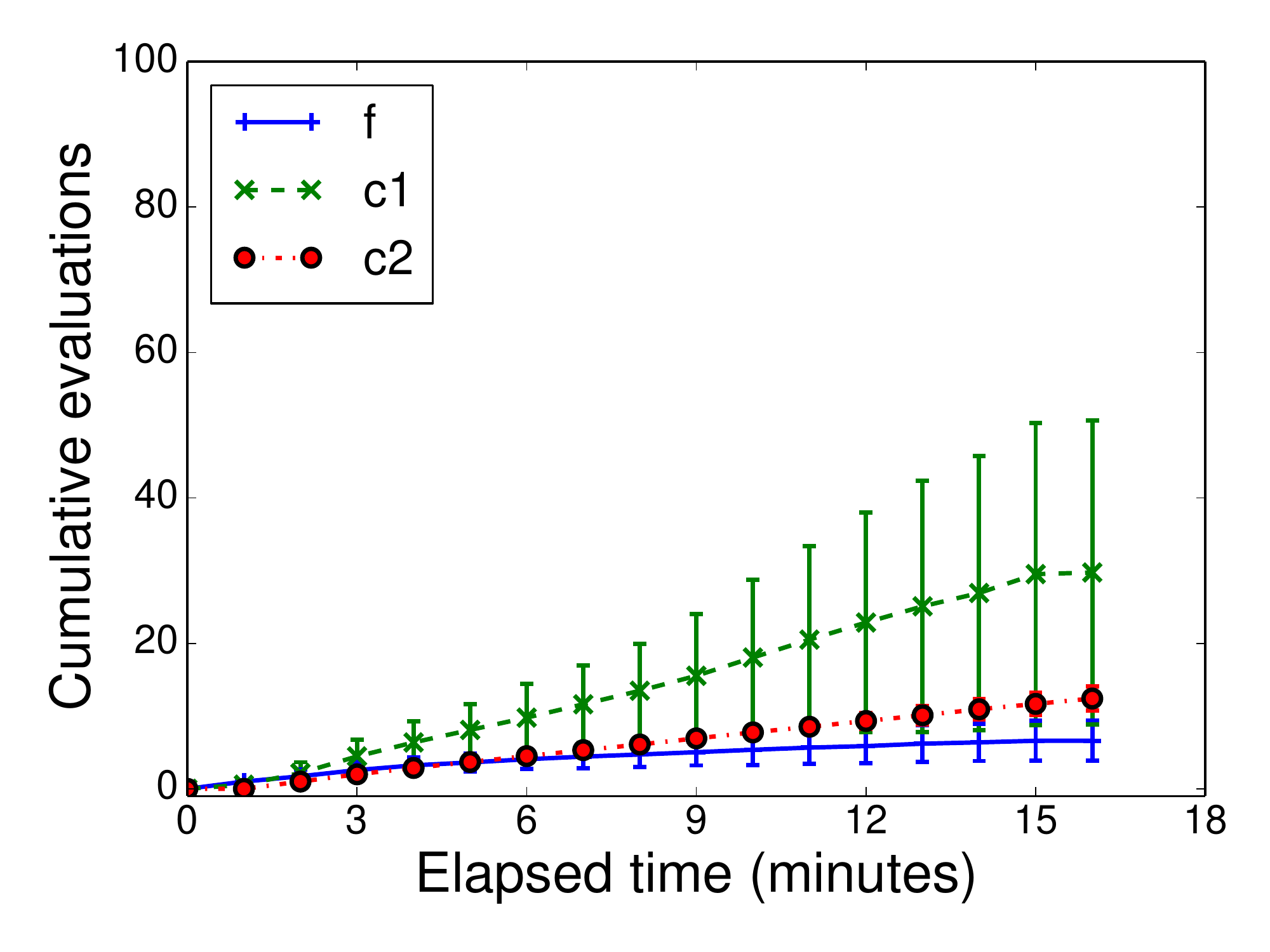}
  \label{fig:experiments:wall-time-toy:evals:fast0}
}
\caption[Evaluating PESC-F with different rationality levels.]{
Results for Coupled and CD PESC-F with $\gamma=\{\infty, 1, 0.1, 0\}$ on the
toy problem given by \cref{eq:experiments:toy-problem}. Evaluations of $f$ are
instantaneous, evaluations of $c_1$ take 2 sec. and evaluations of $c_2$ take 1
min. The maximum experiment time is 15 min. (a)~log utility gap versus
\walltime.  (b-f)~Cumulative function evaluations for (b)~Coupled, (c)~CD
PESC-F with $\gamma=\infty$ (no fast computations), (d)~CD PESC-F with
$\gamma=1.0$, (e)~CD PESC-F with $\gamma=0.1$, and (f) CD PESC-F with
$\gamma=0$ (no slow computations).  Curves reflect the mean over 100 trials.
Error bars in (b-f) are given by the empirical standard deviations.}
\label{fig:experiments:wall-time-toy}
\end{figure}

\Cref{fig:experiments:wall-time-toy:performance} shows the average utility gap
of each method as a function of elapsed time. The coupled approach is the worst
performing one, being outperformed by all the versions of PESC-F with different
$\gamma$. This illustrates the advantages of the decoupled approach.  The
performance of PESC-F is improved as $\gamma$ moves from $\infty$ to $1$ and
then to $0.1$. The reason for this is that, as $\gamma$ is reduced, less time
is spent in the BO computations and more time is spent in the actual collection
of data. However, reducing $\gamma$ too much is detrimental as $\gamma=0$
performs significantly worse than $\gamma = 0.1$ and $\gamma = 1$. The reason
for this is that $\gamma=0$ performs too many fast BO computations, which
produce suboptimal decisions. 

\Cref{fig:experiments:wall-time-toy:evals:coupled,fig:experiments:wall-time-toy:evals:slow,fig:experiments:wall-time-toy:evals:fast1,fig:experiments:wall-time-toy:evals:fast0.1,fig:experiments:wall-time-toy:evals:fast0,table:experiments:toy-wall-time}
are useful to understand the results obtained by the different methods in
\cref{fig:experiments:wall-time-toy:performance}. These figures show, for each
method, the cumulative number of evaluations per task as a function of the
elapsed time. The coupled approach performs very few evaluations of the
different tasks. The reason for this is that it always evaluates all the tasks
the same number of times and this leads to wasting a lot of time by evaluating
too often the slowest task, that is, constraint $c_2$, which is not very
informative about the solution to the problem. The different versions of PESC-F
with $\gamma=\{\infty, 1, 0.1, 0\}$ evaluate more often the most informative
task, that is, $c_1$ and less frequently all the other tasks.  As the
rationality level~$\gamma$ is decreased, less time is spent in the BO
computations, and thus more task evaluations are performed. These correspond to
increases in performance. However, this trend does not continue indefinitely
as~$\gamma$ is decreased. When~$\gamma=0$, performance is significantly
diminished. By not performing slow BO computations, the $\gamma=0$ method is
not able to learn that $c_2$ is uninformative and continues to spend time
evaluating it, thus performing many fewer evaluations of the most informative
task $c_1$. The configuration files for running this experiment are available at
\url{https://github.com/HIPS/Spearmint/tree/PESC/examples/toy-fast-slow}.

\cref{table:experiments:toy-wall-time} shows the time spent by each method in
fast and slow BO computations and in the evaluation of tasks $c_1$ and $c_2$.
We do not include the time spent in the evaluation of task $f$ because it is
always zero. Note that the total time spent in the BO computations and in the
evaluation of the different tasks does not add up exactly to 15 minutes because,
in our implementation, the current iteration is allowed to finish after the
15-minute mark is reached. As expected, the time spent in the BO computations
decreases monotonically as $\gamma$ is decreased. The coupled approach spends a
small amount of time doing BO computations. The reason for this is that this
method does not have to perform step 11 in \cref{algorithm:decoupling:general}
and step 4 is performed less frequently than in the PESC-F methods because
Coupled spends most of its time in the evaluation of $c_2$.

The fifth column in \cref{table:experiments:toy-wall-time} corresponds to the time
spent in the evaluation of $c_1$. The entries in this column are indicative of
the relative performances of each method, since $c_1$ is the most important function in this optimization problem. From these entries we may conclude that this
problem exhibits an optimal
value of $\gamma$ close to $0.1$. This represents an optimal ratio of time
spent in the BO computations to time spent in the evaluation of the different
tasks.  We leave to future work the issue of selecting the optimal value for
$\gamma$. In a highly sophisticated approach this could be done in an online
fashion by using reinforcement learning.

\begin{table}[t!] %[htbp]
\begin{center}
\begin{tabular}{@{}l*{6}{D{.}{.}{-1}}@{}}%{llllllll}
\toprule
\multicolumn{1}{c}{} & \multicolumn{1}{c}{Slow BO} &\multicolumn{1}{c}{Fast BO}  &\multicolumn{1}{c}{\bf Total BO}  &\multicolumn{1}{c}{} &\multicolumn{1}{c}{} &\multicolumn{1}{c}{{\bf Total}}\\
\multicolumn{1}{c}{Method} & \multicolumn{1}{c}{Comp.} &\multicolumn{1}{c}{Comp.}  &\multicolumn{1}{c}{\bf Comp.}  &\multicolumn{1}{c}{$c_1(\x)$} &\multicolumn{1}{c}{$c_2(\x)$} &\multicolumn{1}{c}{\bf Evaluation}\\
% \multicolumn{1}{c}{Method} & \multicolumn{1}{c}{Optimizer (full)} &\multicolumn{1}{c}{Optimizer (fast)}  &\multicolumn{1}{c}{Function evaluations} \\
\midrule
Coupled                    & 2.0  & 0.0 &  2.0  & 0.4 & 13.0 &  13.4 \\  
CD PESC-F, $\gamma=\infty$ & 10.2 & 0.0 &  10.2 & 1.4 & 3.6  &  5.0  \\
CD PESC-F, $\gamma=1$      & 5.2  & 3.0 &  8.2  & 2.2 & 4.7  &  6.9  \\
CD PESC-F, $\gamma=0.1$    & 1.5  & 5.1 &  6.6  & 2.5 & 6.0  &  8.5  \\
CD PESC-F, $\gamma=0.0$    & 0.2  & 1.8 &  2.0  & 1.0 & 12.5 &  13.5\\
\bottomrule
\end{tabular}
\end{center}
\label{table:experiments:toy-wall-time}
\caption[Time allocations of PESC-F with different rationality levels.]{
Time spent by each method in BO computations and in task evaluations.
For each method, the table reports the
mean time in minutes, over 100 independent runs, spent in fast and slow BO computations
and in the evaluation of tasks $c_1$ and $c_2$.}
\end{table}

% \newpage

%!TEX root = main.tex
\section{Conclusions and Future Work}
\label{section:discussion}

We have presented a general framework for solving Bayesian optimization (BO)
problems with unknown constraint functions. In these problems the
objective and the constraints can only be evaluated via expensive queries to
black boxes that may provide noisy values. Our framework allows for problems in
which the objective and the constraints can be split into subsets of functions
that require \emph{coupled} evaluation, meaning that these functions have
always to be jointly evaluated at the same input. We call these subsets of
coupled functions \emph{tasks}. Different tasks may, however, be evaluated
independently at different locations, that is, in a \emph{decoupled} way.
Furthermore, the tasks may or may not compete for a limited set of resources
during their evaluation. Based on this, we have then introduced the notions of
\emph{competitive decoupling} (CD), where two or more tasks compete for the
same resource, and \emph{non-competitive decoupling} (NCD), where the tasks
require to use different resources and can therefore be evaluated in parallel.
The notion of \emph{parallel} BO is a special case in which
one task requires a specific resource, of which many instances are available. We
have then presented a general procedure, given by
\cref{algorithm:decoupling:general}, to solve problems with an arbitrary
combination of coupling and decoupling. This algorithm receives as input a bipartite graph
$\graph$ whose nodes are resources and tasks and whose edges connect each task
with the resource at which it can be evaluated.
\cref{algorithm:decoupling:general} relies on an \acq that can measure the
utility of evaluating any arbitrary subset of functions, that is, of any
possible task. An \acq that satisfies this requirement is said to be
\emph{separable}. 

To implement \cref{algorithm:decoupling:general}, we have proposed a new
information-based approach called Predictive Entropy Search with Constraints
(PESC). At each iteration, PESC collects data at the location that is expected
to provide the highest amount of information about the solution to the
optimization problem.  By introducing a factorization assumption, we obtain an
acquisition function that is additive over the subset of functions to be
evaluated. That is, the amount of information that we approximately gain by
jointly evaluating a set of functions is equal to the sum of the 
gains of information that we approximately obtain by the individual evaluation of each of
the functions. This property means that the \acq of PESC is separable.
Therefore, PESC can be used to solve general constrained BO problems with
decoupled evaluation, something that has not been previously addressed.

We evaluated the performance of PESC in coupled problems, where all the
functions (objective and constraints) are always jointly evaluated at the same
input location. This is the standard setting considered by most prior
approaches to constrained BO. The results of our experiments show that PESC
achieves state-of-the-art results in this scenario. We also evaluated the
performance of PESC in the decoupled setting, where the different tasks can be
evaluated independently at arbitrary input locations. We considered scenarios
with competition (CD) and with non-competition (NCD) and compared the
performances of two versions of PESC: one with decoupling (decoupled PESC) and
another one that always performs coupled evaluations (coupled PESC). Decoupled
PESC is significantly better than coupled PESC when there is competition, that
is, in the CD setting. The reason for this is that some functions can be more
informative than others and decoupled PESC exploits this to make optimal
decisions when deciding which function to evaluate next with limited resources.
In particular, decoupled PESC avoids wasting time in function evaluations that
are unlikely to improve the current estimate of the solution to the
optimization problem.  However, when there is no competition, that is, in the
NCD setting, coupled and decoupled PESC perform similarly. Therefore, in our
experiments, the main advantages of considering decoupling seem to come from
choosing an unequal distribution of tasks to evaluate, rather than from the
additional freedom of evaluating the different tasks at potentially arbitrary
locations. In our experiments we have assumed that the evaluation of all the
functions takes the same amount of time. However, NCD is expected to perform better than
the coupled approach in other settings in which some functions are 
much faster to evaluate than others. Evaluating the performance of NCD in these
settings is left as future work.

For the BO approach to be useful, the time spent performing BO computations
(such as computing and globally optimizing the acquisition function) has to be
significantly shorter than the time spent collecting data. However, decoupled
optimization problems may include some tasks that are fast to evaluate. When
these tasks are informative and their evaluation time is comparable to that of
the BO computations, the BO approach may be inefficient. To address this
issue, we follow a bounded rationality approach and introduce additional
approximations in the computations made by PESC to reduce their cost when
necessary. The new method is called PESC-F and it is able to automatically
switch between fast, but approximate, and slow, but accurate, operations.  A
parameter called the \emph{rationality level} is used in PESC-F to balance the
amount of time that is spent in the BO computations and in the actual
collection of data. Experiments with wall-clock time in a CD scenario show that
PESC-F can be significantly better than the original version of PESC.

In summary, PESC is an effective algorithm for BO problems with unknown
constraints and the separability of its acquisition function makes it a
promising direction towards a unified solution for constrained BO problems. As
new \acqs are proposed in the future, they will hopefully be developed with
separability in mind as an important and desirable property. 

The code for PESC, including decoupling and PESC-F, is available in PESC branch
of the open-source Bayesian optimization package \emph{Spearmint} at
\url{https://github.com/HIPS/Spearmint/tree/PESC}.

%Potential lines of future work include allowing each task to require more than
%one single resource to run, extensions of PESC to handle multiple objectives,
%allowing for more than two levels of computation in PESC-F, that is, fast and
%slow computations, and a theoretical analysis of the performance of
%information-based approaches such as PES and PESC.

%!TEX root = main.tex
%\section{Future work}

One potential line of future work includes extensions to settings where tasks require more
than one resource to run. This could, for example, be formalized using
a framework similar to the one presented in \cref{section:decoupling}, but where the resource
dependencies for each task $t$ are represented as a set of edges $\mathcal
E_{ti} = \{t\sim r\}$ for the $i$th potential allocation of resources. This can
be interpreted as the statement that all resource nodes $\mathcal V_{ti}=\{r:
(t\sim r)\in \mathcal E_{ti} \}$ are required in order to initiate task $t$
using allocation $i$. Note that the union of these edges now specifies a
multigraph with edges $\mathcal E=\bigcup_{t,i} \mathcal E_{ti}$ due to the
fact there may be resources that are required across multiple allocations of a
particular task. This also modifies the pseudocode for
\Cref{algorithm:decoupling:general} where the loop over resources $r$ becomes a
loop over potential allocations such that $\omega(r)<\omega_\text{max}(r)$ for
$r\in\mathcal V_{ti}$ for some $(t,i)$ pair. In the case where each
task requires only a single resource this reduces to the earlier
formulation. Another possibility is for allocations where the resources are
time or iteration dependent. This would require some form of temporal planning.
To make such a procedure feasible, however, it may be necessary to
consider greedy decisions at each point in time. 

Another direction for future work is concerned with the use of \emph{bounded
rationality} in Bayesian optimization. Here we have used a simple heuristic for
selecting between two levels (fast and slow) of computations in PESC-F.
However, we could consider a larger number of levels with increasingly more
accurate computations \citep{hay:2012}. The Bayesian optimization algorithm
would then have to optimally select one of these to determine the next
evaluation location. We could also consider different modeling approaches for
the collected data, with different trade-offs between accuracy and computational cost.
We also leave for future work a theoretical analysis of PESC. This would be in the
line of the work of  \cite{russo:2014c} on {information-directed} sampling.
However, they use simpler models for the data and do not consider problems with
constraints.

Finally, we would like to point out that the approach described here can be
applied in a straightforward manner to address multi-objective Bayesian
optimization problems \citep{knowles2006parego}. In the multi-objective case the
different tasks would be given by groups of objective functions that have to
be evaluated in a coupled manner. An extension of PES for working with multiple
objectives is given by \cite{hernandez2015predictive}.

% Acknowledgements should go at the end, before appendices and references

\acks{Jos\'e Miguel Hern\'andez-Lobato acknowledges support from the Rafael del Pino Foundation.
Zoubin Ghahramani acknowledges support from Google Focused Research Award and EPSRC grant EP/I036575/1.
Matthew W. Hoffman acknowledges support from EPSRC grant EP/J012300/1.}

% Manual newpage inserted to improve layout of file - not

\appendix
%!TEX root = main.tex

\newcommand\scalemath[2]{\scalebox{#1}{\mbox{\ensuremath{\displaystyle #2}}}}

\section{The Expectation Propagation Method Used by PESC}\label{appendix:implementation}

We describe here the expectation propagation (EP) method that is used by PESC to
adjust a Gaussian approximation to the non-Gaussian factor
$f(\mathbf{f},\mathbf{c}_1,\ldots,\mathbf{c}_K|\xopt^j)$ in
\cref{eq:conditioned_predictive_distribution}. This is done after replacing the
infinite set $\mathcal{X}$ with the finite set $\mathcal{Z}$, which contains
only the locations at which the objective $f$ has been evaluated so far, the
value of $\xopt^j$ and $\mathbf{x}$. Recall that $\mathbf{x}$ is
the input to the acquisition function, that is, it contains the location at
which we are planning to evaluate $f,c_1,\ldots,c_K$. When $\mathcal{X}$ is replaced with $\mathcal{Z}$ we have that the
vectors $\mathbf{f},\mathbf{c}_1,\ldots,\mathbf{c}_K$ contain now the result of the
noise-free evaluations of $f,c_1,\ldots,c_K$ at $\mathcal{Z}$, that is,
\begin{align}
\mathbf{f} &= [f(\x_f^1),\ldots,f(\x_f^{N_1}), f(\xopt^j), f(\mathbf{x})]^\text{T}\,, \label{eq:definition_of_f_vector} \\
\mathbf{c}_k &= [c_k(\x_f^1),\ldots,c_k(\x_f^{N_1}),c_k(\xopt^j),c_k(\mathbf{x})]^\text{T}\,,\quad \text{for }\quad k=1,\ldots,K\,.
\label{eq:definition_of_c_k}
\end{align}
where $\x_f^1,\ldots,\x_f^{N_1}$ are the locations at which the objective $f$ has
been evaluated so far. That is, the first $N_1$ entries in $\mathbf{f},\mathbf{c}_1,\ldots,\mathbf{c}_K$
contain the function values at the locations for which there is data for
the objective. These entries are then followed by the function values at $\xopt^j$ and at $\mathbf{x}$.
When we replace $\mathcal{X}$ with $\mathcal{Z}$ we have that
\begin{align}
f(\mathbf{f},\mathbf{c}_1,\ldots,\mathbf{c}_K|\xopt^j) = p(\mathbf{f},\mathbf{c}_1,\ldots,\mathbf{c}_K|\mathcal{D}) \Gamma(\xopt^j)
\left\{ \prod_{i=1}^{N_1} \Psi(\mathbf{x}_f^i) \right\}
\Psi(\mathbf{x})\,.\label{eq:definition_of_f}
\end{align}
In this expression we should have included a factor $\Psi(\xopt^j)$ since
$\xopt^j \in \mathcal{Z}$. We ignore such factor because it is always equal to
1 according to \cref{eq:indicator_psi}. In \cref{eq:definition_of_f} we have
separated the non-Gaussian factor that depends on $\mathbf{x}$, that is,
$\Psi(\mathbf{x})$ from those factors that do not depend on $\mathbf{x}$, that
is, $\Gamma(\xopt^j), \Psi(\mathbf{x}_f^1),\ldots,\Psi(\mathbf{x}_f^{N_1})$.
All these non-Gaussian factors are approximated with Gaussians using EP.  

Finding the next \emph{suggestion} involves maximizing the acquisition
function. This requires to evaluate the acquisition function at many different
$\mathbf{x}$ and recomputing the complete EP approximation for each value of
$\mathbf{x}$ can be excessively expensive.  To avoid this, we first
compute the EP approximation for the factors that do not depend on $\mathbf{x}$
in isolation, store it and then reuse it as we compute the EP approximation for
the remaining factors.  Since most of the factors do not depend on
$\mathbf{x}$, this leads to large speedups when we have to evaluate the
acquisition function at many different $\mathbf{x}$. Therefore, we start by
finding a Gaussian approximation to the factors that do not depend on
$\mathbf{x}$, that is, $\Gamma(\xopt^j),
\Psi(\mathbf{x}_f^1),\ldots,\Psi(\mathbf{x}_f^{N_1})$, while ignoring the other
factor that does depend on $\mathbf{x}$, that is, $\Psi(\mathbf{x})$.

\subsection{Approximating the Non-Gaussian Factors that do not Depend on $\mathbf{x}$}

We use EP to find a Gaussian approximation to $\Gamma(\xopt^j),
\Psi(\mathbf{x}_f^1),\ldots,\Psi(\mathbf{x}_f^{N_1})$ in \cref{eq:definition_of_f}
when $\Psi(\mathbf{x}_1),\ldots,\Psi(\mathbf{x}_{K+1})$ are assumed to be
constant and equal to 1.  Because the data is assumed to be generated from
independent GPs, we have that
$p(\mathbf{f},\mathbf{c}_1,\ldots,\mathbf{c}_K|\mathcal{D})$ in
\cref{eq:definition_of_f} is
\begin{align}
p(\mathbf{f},\mathbf{c}_1,\ldots,\mathbf{c}_K|\mathcal{D}) =
\gaussian(\mathbf{f}\given\mathbf{m}_1^\text{pred},\mathbf{V}_1^\text{pred})
\prod_{k=1}^{K}\left\{\gaussian(\mathbf{c}_k\given\mathbf{m}_{k+1}^\text{pred},\mathbf{V}_{k+1}^\text{pred})\right\}\,,
\end{align}
where $\mathbf{m}^\text{pred}_1$ and $\mathbf{V}^\text{pred}_1$ are the
mean and covariance matrix of the posterior distribution of $\mathbf{f}$ given
the data for the objective and
$\mathbf{m}^\text{pred}_{k+1}$ and $\mathbf{V}^\text{pred}_{k+1}$ are the
mean and covariance matrix of the posterior distribution of $\mathbf{c}_k$ given
the data for constraint $k$.
In particular, from Eqs. (2.22) to (2.24) of
\citep{Rasmussen2006} we have that
\begin{align}
\mathbf{m}^\text{pred}_i & = \mathbf{K}_{\star}^{i}(\mathbf{K}^i+\nu_{i}\identity)^{-1}\mathbf{y}^{i}\,,\\
\mathbf{V}^\text{pred}_i & = 
\mathbf{K}_{\star,\star}^{i}-\mathbf{K}_{\star}^{i}(\mathbf{K}^{i}+\nu_{i}\identity)^{-1}[\mathbf{K}_{\star}^i]\transpose\,,
\quad \text{for}\quad i=1,\ldots,K+1\,,
\end{align}
where $\mathbf{y}^i$ is an $N_i$-dimensional vector with the data for the
$i$-th function in $\{f,c_1,\ldots,c_{K}\}$, $\mathbf{K}_{\star}^{i}$ is an
$(N_1+2)\times N_i$ matrix with the prior cross-covariances between the entries
of the $i$-th vector in $\{\mathbf{f},\mathbf{c}_1,\ldots,\mathbf{c}_K\}$ and
the value of the corresponding function at the locations for which there is
data available for that function and $\mathbf{K}_{\star,\star}^{i}$ is an
$(N_1+2)\times(N_1+2)$ matrix with the prior covariances between entries of the
$i$-th vector in  $\{\mathbf{f},\mathbf{c}_1,\ldots,\mathbf{c}_K\}$ and $\nu_i$
is the noise variance at the black-box for the $i$-th function in
$\{f,c_1,\ldots,c_K\}$.

The exact factors $\Gamma(\xopt^j),
\Psi(\mathbf{x}_f^1),\ldots,\Psi(\mathbf{x}_f^{N_1})$ from
\cref{eq:definition_of_f} are then approximated with the corresponding Gaussian
factors $\widetilde{\Gamma}(\xopt^j),
\widetilde{\Psi}(\mathbf{x}_f^1),\ldots,\widetilde{\Psi}(\mathbf{x}_f^{N_1})$.  Let $\bm
\beta_n(\mathbf{f})=[f(\x_f^n), f(\xopt^j) ]^\text{T}$, where $\x_f^n$ is the
$n$-th location for which there is data for the objective $f$.
Then, we define 
\begin{equation} 
\scalemath{0.975}{
\widetilde{\Psi}(\mathbf{x}_f^n)\propto\exp\left\{-\frac{1}{2}\bm \beta_n(\mathbf{f})^\text{T} \widetilde{\mathbf{A}}_n\bm \beta_n(\mathbf{f})+
\bm \beta_n(\mathbf{f})^\text{T}\widetilde{\mathbf{b}}_n\right\} \prod_{k=1}^K 
\exp\left\{-\frac{1}{2}c_k(\x_f^n)^2 \widetilde{d}_n^k +c_k(\x_f^n) \widetilde{e}_n^k \right\},\label{eq:psi_tilde}
}\end{equation}
where $\widetilde{\mathbf{A}}_n$ and $\widetilde{\mathbf{b}}_n$ are the natural
parameters of a bivariate Gaussian distribution on $\bm \beta_n(\mathbf{f})$
and $\widetilde{d}_n^k$ and $\widetilde{e}_n^k$ are the natural parameters of a
Gaussian distribution on $c_k(\x_f^n)$, that is, the value of constraint $k$ at
the $n$-th location for which there is data for the objective. We also define
\begin{equation} 
\widetilde{\Gamma}(\xopt^j) \propto  \prod_{k=1}^K \exp\left\{-\frac{1}{2}c_k(\xopt^j)^2 \widetilde{g}_k +c_k(\xopt^j)
\widetilde{h}_k \right\}\,,\nonumber\label{eq:gamma_tilde}
\end{equation}
where $\widetilde{g}_k$ and $\widetilde{h}_k$ are the natural parameters of a Gaussian distribution
on $c_k(\xopt^j)$, that is, the value of constraint $k$ at the current
posterior sample of $\xopt$.

The parameters $\widetilde{\mathbf{A}}_n$, $\widetilde{\mathbf{b}}_n$, $\widetilde{d}_n^k$,
$\widetilde{e}_n^k$, $\widetilde{g}_k$ and $\widetilde{h}_k$ are fixed by running EP.
Once the value of these parameters has been fixed, we replace the exact factors
$\Gamma(\xopt^j), \Psi(\mathbf{x}_f^1),\ldots,\Psi(\mathbf{x}_f^{N_1})$ in
\cref{eq:definition_of_f} with their corresponding Gaussian approximations to obtain an
approximation to
$f(\mathbf{f},\mathbf{c}_1,\ldots,\mathbf{c}_K|\xopt^j)$. We denote this
approximation by $q(\mathbf{f},\mathbf{c}_1,\ldots,\mathbf{c}_K)$, where
\begin{equation}
q(\mathbf{f},\mathbf{c}_1,\ldots,\mathbf{c}_K) \propto p(\mathbf{f},\mathbf{c}_1,\ldots,\mathbf{c}_K|\mathcal{D}) \widetilde{\Gamma}(\xopt^j)
\left\{ \prod_{i=1}^{N_1} \widetilde{\Psi}(\mathbf{x}_f^i) \right\}
\left\{ \prod_{k=1}^{K+1} \widetilde{\Psi}(\mathbf{x}_k) \right\}\,.
\end{equation}
Since the approximate factors are Gaussian and $p(\mathbf{f},\mathbf{c}_1,\ldots,\mathbf{c}_K|\mathcal{D})$
is also Gaussian, we have that $q(\mathbf{f},\mathbf{c}_1,\ldots,\mathbf{c}_K)$ is also Gaussian:
\begin{align}
q(\mathbf{f},\mathbf{c}_1,\ldots,\mathbf{c}_K) =
\mathcal{N}(\mathbf{f}|\mathbf{m}_1,\mathbf{V}_1) \prod_{k=1}^{K}
\mathcal{N}(\mathbf{c}_k|\mathbf{m}_{k+1},\mathbf{V}_{k+1})\,,\label{eq:definition_of_q_appendix}
\end{align}
where, by applying the formula for products of Gaussians, we obtain
\begin{align}
\mathbf{V}_i & = \left[ [\mathbf{V}^{\text{pred}}_i]^{-1}+\widetilde{\mathbf{S}}_i\right]^{-1}\,,\\
\mathbf{m}_i & = \mathbf{V}_i\left[[\mathbf{V}^{\text{pred}}_i]^{-1}\mathbf{m}^\text{pred}_i + \widetilde{\mathbf{t}}_i\right]\,, 
\quad \text{for}\quad i=1,\ldots,K+1\,,
\end{align}
with the following definitions for $\widetilde{\mathbf{S}}_i$ and $\widetilde{\mathbf{t}}_i$:
\begin{itemize}
\item $\widetilde{\mathbf{S}}_1$ is an $(N_1+2)\times(N_1+2)$ matrix whose non-zero entries are
\begin{itemize}
\item $[\widetilde{\mathbf{S}}_1]_{n,n}=\left[\mathbf{A}_n\right]_{1,1}$ for $n=1,\ldots,N_1\,$,
\item $[\widetilde{\mathbf{S}}_1]_{N_1+1,n}=[\widetilde{\mathbf{S}}_1]_{n,N_1+1}=\left[\mathbf{A}_n\right]_{1,2}$ for $n=1,\ldots,N_1\,$,
\item $[\widetilde{\mathbf{S}}_1]_{N+1,N+1}=\sum_{n=1}^{N_1}\left[\mathbf{A}_n\right]_{2,2}\,$.
\end{itemize}
\item $\widetilde{\mathbf{S}}_{k+1}$, for $k=1,\ldots,K$, is an $(N_1+2)\times(N_1+2)$ matrix whose non-zero entries are
\begin{itemize}
\item $[\widetilde{\mathbf{S}}_{k+1}]_{n,n}=d_n$ for $n=1,\ldots,N_1\,$,
\item $[\widetilde{\mathbf{S}}_{k+1}]_{N_1+1,N_1+1}=g_n$ for $n=1,\ldots,N_1\,$.
\end{itemize}
\item $\widetilde{\mathbf{t}}_1$ is an $(N_1+2)$-dimensional vector whose non-zero entries are
\begin{itemize}
\item $[\widetilde{\mathbf{t}}_1]_n=[\widetilde{\mathbf{b}}_n]_1$ for $n=1,\ldots,N_1\,$,
\item $[\widetilde{\mathbf{t}}_1]_{N_1+1}=\sum_{n=1}^{N_1}[\widetilde{\mathbf{b}}_{n}]_2\,$.
\end{itemize}
\item $\widetilde{\mathbf{t}}_{k+1}$, for $k=1,\ldots,K$, is an $(N_1+2)$-dimensional vector whose non-zero entries are
\begin{itemize}
\item $[\widetilde{\mathbf{t}}_{k+1}]_n=\widetilde{e}_n^k$ for $n=1,\ldots,N_1\,$,
\item $[\widetilde{\mathbf{t}}_{k+1}]_{N_1+1}=\widetilde{h}_k\,$.
\end{itemize}
\end{itemize}
We now explain how to obtain the values of all the $\widetilde{\mathbf{A}}_n$,
$\widetilde{\mathbf{b}}_n$, $\widetilde{d}_n^k$, $\widetilde{e}_n^k$,
$\widetilde{g}_k$ and $\widetilde{h}_k$ using EP. 

\subsubsection{Adjusting $\widetilde{\Psi}(\mathbf{x}_f^n)$  by EP} \label{appendix:section_ep_part_1}

We explain how to adjust the parameters $\widetilde{\mathbf{A}}_n$,
$\widetilde{\mathbf{b}}_n$, $\widetilde{d}_n^k$ and $\widetilde{e}_n^k$ of the
approximate factor $\widetilde{\Psi}(\mathbf{x}_f^n)$ using EP. EP performs
this operation by minimizing the following Kullback-Leibler divergence:
\begin{align}
\text{KL}[\Psi(\mathbf{x}_f^n)q^{\neg n}(\mathbf{f},\mathbf{c}_1,\ldots,\mathbf{c}_K)||
\widetilde{\Psi}(\mathbf{x}_f^n)q^{\neg n}(\mathbf{f},\mathbf{c}_1,\ldots,\mathbf{c}_K)]\,,\label{eq:KL_divergence}
\end{align}
where $q^{\neg n}(\mathbf{f},\mathbf{c}_1,\ldots,\mathbf{c}_K)$ is the
cavity distribution given by
\begin{align}
q^{\neg n}(\mathbf{f},\mathbf{c}_1,\ldots,\mathbf{c}_K) = 
q(\mathbf{f},\mathbf{c}_1,\ldots,\mathbf{c}_K)[\widetilde{\Psi}(\mathbf{x}_f^n)]^{-1}\,,
\end{align}
If we marginalize out all variables except those which
$\widetilde{\Psi}(\mathbf{x}_f^n)$ depends on, namely
$\bm \beta_n(\mathbf{f})$ and $c_1(\mathbf{x}_f^n),\ldots,c_K(\mathbf{x}_f^n)$,
then $q^{\neg n}$ takes the form
\begin{align}
q^{\neg n}[\bm \beta_n(\mathbf{f}),c_1(\mathbf{x}_f^n),\ldots,c_K(\mathbf{x}_f^n)] \propto 
\gaussian(\bm\beta_n(\mathbf{f})\given\mathbf{b}^{\neg n},\mathbf{A}^{\neg n})
\left\{\prod_{k=1}^K \gaussian(c_k(\mathbf{x}_f^n)\given e^{\neg n}_k, d^{\neg n}_k)\right\} \, ,
\end{align}
where the parameters $\mathbf{b}^{\neg n}$, $\mathbf{A}^{\neg n}$, $e^{\neg n}_k$ and $d^{\neg n}_k$ of these Gaussian distributions
are obtained from the ratio of
$q$
and
$\widetilde{\Psi}(\mathbf{x}_f^n)$
by using the formula for dividing Gaussians:
\begin{align}
\mathbf{A}^{\neg n} & = \left\{ \mathbf{V}_{\bm \beta_n(\mathbf{f})}^{-1} - \widetilde{\mathbf{A}}_n \right\}^{-1}\,, &
\mathbf{b}^{\neg n} & = \mathbf{A}^{\neg n} \left\{ \mathbf{V}_{\bm \beta_n(\mathbf{f})}^{-1} \mathbf{m}_{\bm \beta_n(\mathbf{f})}
 - \widetilde{\mathbf{b}}_n \right\}\,,\\
d^{\neg n}_k & = \left\{ v_{c_k(\mathbf{x}_f^n)}^{-1} - \widetilde{d}_k \right\}^{-1}\,, &
e^{\neg n}_k & = d^{\neg n}_k \left\{ v_{c_k(\mathbf{x}_f^n)}^{-1} m_{c_k(\mathbf{x}_f^n)} - \widetilde{e}_k \right\}^{-1}\,,
\end{align}
where $\mathbf{V}_{\bm \beta_n(\mathbf{f})}$ is the $2\times2$ covariance matrix for $\bm
\beta_n(\mathbf{f})$ given by $q(\mathbf{f},\mathbf{c}_1,\ldots,\mathbf{c}_K)$ in
\cref{eq:definition_of_q_appendix},
$\mathbf{m}_{\bm \beta_n(\mathbf{f})}$ is the corresponding $2$-dimensional mean vector,
$v_{c_k(\mathbf{x}_f^n)}$ is the variance for $c_k(\mathbf{x}_f^n)$
given by $q(\mathbf{f},\mathbf{c}_1,\ldots,\mathbf{c}_K)$ in
\cref{eq:definition_of_q_appendix}
and $m_{c_k(\mathbf{x}_f^n)}$ is the corresponding mean parameter.

To minimize \cref{eq:KL_divergence} we match the 1st and 2nd moments of
$\Psi(\mathbf{x}_f^n)q^{\neg n}(\mathbf{f},\mathbf{c}_1,\ldots,\mathbf{c}_K)$
and $\widetilde{\Psi}(\mathbf{x}_f^n)q^{\neg
n}(\mathbf{f},\mathbf{c}_1,\ldots,\mathbf{c}_K)]$. The moments of
$\Psi(\mathbf{x}_f^n)q^{\neg n}(\mathbf{f},\mathbf{c}_1,\ldots,\mathbf{c}_K)$
can be obtained from the derivatives of the logarithm of its normalization constant $Z$, which is given by
\begin{equation}
\scalemath{0.975}{
Z = 
\int \Psi(\mathbf{x}_f^n)q^{\neg n}(\mathbf{f},\mathbf{c}_1,\ldots,\mathbf{c}_K)\,d\mathbf{f}\,d\mathbf{c}_1\,\cdots d\mathbf{c}_K
 = \Phi(\alpha_n)\prod_{k=1}^K\Phi\left[\alpha_n^k\right]+ 1-\prod_{k=1}^K\Phi\left[\alpha_n^k\right]\,,
}
\end{equation}
where $\alpha_n^k=m_{c_k(\mathbf{x}_f^n)}v_{c_k(\mathbf{x}_f^n)}^{-1/2}$
and $\alpha_n=[1,\,-1]\mathbf{m}_{\bm \beta_n(\mathbf{f})}([1,\;-1]\mathbf{V}_{\bm \beta_n(\mathbf{f})}[1,\,-1]\transpose)^{-1/2}$
and $\Phi$ is the standard Gaussian cdf.
We follow Eqs. (5.12) and (5.13) in \citep{minka2001family} to update $\widetilde{d}_n^k$ and
$\widetilde{e}_n^k$ in \cref{eq:psi_tilde}. 
However, we use the second
partial derivative with respect to $e^{\neg n}_k$ rather than first partial
derivative with respect to $d^{\neg n}_k$ for numerical robustness. These
derivatives are given by
\begin{align}
\frac{\partial \log Z}{\partial e^{\neg n}_k} & =
\frac{(Z-1)\phi(\alpha_n^k)}{Z\Phi(\alpha_n^k)\sqrt{d^{\neg n}_k}}\,, &
\frac{\partial^2\log Z}{\partial[e^{\neg n}_k]^{2}} & =
-\frac{\partial \log Z}{\partial e^{\neg n}_k} \cdot \frac{\alpha_n^k}{\sqrt{d^{\neg n}_k}}
- \left[\frac{\partial \log Z}{\partial e^{\neg n}_k}\right]^2\,,
\end{align}
where $\phi$ is the standard Gaussian pdf. The update equations for the
parameters $\widetilde{d}_n^k$ and $\widetilde{e}_n^k$ of the approximate
factor $\widetilde{\Psi}(\mathbf{x}_f^n)$ are then 
\begin{align}
\scalemath{0.875}{
[\widetilde{d}_n^k]_\text{new} = 
-\left\{\left(\frac{\partial^2 \log Z}{\partial [e^{\neg n}_k ]^2}\right)^{-1} + d^{\neg n}_k \right\}^{-1}\,,\quad
[\widetilde{e}_n^k]_\text{new} = \left\{ d^{\neg n}_k -\left[\frac{\partial^2\log Z}{\partial[e^{\neg n}_k]^{2}}\right]^{-1}
\frac{\partial\log Z}{\partial e^{\neg n}_k}\right\} [\widetilde{d}_n^k]^\text{new}\,,
}
\label{eq:pesc-supplement:bhat-h-thingy}
\end{align}
We now perform the analogous operations to update $\widetilde{\mathbf{A}}_n$
and $\widetilde{\mathbf{b}}_n$. We need to compute
\begin{align}
\frac{\partial\log Z}{\partial \mathbf{b}^{\neg n}} & = 
\frac{\left\{ \prod_{k=1}^K\Phi[\alpha_n^k]\right\} \phi(\alpha_n)}{Z\sqrt{s}}[1,\,-1]\,,\\
\frac{\partial\log Z}{\partial \mathbf{A}^{\neg n}} & = -\frac{1}{2}[1,\,-1]\transpose[1,\,-1]\frac{\left\{ 
\prod_{k=1}^K\Phi[\alpha_n^k]\right\} \phi(\alpha_n)\alpha_n}{Z s}\,,
\end{align}
where $s=[-1,\,1]\mathbf{A}^{\neg n}[-1,\,1]\transpose$.
We then compute the mean vector and covariance matrix for $\bm \beta_n(\mathbf{f})$
with respect to $\Psi(\mathbf{x}_f^n)q^{\neg n}(\mathbf{f},\mathbf{c}_1,\ldots,\mathbf{c}_K)$:
\begin{align}
[\mathbf{V}_{\bm \beta_n(\mathbf{f})}]_\text{new} & = 
\mathbf{A}^{\neg n} -
\mathbf{A}^{\neg n}
\left[\frac{\partial\log Z}{\partial \mathbf{b}^{\neg n}}\left(\frac{\partial\log Z}{\partial \mathbf{b}^{\neg n}}\right)\transpose-
2\frac{\partial\log Z}{\partial \mathbf{A}^{\neg n}}\right] \mathbf{A}^{\neg n}\,,\label{eq:parameters_new_q_1}\\
[\mathbf{m}_{\bm \beta_n(\mathbf{f})}]_\text{new} & = 
\mathbf{b}^{\neg n} + \mathbf{A}^{\neg n}\frac{\partial\log Z}{\partial \mathbf{b}^{\neg n}}\,.\label{eq:parameters_new_q_2}
\end{align}
Next, we divide the Gaussian with mean and covariance parameters
given by \cref{eq:parameters_new_q_1,eq:parameters_new_q_2}
by the marginal for $\bm \beta(\mathbf{f})$ in the cavity distribution
$q^{\neg n}(\mathbf{f},\mathbf{c}_1,\ldots,\mathbf{c}_K)$. Therefore, the new
parameters $\widetilde{\mathbf{A}}_n$ and $\widetilde{\mathbf{b}}_n$
of the approximate factor 
$\widetilde{\Psi}(\mathbf{x}_f^n)$
are obtained using the formula for the ratio of two Gaussians: 
\begin{align}
\widetilde{\mathbf{A}}_n^{\text{new}} & = [\mathbf{V}_{\bm \beta_n(\mathbf{f})}]_\text{new}^{-1} - \left[\mathbf{A}^{\neg n}\right]^{-1} \,,\\
\widetilde{\mathbf{b}}_n^{\text{new}} & = 
[\mathbf{V}_{\bm \beta_n(\mathbf{f})}]_\text{new}^{-1}[\mathbf{m}_{\bm \beta_n(\mathbf{f})}]_\text{new} -
\left[\mathbf{A}^{\neg n}\right]^{-1}\mathbf{b}^{\neg n}\,.
\end{align}

\subsubsection{Adjusting $\widetilde{\Gamma}(\xopt^j)$  by EP}\label{appendix:section_ep_part_2}

We explain how to adjust the parameters 
$\widetilde{g}_k$ and $\widetilde{h}_k$
of the
approximate factor $\widetilde{\Gamma}(\xopt^j)$ using EP. EP performs
this operation by minimizing the following Kullback-Leibler divergence:
\begin{align}
\text{KL}[\Gamma(\xopt^j)q^{\neg}(\mathbf{f},\mathbf{c}_1,\ldots,\mathbf{c}_K)||
\widetilde{\Gamma}(\xopt^j)q^{\neg}(\mathbf{f},\mathbf{c}_1,\ldots,\mathbf{c}_K)]\,,\label{eq:KL_divergence_2}
\end{align}
where $q^{\neg}(\mathbf{f},\mathbf{c}_1,\ldots,\mathbf{c}_K)$ is the
cavity distribution given by
\begin{align}
q^{\neg}(\mathbf{f},\mathbf{c}_1,\ldots,\mathbf{c}_K) = 
q(\mathbf{f},\mathbf{c}_1,\ldots,\mathbf{c}_K)[\widetilde{\Gamma}(\xopt^j)]^{-1}\,,
\end{align}
We integrate out in $q^{\neg}$ all the variables except those which $\widetilde{\Gamma}(\xopt^j)$ does depend on, 
namely, $c_1(\xopt^j),\ldots,c_K(\xopt^j)$.
Then $q^{\neg}$ takes the form
\begin{align}
q^{\neg}[c_1(\xopt^j),\ldots,c_K(\xopt^j)] \propto 
\prod_{k=1}^K \gaussian(c_k(\mathbf{x}_f^n)\given h^{\neg}_k, g^{\neg}_k)\,,
\end{align}
where the parameters $h^{\neg}_k$ and $g^{\neg}_k$ of these Gaussian
distributions are obtained by using the formula for dividing Gaussians:
\begin{align}
g^{\neg}_k & = \left\{ v_{c_k(\xopt)}^{-1} - \widetilde{g}_k \right\}^{-1}\,, &
h^{\neg}_k & = g^{\neg }_k \left\{ v_{c_k(\mathbf{x}_f^n)}^{-1} m_{c_k(\mathbf{x}_f^n)} - \widetilde{e}_k \right\}^{-1}\,,
\end{align}
where $v_{c_k(\xopt^j)}$ is the variance for $c_k(\xopt^j)$
given by $q(\mathbf{f},\mathbf{c}_1,\ldots,\mathbf{c}_K)$ in
\cref{eq:definition_of_q_appendix}
and $m_{c_k(\xopt^j)}$ is the corresponding mean parameter.

To minimize \cref{eq:KL_divergence_2} we match the 1st and 2nd moments of
$\Gamma(\xopt^j)q^{\neg}(\mathbf{f},\mathbf{c}_1,\ldots,\mathbf{c}_K)$ and
$\widetilde{\Gamma}(\xopt^j)q^{\neg
}(\mathbf{f},\mathbf{c}_1,\ldots,\mathbf{c}_K)]$. The moments of
$\Gamma(\xopt^j)q^{\neg}(\mathbf{f},\mathbf{c}_1,\ldots,\mathbf{c}_K)$ can be
obtained from the derivatives of the logarithm of its normalization constant
$Z$, which is given by
\begin{equation}
Z = 
\int \Gamma(\xopt^j)q^{\neg }(\mathbf{f},\mathbf{c}_1,\ldots,\mathbf{c}_K)\,d\mathbf{f}\,d\mathbf{c}_1\,\cdots d\mathbf{c}_K
 = \prod_{k=1}^K\Phi\left[\alpha_n^k\right]\,,
\end{equation}
where $\alpha_n^k=m_{c_k(\xopt^j)}v_{c_k(\xopt^j)}^{-1/2}$.
We follow Eqs. (5.12) and (5.13) in \citep{minka2001family} to update $\widetilde{g}_k$ and
$\widetilde{h}_k$ in \cref{eq:gamma_tilde}. 
However, we use the second
partial derivative with respect to $g^{\neg}_k$ rather than first partial
derivative with respect to $h^{\neg}_k$ for numerical robustness. These
derivatives are given by
\begin{align}
\frac{\partial \log Z}{\partial h^{\neg}_k} & =
\frac{(Z-1)\phi(\alpha_n^k)}{Z\Phi(\alpha_n^k)\sqrt{g^{\neg}_k}}\,, &
\frac{\partial^2\log Z}{\partial[h^{\neg}_k]^{2}} & =
-\frac{\partial \log Z}{\partial h^{\neg}_k} \cdot \frac{\alpha_n^k}{\sqrt{g^{\neg}_k}}
- \left[\frac{\partial \log Z}{\partial h^{\neg}_k}\right]^2\,.
\end{align}
The update equations for the parameters $\widetilde{g}_k$ and $\widetilde{h}_k$ of the approximate
factor $\widetilde{\Gamma}(\xopt^j)$ are
\begin{align}
\scalemath{0.95}{
[\widetilde{g}_k]_\text{new} = 
-\left\{\left(\frac{\partial^2 \log Z}{\partial [h^{\neg}_k ]^2}\right)^{-1} + g^{\neg}_k \right\}^{-1}\,,\quad
[\widetilde{h}_k]_\text{new} = \left\{ g^{\neg}_k -\left[\frac{\partial^2\log Z}{\partial[h^{\neg}_k]^{2}}\right]^{-1}
\frac{\partial\log Z}{\partial g^{\neg }_k}\right\} [\widetilde{g}_k]^\text{new}\,.\nonumber
}
\end{align}

\subsection{Approximating the Non-Gaussian Factor that Depends on $\mathbf{x}$}

Expectation propagation performs the operations described in
\cref{appendix:section_ep_part_2,appendix:section_ep_part_1} until the Gaussian
approximations to $\Gamma(\xopt^j),
\Psi(\mathbf{x}_f^1),\ldots,\Psi(\mathbf{x}_f^{N_1})$ converge. Importantly
the EP operations described in
\cref{appendix:section_ep_part_2,appendix:section_ep_part_1} can be implemented
independently of the value of $\mathbf{x}$, that is, the location at which we
will be evaluating PESC's acquisition function.
After EP has converged, the next step is to approximate with Gaussians the other factor
in \cref{eq:definition_of_f} that does depend on $\mathbf{x}$, that is,
$\Psi(\mathbf{x})$.
For this, we first replace 
the exact factors
$\Gamma(\xopt^j),
\Psi(\mathbf{x}_f^1),\ldots,\Psi(\mathbf{x}_f^{N_1})$
in \cref{eq:definition_of_f} with their Gaussian approximations. This results in the following approximation:
\begin{align}
f(\mathbf{f},\mathbf{c}_1,\ldots,\mathbf{c}_K|\xopt^j) \approx
\widetilde{f}(\mathbf{f},\mathbf{c}_1,\ldots,\mathbf{c}_K|\xopt^j) = q(\mathbf{f},\mathbf{c}_1,\ldots,\mathbf{c}_K)
\Psi(\mathbf{x})\,,\label{eq:definition_of_f_tilde}
\end{align}
where $q(\mathbf{f},\mathbf{c}_1,\ldots,\mathbf{c}_K)$, as given by \cref{eq:definition_of_q_appendix},
approximates the product of
$p(\mathbf{f},\mathbf{c}_1,\ldots,\mathbf{c}_K|\mathcal{D})$ and
$\Gamma(\xopt^j), \Psi(\mathbf{x}_f^1),\ldots,\Psi(\mathbf{x}_f^{N_1})$ in
\cref{eq:definition_of_f}.
Next, we find a Gaussian approximation to the right-hand-side of \cref{eq:definition_of_f_tilde}.
For this, we first marginalize out in $q$ all the variables except those which 
$\Psi(\mathbf{x})$ does depend on, that is,
$\bm \gamma(\mathbf{f})$ and $c_1(\mathbf{x}),\ldots,c_K(\mathbf{x})$,
where $\bm \gamma(\mathbf{f}) = [ f(\mathbf{x}), f(\xopt^j) ]^\text{T}$,
we obtain
\begin{align}
q[ \bm \gamma(\mathbf{f}), c_1(\mathbf{x}),\ldots,c_K(\mathbf{x})] = 
\mathcal{N}(\bm \gamma(\mathbf{f})|\mathbf{m}_{\gamma(\mathbf{f})},\mathbf{V}_{\gamma(\mathbf{f})})
\left\{\prod_{k=1}^K \mathcal{N}(c_k(\mathbf{x})|m_{c_k(\mathbf{x})},v_{c_k(\mathbf{x})})\right\}\,,\label{eq:marginals_for_q}
\end{align}
where $\mathbf{V}_{\bm \gamma(\mathbf{f})}$ is the $2\times2$ covariance matrix
for $\bm \gamma(\mathbf{f})$ given by
$q(\mathbf{f},\mathbf{c}_1,\ldots,\mathbf{c}_K)$ in
\cref{eq:definition_of_q_appendix}, $\mathbf{m}_{\bm \gamma(\mathbf{f})}$ is
the corresponding $2$-dimensional mean vector, $v_{c_k(\mathbf{x})}$ is the
variance for $c_k(\mathbf{x})$ given by
$q(\mathbf{f},\mathbf{c}_1,\ldots,\mathbf{c}_K)$ in
\cref{eq:definition_of_q_appendix} and $m_{c_k(\mathbf{x})}$ is the
corresponding mean parameter.

Let $\mathbf{m}_1'$ and $\mathbf{V}_1'$ be the mean vector and covariance
matrices for the first $N_1+1$ elements of $\mathbf{f}$ in
\cref{eq:definition_of_f_vector}, according to $q$ in
\cref{eq:definition_of_q_appendix}.
Similarly, $\mathbf{m}_{k+1}'$ and $\mathbf{V}_{k+1}'$ be the mean vector and covariance
matrices for the first $N_1+1$ elements of $\mathbf{c}_k$ in
\cref{eq:definition_of_c_k}, according to $q$ in
\cref{eq:definition_of_q_appendix}, for $k=1,\ldots,K$.
After the execution of EP, we compute
and store $\mathbf{m}_i'$ and $\mathbf{V}_i'$, for $i=1,\ldots,K$.
These parameters can then be used to efficiently compute
$\mathbf{V}_{\bm \gamma(\mathbf{f})}$,
$\mathbf{m}_{\bm \gamma(\mathbf{f})}$, 
$v_{c_1(\mathbf{x})},\ldots,v_{c_K(\mathbf{x})}$ and
$m_{c_1(\mathbf{x})},\ldots,m_{c_K(\mathbf{x})}$ for any arbitrary value of $\mathbf{x}$.
For this, we use Eqs. (3.22) and (3.24) in \citep{Rasmussen2006} to obtain
\begin{align}
\left[ \mathbf{m}_{\bm \gamma(\mathbf{f})} \right]_1 & = 
\mathbf{k}^1(\x)\transpose\left[\mathbf{K}_{\star,\star}^{1}\right]^{-1} \mathbf{m}_1'\,,\nonumber\\
\left[ \mathbf{m}_{\bm \gamma(\mathbf{f})} \right]_{2} & = \left[ \mathbf{m}_1' \right]_{N_1+1}\,,\nonumber\\
\left[ \mathbf{V}_{\bm \gamma(\mathbf{f})} \right]_{1,1} & = k^1(\x,\x)-
\mathbf{k}^{1}(\x)\transpose\left\{ \left[\mathbf{K}_{\star,\star}^{1}\right]^{-1}+\left[\mathbf{K}_{\star,\star}^{1}\right]^{-1}
\mathbf{V}'_1 \left[\mathbf{K}_{\star,\star}^{1}\right]^{-1}\right\} \mathbf{k}^{1}(\x)\,,\nonumber\\
\left[ \mathbf{V}_{\bm \gamma(\mathbf{f})} \right]_{2,2} & = \left[ \mathbf{V}'_1 \right]_{N_1+1,N_1+1}\,,\nonumber\\
\left[ \mathbf{V}_{\bm \gamma(\mathbf{f})} \right]_{1,2} & =
k^{1}(\x,\xopt^j)-\mathbf{k}^{1}(\x)\transpose\left\{ [\mathbf{K}_{\star,\star}^{1}]^{-1}+
[\mathbf{K}_{\star,\star}^{1}]^{-1} \mathbf{V}_1'
[\mathbf{K}_{\star,\star}^{1}]^{-1}\right\} \mathbf{k}^{1}(\xopt^j)\,,\nonumber\\
m_{c_k(\mathbf{x})} & = \mathbf{k}^{k+1}(\x)\transpose\left[\mathbf{K}_{\star,\star}^{k+1}\right]^{-1}\mathbf{m}_{k+1}'\,,\nonumber\\
v_{c_k(\mathbf{x})} & = k^{k+1}(\x,\x)-\mathbf{k}^{k+1}(\x)\transpose\left\{ \left[\mathbf{K}_{\star,\star}^{k+1}\right]^{-1}+
\left[\mathbf{K}_{\star,\star}^{k+1}\right]^{-1} \mathbf{V}_{k+1}' 
\left[\mathbf{K}_{\star,\star}^{k+1}\right]^{-1}\right\} \mathbf{k}^{k+1}(\x)\,,\nonumber
\end{align}
for $k=1,\ldots,K$, where $\mathbf{k}^i(\x)$ is the
$(N_1+1)$-dimensional vector with the prior cross-covariances between the value
of the $i$-th function in $\{f,c_1,\ldots,c_K\}$ at $\x$ and the values of that
function at $\x_f^1,\ldots,\x^{N_1}_f,\xopt^j$, $\mathbf{K}_{\star,\star}^{i}$
is an $(N_1+1)\times(N_1+1)$ matrix with the prior covariances between the
values of that function at $\x_f^1,\ldots,\x^{N_1}_f,\xopt^j$ and $k^i(\x,\x)$
contains the prior variance of the values of that function at $\x$, for $i = 1,\ldots,K+1$. Finally,
$k^1(\x,\xopt^j)$ contains the prior covariance between $f(\x)$ and $f(\xopt^j)$.

Once we have computed the parameters of $q[ \bm \gamma(\mathbf{f})
c_1(\mathbf{x}),\ldots,c_K(\mathbf{x}) ]$ in \cref{eq:marginals_for_q} using
the formulas above, we obtain the marginal means and variances for
$f(\mathbf{x}),c_1(\x),\ldots,c_K(\x)$ with respect to $q[ \bm
\gamma(\mathbf{f}), c_1(\mathbf{x}),\ldots,c_K(\mathbf{x}) ]\Psi(\x)$.
Let $m_1(\x),\ldots,m_{K+1}(\x)$ and $v_1(\x),\ldots,v_{K+1}(\x)$ be these marginal means and variances.
Then, we have the approximation
\begin{align}
\scalemath{0.95}{
\int q[ \bm \gamma(\mathbf{f}), c_1(\mathbf{x}),\ldots,c_K(\mathbf{x}) ]\Psi(\x) df(\xopt^j) \approx
\mathcal{N}(f(\mathbf{x})|m_1(\mathbf{x}),v_1(\mathbf{x}))
\prod_{k=1}^{K} \mathcal{N}(c_k(\x)|m_{k+1}(\mathbf{x}),v_{k+1}(\mathbf{x}))\,,\nonumber
}
\end{align}
where $m_1(\x),\ldots,m_{K+1}(\x)$ and $v_1(\x),\ldots,v_{K+1}(\x)$
can be obtained from
the normalization constant of
$q[ \bm \gamma(\mathbf{f}), c_1(\mathbf{x}),\ldots,c_K(\mathbf{x}) ]\Psi(\x)$
using Eqs. (5.12) and (5.13) in \citep{minka2001family}.
This normalization constant is given by
\begin{equation}
Z = \int q[ \bm \gamma(\mathbf{f}), c_1(\mathbf{x}),\ldots,c_K(\mathbf{x}) ]\Psi(\x)\,d\gamma(\mathbf{f})\,dc_1(\mathbf{x})\,dc_K(\mathbf{x})
= \Phi(\alpha) \prod_{k=1}^K\Phi(\alpha_k) + 1-\prod_{k=1}^K\Phi(\alpha_k)\,,\nonumber
\end{equation}
where
\begin{align}
\alpha_k & = \frac{m_{c_k(\mathbf{x})}}{\sqrt{v_{c_k(\mathbf{x})}}}\,, &
\alpha & = \frac{[1,\,-1]\mathbf{m}_{\bm \gamma(\mathbf{f})}}{\sqrt{s}}\,, & 
s & = \left[\mathbf{V}_{\bm \gamma(\mathbf{f})} \right]_{1,1}+\left[\mathbf{V}_{\bm \gamma(\mathbf{f})} 
\right]_{2,2}-2\left[\mathbf{V}_{\bm \gamma(\mathbf{f})} \right]_{1,2}\,.
\end{align}
Given $Z$, we then compute $m_1(\x),\ldots,m_{K+1}(\x)$ and $v_1(\x),\ldots,v_{K+1}(\x)$ 
using Eqs. (5.12) and (5.13) in \citep{minka2001family}:
\begin{align}
v_1(\x) & = \left[ \mathbf{V}_{\bm \gamma(\mathbf{f})} \right]_{1,1}-\frac{\beta}{s}\left(\beta+\alpha\right)
\left\{\left[ \mathbf{V}_{\bm \gamma(\mathbf{f})}\right]_{1,1}-\left[\mathbf{V}_{\bm \gamma(\mathbf{f})}\right]_{1,2}\right\}^{2}\,,
\label{eq:final_v_f}\\
m_1(\x) & = \left[ \mathbf{m}_{\bm \gamma(\mathbf{f})} \right]_{1}+\left\{\left[  \mathbf{V}_{\bm \gamma(\mathbf{f})} \right]_{1,1}-
\left[  \mathbf{V}_{\bm \gamma(\mathbf{f})} \right]_{1,2}\right\}\frac{\beta}{\sqrt{s}}\,,
\label{eq:final_m_f}\\
v_{k+1}(x) & = \left\{ v_{c_k(\mathbf{x})}^{-1}+\tilde{a}_k\right\} ^{-1}\,, \quad\text{for}\quad k=1,\ldots,K\,,
\label{eq:final_v_k}\\
m_{k+1}(x) & = v_{k+1}(x)\left\{ m_{c_k(\mathbf{x})} v_{c_k(\mathbf{x})}^{-1}+\tilde{b}_k\right\}\,, \quad\text{for}\quad k=1,\ldots,K\,,
\label{eq:final_m_k}
\end{align}
where
\begin{align}
\beta & = Z^{-1} \phi(\alpha)\prod_{k=1}^K\Phi[\alpha_k]\,, &
\tilde{a}_k & = -\left\{ \frac{\partial^2\log Z}{\partial m_{c_k(\mathbf{x})}^2}+ v_{c_k(\mathbf{x})}\right\}^{-1}\,,\\
\tilde{b}_k & = \tilde{a}_k\left\{ m_{c_k(\mathbf{x})}+\frac{\sqrt{v_{c_k(\mathbf{x})}}}{\alpha_k+\beta_k}\right\}\,, &
\frac{\partial^2\log Z}{\partial m_{c_k(\mathbf{x})}^2} & = -\frac{\beta_k\left\{ \alpha_k+\beta_k\right\} }{ v_{c_k(\mathbf{x})} }\,,\\
\beta_k & = \frac{\phi(\alpha_n)}{Z\Phi(\alpha_n)}(Z-1)\,.
\end{align}
\cref{eq:final_v_f,eq:final_m_f,eq:final_m_k,eq:final_v_k}
are the output of our EP algorithm. These quantities are used in
\cref{eq:ep_approximation} to evaluate PESC's acquisition function.

\section{Implementation Considerations}\label{appendix:details}

We give details on the practical implementation of PESC.

\subsection{Initialization, Convergence of EP and Parallel EP Updates}
\label{section:implementation:ep-pesc:initialization-convergence}

We start by fixing the parameters of all the approximate factors
$\widetilde{\Gamma}(\xopt^j),
\widetilde{\Psi}(\mathbf{x}_f^1),\ldots,\widetilde{\Psi}(\mathbf{x}_f^{N_1})$
to be zero. We stop EP when the absolute change in the means and covariance
matrices for the first $N_1+1$  elements of $\mathbf{f}$ and
$\mathbf{c}_1,\ldots,\mathbf{c}_K$ in
\cref{eq:definition_of_f_vector,eq:definition_of_c_k}, according to $q$ in
\cref{eq:definition_of_q_appendix}, is below $10^{-4}$.  The approximate
factors $\widetilde{\Gamma}(\xopt^j),
\widetilde{\Psi}(\mathbf{x}_f^1),\ldots,\widetilde{\Psi}(\mathbf{x}_f^{N_1})$
are updated in parallel to speed up convergence \citep{gerven2009bayesian}.
With parallel updates $q$ in \cref{eq:ep_approximation} is only updated once per
iteration, after all the approximate factors have been refined.

\subsection{EP with Damping}
\label{section:implementation:ep-pesc:damping}

To improve the convergence of EP, we use damping \citep{minka2002damping}. If
$\widetilde{\Psi}(\mathbf{x}_f^n)^\text{new}$ is the value of an approximate
factor that minimizes the KL-divergence, damping entails using instead
$\widetilde{\Psi}(\mathbf{x}_f^n)^\text{damped}$ as the new factor value, as
defined below:
\begin{align}
\widetilde{\Psi}(\mathbf{x}_f^n)^\text{damped} = [\widetilde{\Psi}(\mathbf{x}_f^n)^\text{new}]^\epsilon +
[\widetilde{\Psi}(\mathbf{x}_f^n)^\text{old}]^{1-\epsilon} \, ,
\end{align}
where $\widetilde{\Psi}(\mathbf{x}_f^n)^\text{old}$ is the factor value before
performing the update. We do the same for $\widetilde{\Gamma}(\xopt^j)$.  The parameter
$\epsilon$ controls the amount of damping, with $\epsilon=1$ corresponding to
no damping. We initialize $\epsilon$ to 1 and multiply it by a factor of $0.99$
at each iteration. 

During the execution of EP, some covariance matrices in $q$ or in the cavity
distributions may become non-positive-definite due to an excessively large step
size (i.e. large $\epsilon$). If this issue is encountered during an EP
iteration, the damping parameter is reduced by half and the iteration is
repeated.

% \subsection{Numerical stability of EP}
% \label{section:implementation:ep-pesc:numerical-stability}
% The EP approximation in PESC can easily become numerically unstable unless care is taken. The calculations involve many matrix inversions. These should be avoided whenever possible by replacing the inversion with a Cholesky decomposition plus solving a triangular system. However, matrix inversion cannot be avoided entirely. As mentioned in \cref{section:implementation:GPs}, it is useful to add jitter to the diagonal of the kernel matrix before computing its Cholesky decomposition when making GP predictions. We found that this jitter was not needed in other places; it is sufficient to add it when making GP predictions.

\subsection{Sampling $\xopt$ in PESC}
\label{section:implementation:pesc-sampling-x-star}

We sample~$\xopt$ from its posterior distribution using an extension of the
method described by \citet{hernandez2014} to sample $\xopt$ in the
unconstrained setting. We perform a finite basis approximation to the GPs used
to describe the data for the objective and the constraints. This allows us to
sample analytic approximate samples from the the GP posterior distribution. We
then solve the optimization problem given by \cref{eq:problem}, when the
functions $f$, $c_1,\ldots,c_K$ are replaced by the generated samples. For
this, we use a numerical method for solving constrained optimization problems:
the Method of Moving Asymptotes (MMA) \citep{svanberg2002class-MMA} as
implemented in the NLopt package \citep{nlopt}.  We evaluate the sampled
functions in a uniform grid of size $10^3$ and obtain the best feasible result
in that grid. We add to the points in the uniform grid the evaluation locations
for which we have already collected data.  This is then used as the initial
point for the MMA method. The number of basis functions in the approximation to
the GP is $10^3$. The NLopt convergence tolerance is $10^{-6}$ in the scaled
input space units. 

The finite basis approximation to the GP is given by a Bayesian Gaussian linear
model build on top of a collection of basis functions \citep{hernandez2014}. Drawing an approximate sample
from the GP posterior distribution involves then sampling from the posterior
distribution of that linear model given the observed data.  When the number of
basis functions is larger than the observed data points, this can be done
efficiently as described by \cite{hernandez2014}. In this case, the covariance
matrix of the Gaussian posterior distribution for the linear model is the sum
of a low rank matrix and a diagonal matrix, we can then use an efficient method
to sample from that Gaussian posterior distribution. This method is outlined in
Appendix B.2 of \citet{seeger2008bayesian}. The cost is $\bigO(N^2 M)$ where
$N$ is the number of collected data points and $M$ is the number of basis
functions.  Sampling with the naive method takes $\bigO(M^3)$ operations
because we must take the Cholesky decomposition of an $M\times M$ covariance
matrix. Given that in our implementation $M=10^3$ and typically $N< 100$, 
this method can speed up this sampling procedure by orders of magnitude.
A more efficient implementation 
could also be obtained by using quasi-random numbers to generate the
basis functions, thus reducing the number of basis functions needed to
attain the same approximation quality \citep{YangSAM}.

% for matern we sample from Student T, which do we with Gaussian/Gamma

% future workfuture work: use quazi-random when sampling random feature stuff-- will speed things up a lot-- could use much less random features- for "w" not "randomness"
% two randomness:
% 1. generate random features- - could be quazi
% 2. sample from posterior -- should not because would introduce bias
% dont really need to mention this

\subsection{Cholesky Update in PESC-F}

\label{section:implementation:pesc:rank-one-cholesky-update}

%PESC-F, introduced in \cref{section:pesc:fast-updates}, is a method to perform the optimization step quickly by not updating the GP hyperparameters, the $\xopt$ samples, or the EP approximation to the PESC \acq. 

In PESC-F, during the fast BO computations, the GP hyperparameters (and in
particular the length scales) are not changed from the ones used during last
iteration. Because of this, the GP kernel matrix is unchanged except for the
addition of a new row and column. Given this, we can compute the Cholesky
decomposition of the new kernel matrix with a rank-one update of the Cholesky
decomposition of the current kernel matrix. The $\bigO(N^3)$ computation of the
Cholesky decomposition of the kernel matrix is the main bottleneck for GP-based
\bo. As $N$ gets large, this trick can significantly speed up the fast BO
computations in PESC-F. In fact, this trick applies more generally beyond
PESC-F or even any fast-update method: any \bo method that does not update the
GP hyperparameters at every iteration can take advantage of the rank-one
Cholesky update. This update technique is described in more detail by
\cite{gill1974} and is commonly used in the setting of \bo as seen in
\citep{osborne:2010}.

% \section*{Appendix A.}
% \label{app:theorem}

% Note: in this sample, the section number is hard-coded in. Following
% proper LaTeX conventions, it should properly be coded as a reference:

%In this appendix we prove the following theorem from
%Section~\ref{sec:textree-generalization}:

% In this appendix we prove the following theorem from
% Section~6.2:

\vskip 0.2in
\bibliography{thesis}

\end{document}